\def\year{2021}\relax
\documentclass[letterpaper]{article} 
\usepackage{aaai21}  
\usepackage{times}  
\usepackage{helvet} 
\usepackage{courier}  
\usepackage[hyphens]{url}  
\usepackage{graphicx} 
\usepackage{natbib}  
\usepackage{caption} 
\usepackage{amsmath}
\usepackage{amssymb}
\usepackage{array}
\usepackage{siunitx}
\usepackage{multirow}
\usepackage{subcaption}
\usepackage{appendix}

\newcolumntype{C}[1]{>{\centering\let\newline\\\arraybackslash\hspace{0pt}}m{#1}}
\newcolumntype{L}[1]{>{\raggedright\let\newline\\\arraybackslash\hspace{0pt}}m{#1}}

\newcommand\Tstrut{\rule{0pt}{2.2ex}}       
\newcommand\Bstrut{\rule[-1.0ex]{0pt}{0pt}} 
\newcommand{\TBstrut}{\Tstrut\Bstrut}       

\let\appendixpagenameorig\appendixpagename
\renewcommand{\appendixpagename}{\Large\appendixpagenameorig}

\title{Learning to Sit: Synthesizing Human-Chair Interactions via Hierarchical Control}
\author{
 Yu-Wei Chao,\textsuperscript{\rm 1}\thanks{Work mostly done at the University of Michigan, Ann Arbor.}
 Jimei Yang,\textsuperscript{\rm 2}
 Weifeng Chen,\textsuperscript{\rm 3}
 Jia Deng\textsuperscript{\rm 4} \\
}
\affiliations {
 \textsuperscript{\rm 1}NVIDIA \\
 \textsuperscript{\rm 2}Adobe Research \\
 \textsuperscript{\rm 3}University of Michigan, Ann Arbor \\
 \textsuperscript{\rm 4}Princeton University \\
 {\tt\small ychao@nvidia.com}, {\tt\small jimyang@adobe.com}, {\tt\small wfchen@umich.edu}, {\tt\small jiadeng@cs.princeton.edu}
}
\begin{document}

\maketitle

\begin{abstract}
Recent progress on physics-based character animation has shown impressive
breakthroughs on human motion synthesis, through imitating motion capture data
via deep reinforcement learning. However, results have mostly been demonstrated
on imitating a single distinct motion pattern, and do not generalize to
interactive tasks that require flexible motion patterns due to varying
human-object spatial configurations. To bridge this gap, we focus on one class
of interactive tasks---sitting onto a chair. We propose a hierarchical
reinforcement learning framework which relies on a collection of subtask
controllers trained to imitate simple, reusable mocap motions, and a meta
controller trained to execute the subtasks properly to complete the main task.
We experimentally demonstrate the strength of our approach over different
non-hierarchical and hierarchical baselines. We also show that our approach can
be applied to motion prediction given an image input. A supplementary video can
be found at~\texttt{https://youtu.be/3CeN0OGz2cA}.
\end{abstract}

\section{Introduction}

The capability of synthesizing realistic human-scene interactions is an
important basis for simulating human living space, where robots can be trained
to collaborate with humans, e.g. avoiding collisions or expediting the
completion of assistive tasks.

Motion capture (mocap) data, by offering high quality recordings of articulated
human pose, has provided a crucial resource for human motion synthesis. With
large mocap datasets and deep learning algorithms, kinematics-based approaches
have recently made rapid progress on motion synthesis and
prediction~\cite{fragkiadaki:iccv2015,jain:cvpr2016,holden:siggraph2016,ghosh:3dv2017,butepage:cvpr2017,martinez:cvpr2017,holden:siggraph2017,zhou:iclr2018,li:cvpr2018,gui:eccv2018a,gui:eccv2018b,yan:eccv2018}.
However, the lack of physical interpretability in the synthesized motion has
been a major limitation of these approaches. Such problem becomes impermissible
when it comes to motions that involve substantial human-object or human-human
interactions. Without modeling the physics, the sythensized interactions are
often physically unrealistic, e.g. body parts penetrating obstacles or not
reacting to collision. This generally limits the use of these approaches to
either non-interactive motions, or a carefully set up virtual scene with high
fidelity to the captured one.

\begin{figure}[t]
 \centering
 \begin{minipage}{0.226\textwidth} \centering \includegraphics[width=1.00\textwidth]{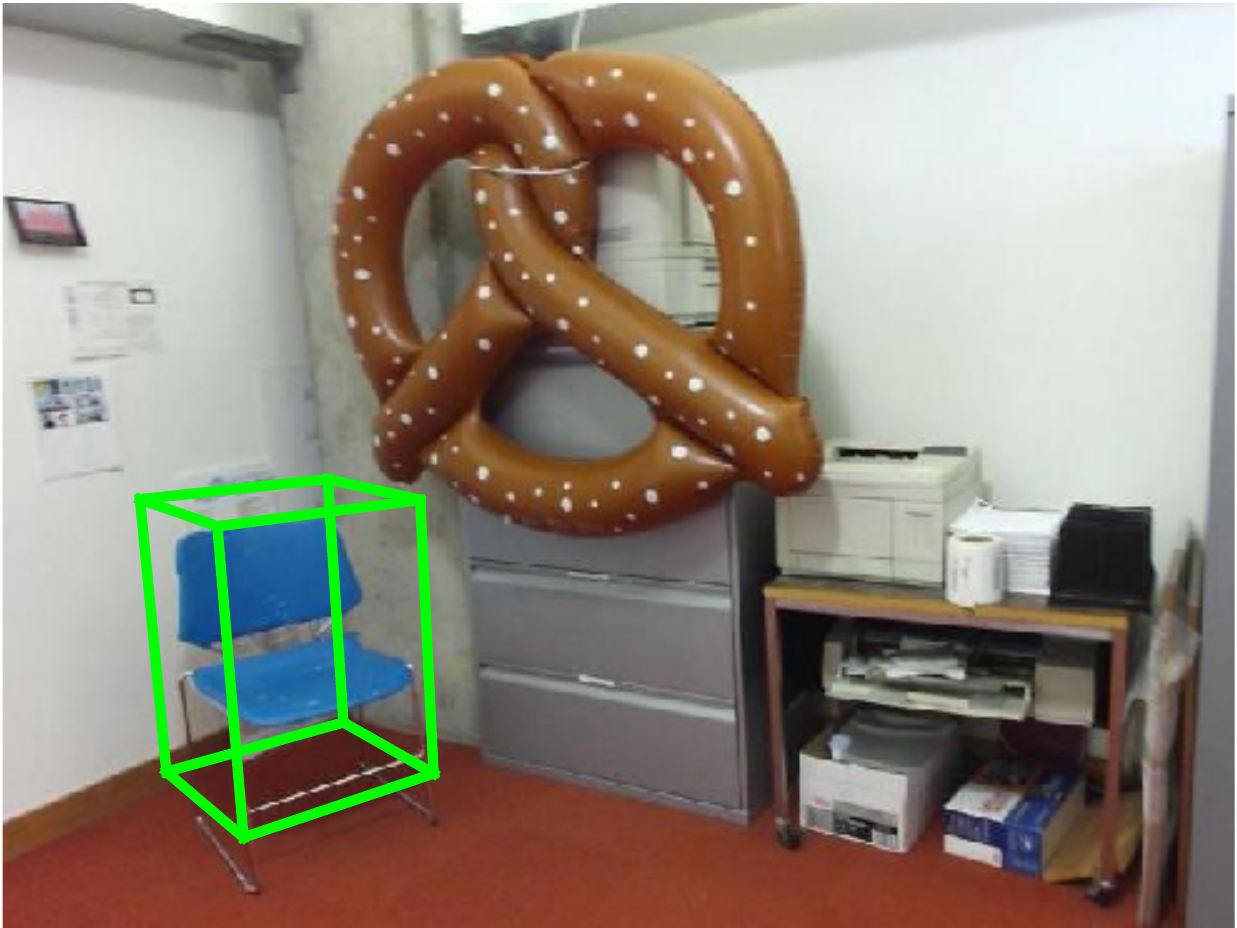} \end{minipage}~
 \begin{minipage}{0.226\textwidth} \centering \includegraphics[width=1.00\textwidth]{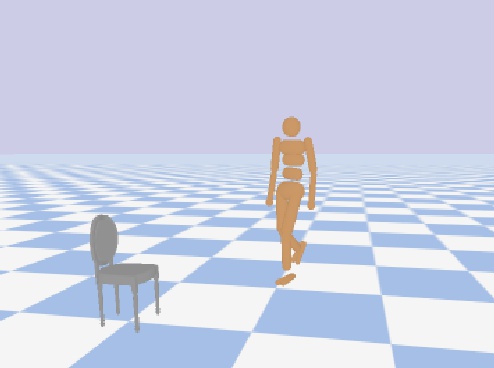} \end{minipage}
 \\ \vspace{1.3mm}
 \begin{minipage}{0.226\textwidth} \centering \includegraphics[width=1.00\textwidth]{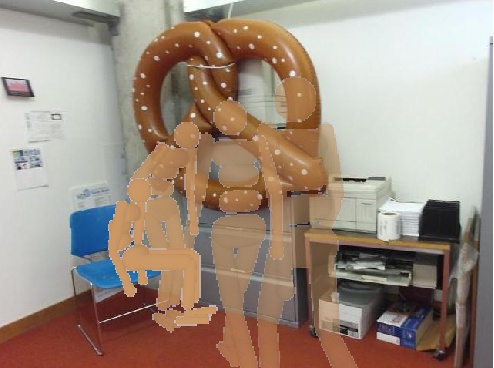} \end{minipage}~
 \begin{minipage}{0.226\textwidth} \centering \includegraphics[width=1.00\textwidth]{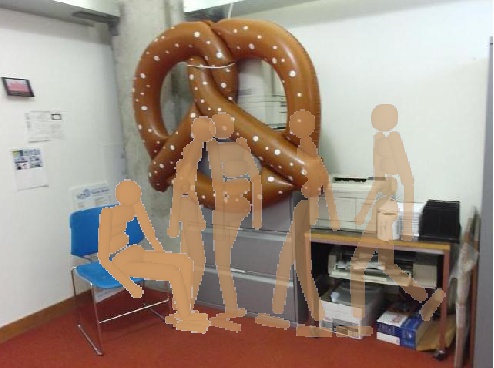} \end{minipage}
 \caption{\small Synthesizing the motion of sitting. \textbf{Top left:} Input
image and 3D chair detection. \textbf{Top right:} Physics simulated environment
for learning human-chair interactions. \textbf{Bottom:} Two examples of
synthesized motions.}
 \label{fig:pull_fig}
\end{figure}

The graphics community has recently witnessed impressive progress on
physics-based character
animation~\cite{peng:siggraph2017,peng:siggraph2018,peng:siggraphasia2018}.
These approaches, through imitating mocap examples via deep reinforcement
learning, can synthesize realistic motions in physics simulated environments.
Consequently, they can adapt to different physical contexts and thus attain a
better generalization performance for interaction-based motions, e.g. walking
on uneven terrain or stunt performance under obstacle disturbance. Nonetheless,
these approaches still suffer from a drawback---a single model is trained for
performing a single task with a distinct motion pattern (often time from a
single mocap clip). As a result, they might not generalize to higher-level
interactive tasks that require flexible motion patterns. Take the example of a
person sitting down on a chair. A person can start in any location and
orientation relative to the chair (Fig.~\ref{fig:pull_fig}). A fixed motion
pattern (e.g. turn left and sit) will be incapable of handling such variations.

In this paper, we focus on one class of high-level interactive tasks---sitting
onto a chair. As earlier mentioned, there are many possible human-chair
configurations and different configurations may require different sequences of
actions to accomplish the goal. For example, if the human is facing the chair,
it needs to walk, turn either left or right, and sit; if the human is behind
the chair, it needs to walk, side-walk and sit. To this end, we propose a
hierarchical reinforcement learning (RL) method to address the challenge of
generalization. Our key idea is the use of hierarchical control: (1) we assume
the main task (e.g. sitting onto a chair) can be decomposed into several
subtasks (e.g. walk, turn, sit, etc.), where the motion of each subtask can be
reliably learned from mocap data, and (2) we train a meta controller using RL
which can execute the subtasks properly to ``complete'' the main task from a
given configuration. Such strategy is in line with the observation that humans
have a repertoire of motion skills, and different subset of skills is selected
and executed for different high-level tasks.

Our contributions are two folds: (1) we extend the prior work on physics-based
motion imitation to the context of higher-level interactive tasks using a
hierarchical approach, and (2) we experimentally demonstrate the strength of
our approach over different non-hierarchical and hierarchical baselines. We
also show in the supplementary video that our approach can be applied to motion
synthesis in human living space with the help of 3D scene reconstruction.

\section{Related Work}

\paragraph{Kinematics-based Models} Kinematic modeling of human motions has a
substantial literature in both vision and graphics domains. Conventional
methods such as motion graphs~\cite{kovar:siggraph2002} require a large corpus
of mocap data and face challenges in generalizing to new behaviors in new
context. Recent progress in deep learning enables researchers to explore more
efficient algorithms to model human motions, again, from large-scale mocap
data. The focus in the vision community is often motion
prediction~\cite{fragkiadaki:iccv2015,jain:cvpr2016,ghosh:3dv2017,butepage:cvpr2017,martinez:cvpr2017,zhou:iclr2018,li:cvpr2018,villegas:cvpr2018,gui:eccv2018a,gui:eccv2018b,yan:eccv2018},
where a sequence of mocap poses is given as historical observation and the goal
is to predict future poses. Recent work has even started to predict motions
directly from a static image~\cite{chao:cvpr2017,walker:iccv2017,yao:cvpr2018}.
In the graphics community, the focus has been primarily on motion synthesis,
which aims to synthesis realistic motions from mocap
examples~\cite{yamane:siggraph2004,agrawal:siggraph2016,holden:siggraph2016,holden:siggraph2017}.
Regardless of the focus, this class of approaches still faces the challenge of
generalization due to the lack of physical plausibility in the synthesized
motion, e.g. foot sliding and obstacle penetrations.

\vspace{-3mm}

\paragraph{Physics-based Models} Physics simulated character animation has a
long history in computer
graphics~\cite{liu:tog2016,liu:tog2017,peng:siggraph2017,liu:siggraph2018,peng:siggraph2018,clegg:siggraphasia2018,peng:siggraphasia2018}.
Our work is most related to the recent work by Peng et
al.~\cite{peng:siggraph2017,peng:siggraph2018}, which trained a virtual
character to imitate mocap data using deep reinforcement learning. They
demonstrated robust and realistic looking motions on a wide array of skills
including locomotion and acrobatic motions. Notably, they have used a
hierarchical model for the task of navigating on irregular
terrain~\cite{peng:siggraph2017}. However, their meta task only requires a
single subtask (i.e. walk), and the meta controller focuses solely on steering.
We address a more complex task (i.e. sitting onto a chair) which requires the
execution of diverse subtasks (e.g. walk, turn, and sit). Another recent work
that is closely related to ours is that of Clegg et
al.~\cite{clegg:siggraphasia2018}, which addressed the task of dressing also
with a hierarchical model. However, their subtasks are executed in a
pre-defined order, and the completion of subtasks is determined by hand-coded
rules. In contrast, our meta controller is trained and is free to select any
subtask at any time point. This is crucial when the main task cannot always be
completed by a fixed order of subtasks.

Humanoid control in physics simulated environments is also a widely-used
benchmark task in the RL community. Some have investigated how to ease the
design of the reward function, but focused less on realistic
motions~\cite{heess:arxiv2017,merel:arxiv2017}. Our work is closest to a few
recent work on hierarchical control of
humanoids~\cite{merel:iclr2019a,merel:iclr2019b,peng:neurips2019}. However,
they focus mainly on locomotion related tasks.

\vspace{-3mm}

\paragraph{Hierarchical Reinforcement Learning} Our model is inspired by a
series of recent work on hierarchical control in deep reinforcement
learning~\cite{heess:arxiv2016,kulkarni:nips2016,tessler:aaai2017}. Although in
different contexts, they share the same attribute that the tasks of concern
have high-dimensional action space, but can be decomposed into simpler,
reusable subtasks. Such decomposition may even help in generalizing to new
high-level tasks due to the shared subtasks.

\vspace{-3mm}

\paragraph{Object Affordances} Our work is connected to the learning of object
affordances in the vision domain. Affordances express the functionality of
objects and how humans can interact with them. Prior work attempted to detect
affordances of a scene, represented as a set of plausible human poses, by
training on large video
corpora~\cite{delaitre:eccv2012,zhu:cvpr2015,wang:cvpr2017}. Instead, we learn
the motion in a physics simulated environment using limited mocap examples and
reinforcement learning. Another relevant work also detected affordances using
mocap data~\cite{gupta:cvpr2011}, but focused only on static pose rather than
motion.

\section{Overview}

\begin{figure*}[t]
 \centering
 \begin{minipage}{0.66\textwidth}
  \centering
  \includegraphics[width=0.96\linewidth]{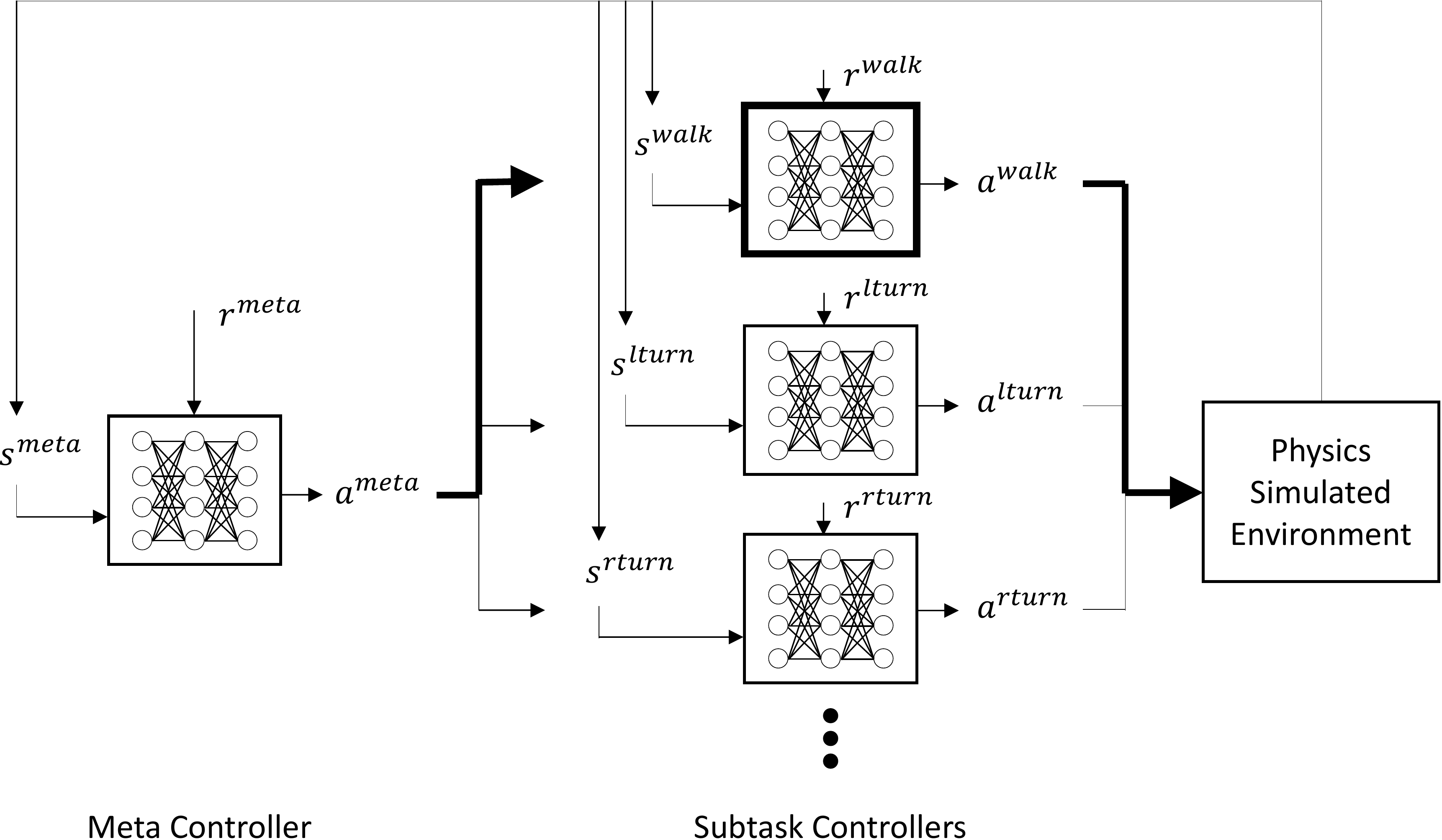}
 \end{minipage}
 \begin{minipage}{0.33\textwidth}
  \centering
  \begin{minipage}{0.41\textwidth}
   \centering
   \begin{subfigure}[b]{0.96\textwidth}
    \includegraphics[width=1.00\textwidth,trim={130 4 50 26},clip]{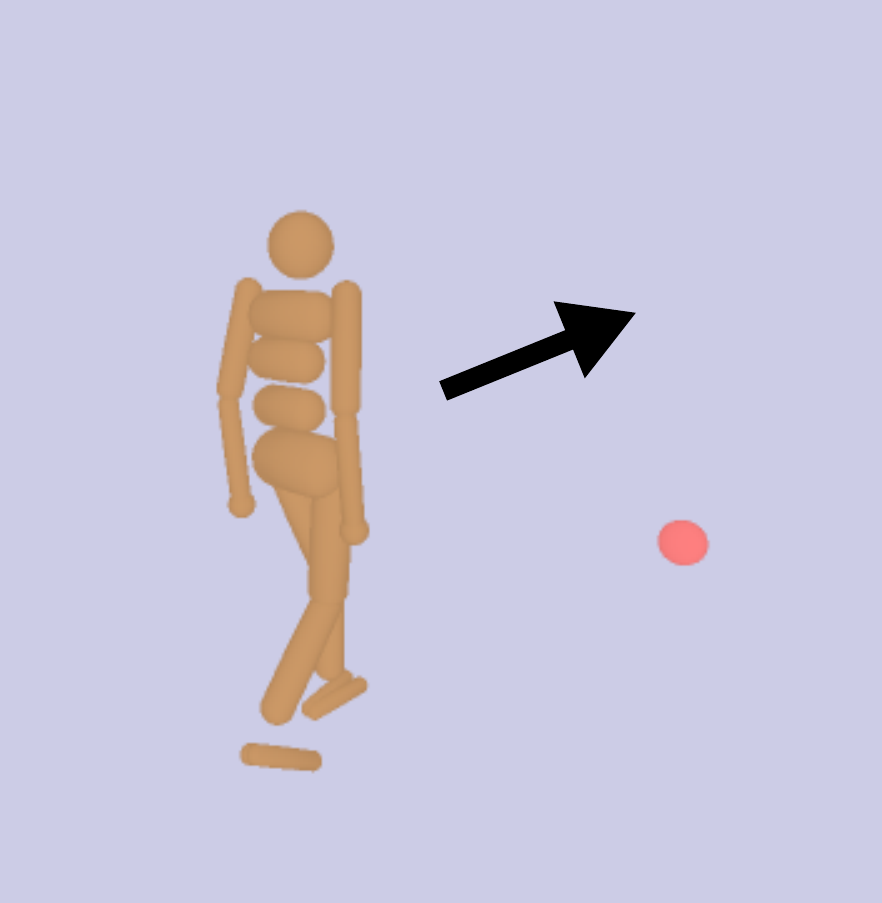}
    \caption{\textit{walk}}
   \end{subfigure}
  \end{minipage}
  ~~
  \begin{minipage}{0.41\textwidth}
   \centering
   \begin{subfigure}[b]{0.96\textwidth}
    \includegraphics[width=1.00\textwidth,trim={90 26 90 4},clip]{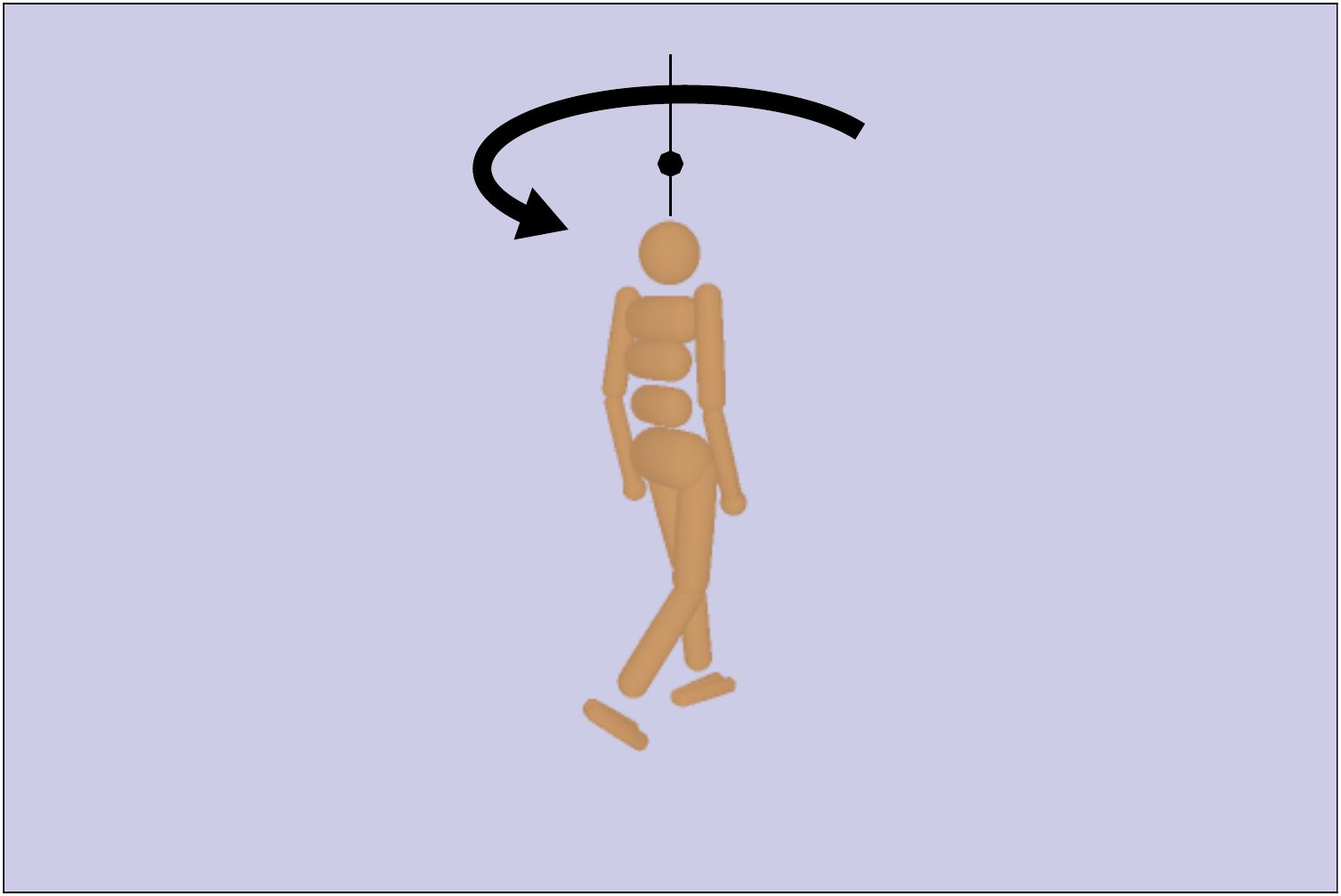}
    \caption{\textit{left turn}}
   \end{subfigure}
  \end{minipage}
  \\ \vspace{1mm}
  \begin{minipage}{0.41\textwidth}
   \centering
   \begin{subfigure}[b]{0.96\textwidth}
    \includegraphics[width=1.00\textwidth,trim={90 26 90 4},clip]{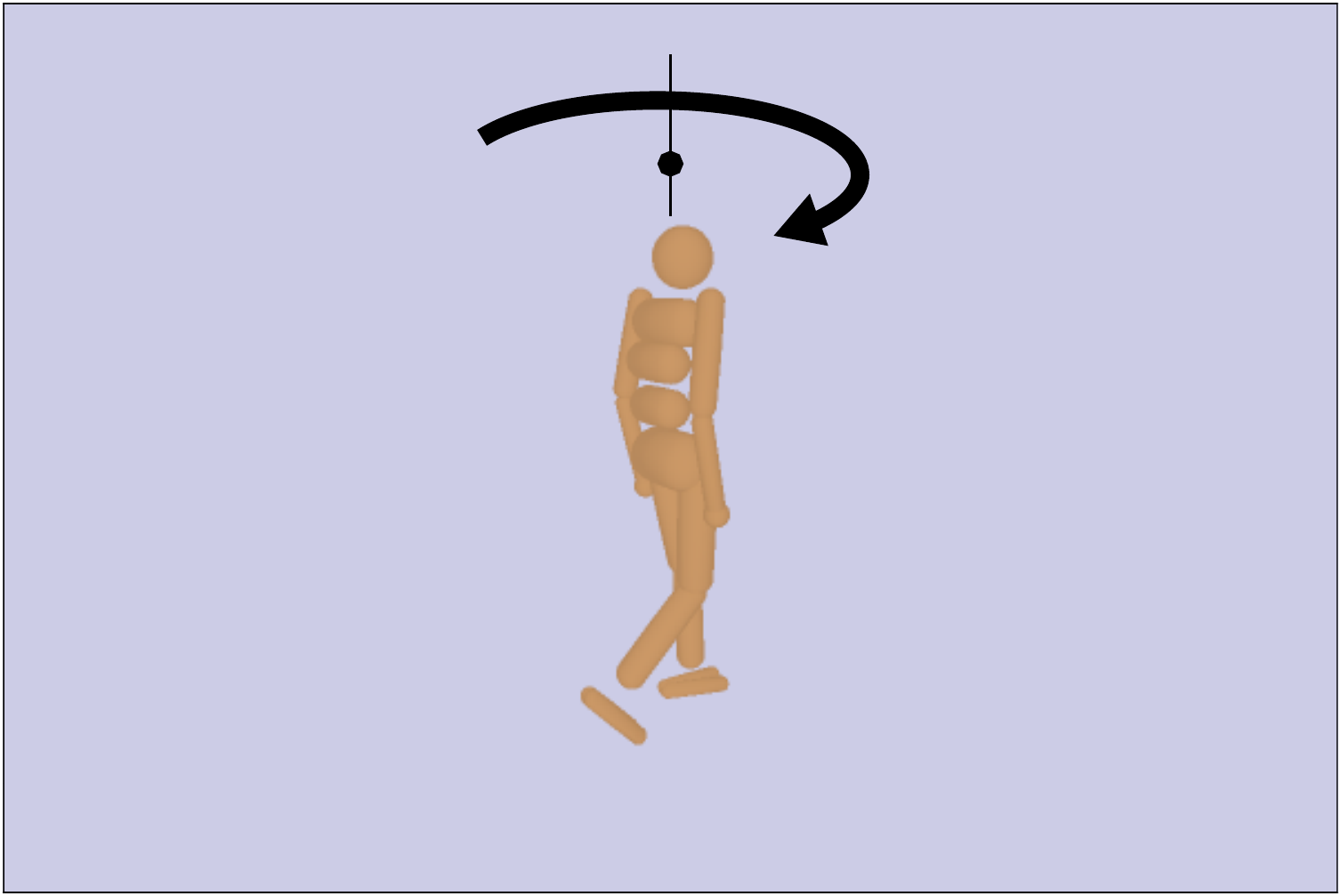}
    \caption{\textit{right turn}}
   \end{subfigure}
  \end{minipage}
  ~~
  \begin{minipage}{0.41\textwidth}
   \centering
   \begin{subfigure}[b]{0.96\textwidth}
    \includegraphics[width=1.00\textwidth,trim={70 4 110 26},clip]{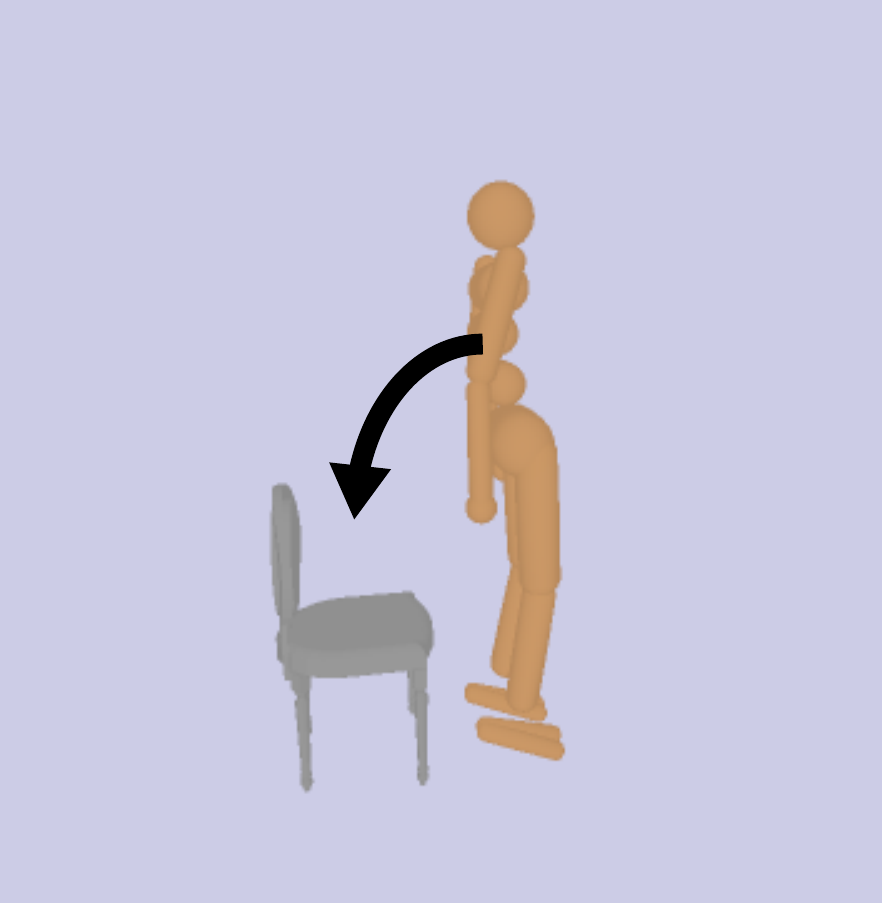}
    \caption{\textit{sit}}
   \end{subfigure}
  \end{minipage}
  \\ \vspace{2mm}
  Subtasks
 \end{minipage}
 \caption{\small \textbf{Left:} Overview of the hierarchical system.
\textbf{Right:} Illustration of the subtasks.}
 \label{fig:hierarchical}
\end{figure*}

Our main task is the following: given a chair and a skeletal pose of a human in
the 3D space, generate a sequence of skeletal poses that describes the motion
of the human sitting onto the chair from the given pose
(Fig.~\ref{fig:pull_fig}). Our system builds upon a physics simulated
environment which contains an articulated structured humanoid and a rigid body
chair model. Each joint of the humanoid (except the root) can receive a control
signal and produce dynamics from the physics simulation. The goal is to learn a
policy that controls the humanoid to successfully sit on the chair.

Fig.~\ref{fig:hierarchical} (left) illustrates the hierarchical architecture of
our policy. At the lower level is a set of subtask controllers, each
responsible for generating the control input of a particular subtask. As
illustrated in Fig.~\ref{fig:hierarchical} (right), we consider four subtasks:
\textit{walk}, \textit{left turn}, \textit{right turn}, and \textit{sit}. To
synthesize realistic motions, the subtask policies are trained on mocap data to
imitate real human motions. At the higher level, a meta controller is
responsible for controlling the execution of subtasks to ultimately accomplish
the main task. The subtask and meta controller run at 60 Hz and 2Hz
respectively, and the physics simulation runs at 240 Hz.

\section{Subtask Controller}

A subtask controller is a policy network $\pi(a_t|s_t)$ that maps a state
vector $s_t$ to an action $a_t$ at each timestep $t$. The state representation
$s$ is extracted from the current configuration of the simulation environment,
and may vary for different subtasks. For example, \textit{turn} requires only
proprioceptive information of the humanoid, while \textit{sit} requires not
only such information, but also the pose of the chair relative to the humanoid.
The action $a$ is the signal for controlling the humanoid joints for each
subtask. We use a humanoid model with 21 degrees of freedom, i.e.
$a\in\mathbb{R}^{21}$. The network architecture is fixed across the subtasks:
we use a multi-layer perceptron with two hidden layers of size 64. The output
of the network parameterizes the probability distribution of $a$, modeled by a
Gaussian distribution with a fixed diagonal covariance matrix, i.e.
$\pi(a|s)=\mathcal{N}(\mu(s),\Sigma)$ and $\Sigma=diag(\{\sigma_i\})$. We can
generate $a_t$ at each timestep by sampling from $\pi(a_t|s_t)$.

Each subtask is formulated as an independent RL problem. At timestep $t$, the
state $s_t$ given by the simulation environment is fed into the policy network
to output an action $a_t$. The action $a_t$ is then fed back to the simulation
environment to generates the state $s_{t+1}$ at the next timestep and a reward
signal $r_t$. The design of the reward function is crucial and plays a key role
in shaping the style of the humanoid's motion. A heuristically crafted reward
may yield a task achieving policy, but may result in unnatural looking motions
and behaviors~\cite{heess:arxiv2017}. Inspired by~\cite{peng:siggraph2018}, we
set the reward function of each subtask by a sum of two terms:
\begin{equation}
 r^{sub} = r^{S} + r^{G}.
\end{equation}
where $r^{S}$ encourages similar motion to the mocap reference and $r^{G}$
encourages the achievement of the subtask goal. We use a consistent similarity
reward $r^{S}$ across all subtasks:
\begin{equation}
 r^{S} = \omega^{p}r^{p} + \omega^{v}r^{v},
\end{equation}
where $r^{p}$ and $r^{v}$ encourage the similarity of local joint angles $q_j$
and velocities $\dot{q}_j$ between the humanoid and the reference motion, and
$\omega^{p}$ and $\omega^{v}$ are the respective weights. Specifically,
\begin{equation}
 \begin{aligned}
  &r^{p} = \exp\left(-\alpha^{p}\sum_jd(q_j,\hat{q}_j)^2\right) \\
  &r^{v} = \exp\left(-\alpha^{v}\sum_j(\dot{q}_j-\hat{\dot{q}}_j)^2\right),
 \end{aligned}
\end{equation}
where $d(\cdot,\cdot)$ computes the angular difference between two angles. We
empirically set $\omega^{p} = 0.5$, $\omega^{v} = 0.05$, $\alpha^{p} = 1$, and
$\alpha^{v} = 10$. Next, we detail the state representation $s$ and the goal
reward $r^{G}$ for each subtask.

\vspace{-3mm}

\begin{figure}[t]
 \centering
 \includegraphics[width=0.96\linewidth]{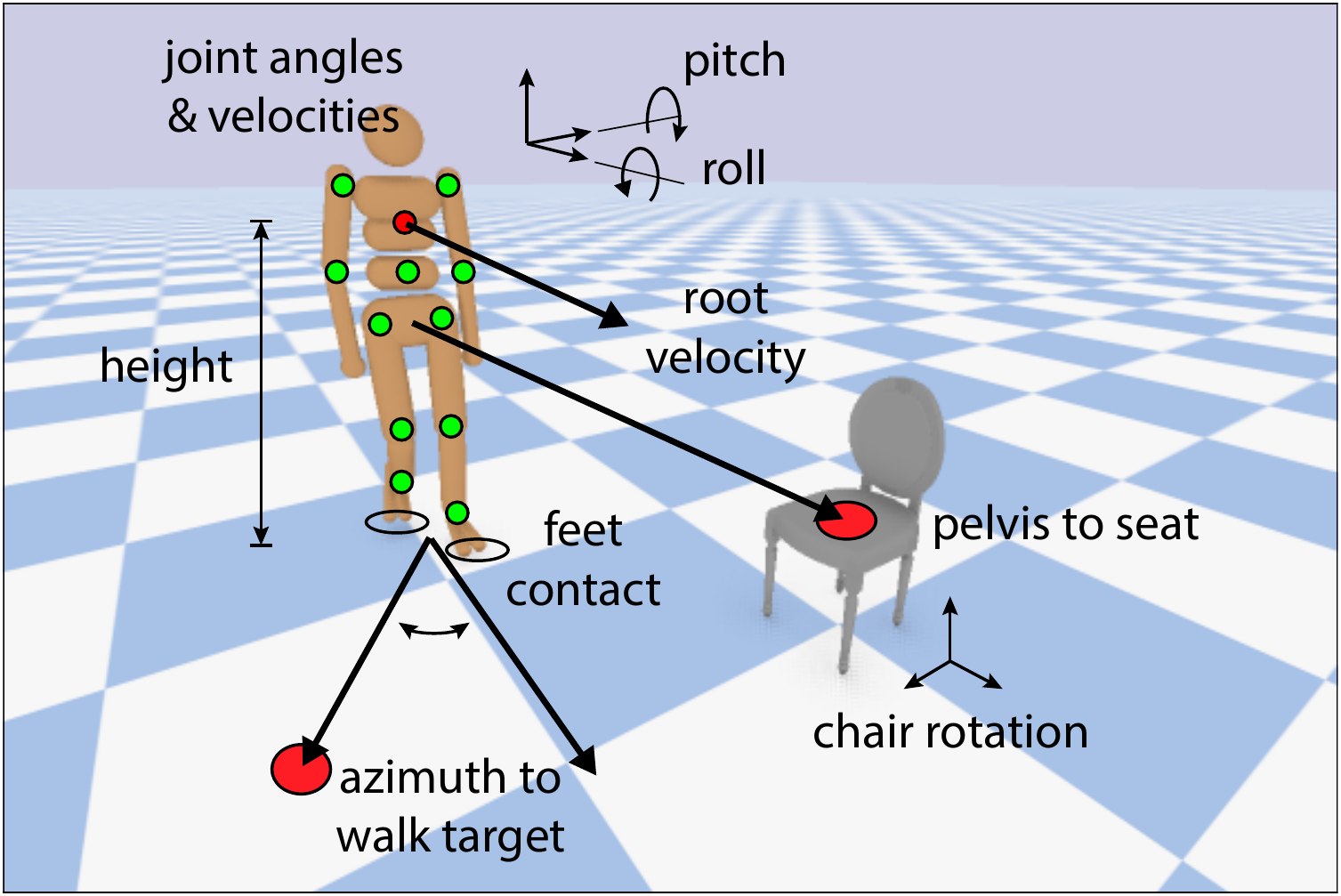}
 \caption{\small State representation of the humanoid and chair. The red and
green dots on the humanoid denote the root and non-root joints. The red dots on
the ground and chair denote the walk target and the center of the seat
surface.}
 \label{fig:state}
\end{figure}

\paragraph{1) Walk} The state $s^{walk}\in\mathbb{R}^{52}$ consists of a 50-d
proprioceptive feature and a 2-d goal feature that specifies an intermediate
walking target. The proprioceptive feature includes the local joint angles and
velocities, the height and linear velocity of the root (i.e. torso) as well as
its pitch and roll angles, and a 2-d binary vector indicating the contact of
each foot with the ground (Fig.~\ref{fig:state}). Rather than walking in random
directions, target-directed locomotion~\cite{agrawal:siggraph2016} is necessary
for accomplishing high-level tasks. Assuming a target is given, represented by
a 2D point on the ground plane, the 2-d goal feature is given by $[\sin(\psi),
\cos(\psi)]^{\top}$, where $\psi$ is the azimuth angle to the target in the
humanoid centric coordinates. The generation of targets will be detailed in the
Meta Controller section.

We observe that it is challenging to directly train a target-directed walking
policy with mocap examples. Therefore we adopt a two-stage training strategy
where each stage uses a distinct goal reward. In the first stage, we encourage
similar steering patterns to the reference motion, i.e. the linear velocity of
the root $v\in\mathbb{R}^3$ should be similar between the humanoid and
reference motion:
\begin{equation}
 r^{G} = 0.5\cdot\exp\left(-10\cdot\sum_i(v_i-\hat{v_i})^2\right).
\end{equation}
In the second stage, we reward motion towards the target:
\begin{equation}
 r^{G} = 0.1\cdot\frac{1}{1+\exp(10\cdot{V^{walk}})},
\end{equation}
where $V^{walk}=(D^{walk}_{t+1}-D^{walk}_t)/\delta{t}$. $D^{walk}_t$ denotes
the horizontal distance between the root and the target, and $\delta{t}$ is the
length of the timestep.

\vspace{-3mm}

\paragraph{2) Left/Right Turn} The states $s^{lturn}$,
$s^{rturn}\in\mathbb{R}^{50}$ reuse the 50-d proprioceptive feature from the
walk subtask. The goal reward encourages the rotation of the root to be matched
between the humanoid and reference motion:
\begin{equation}
 r^{G} = 0.1\cdot\exp\left(-10\cdot\sum_id(\theta_i,\hat{\theta_i})^2\right),
\end{equation}
where $\theta\in\mathbb{R}^3$ consists of the root's pitch, yaw, and roll.

\vspace{-3mm}

\paragraph{3) Sit} The sit subtask assumes that the humanoid is initially
standing roughly in front of the chair and facing away. The task is simply to
lower the body and be seated. Different from \textit{walk} and \textit{turn},
the state for \textit{sit} should capture the pose information of the chair.
Our state $s^{sit}\in\mathbb{R}^{57}$ consists of the same 50-d proprioceptive
feature used in \textit{walk} and \textit{turn}, and additionally a 7-d feature
describing the state of the chair in the humanoid centric coordinates. The 7-d
chair state includes the displacement vector from the pelvis to the center of
the seat surface, and the rotation of the chair in the humanoid centric
coordinates represented as a quaternion (Fig.~\ref{fig:state}). The goal reward
encourages the pelvis to move towards the center of the seat surface:
\begin{equation}
 r^{G} = 0.5\cdot(-V^{sit}),
 \label{eq:sit}
\end{equation}
where $V^{sit}=(D^{sit}_{t+1}-D^{sit}_t)/\delta{t}$ and $D^{sit}_t$ is the 3D
distance between the pelvis and the center of the seat surface.

\section{Meta Controller}

The meta controller is also a policy network and shares the same architecture
as the subtask controllers. We reuse the 57-d state representation from the sit
subtask which contains both the proprioceptive and chair information. Rather
than directly controlling the humanoid joints, the output action $a^{meta}$ now
controls the execution of subtasks. Specifically,
$a^{meta}=\{a^{switch},a^{target}\}$ consists of two components.
$a^{switch}\in\{\text{\textit{walk}},\text{\textit{left
turn}},\text{\textit{right turn}},\text{\textit{sit}}\}$ is a discrete output
which at each timestep picks a single subtask out of the four to execute.
$a^{target}\in\mathbb{R}^2$ specifies the 2D target for the walk subtask, which
is used to compute the goal state in $s^{walk}$. Note that $a^{target}$ is only
used when the walk subtask is picked for execution. The output of the policy
network parameterizes the probability distributions of both $a^{switch}$ and
$a^{target}$, where $a^{switch}$ is modeled by a categorical distribution as in
standard classification problems, and $a^{target}$ is modeled by a Gaussian
distribution following the subtask controllers.

The meta task is also formulated as an independent RL problem. At timestep $t$,
the policy network takes the state $s^{meta}_t$ from the simulation environment
and output an action $a^{meta}_t$. $a^{meta}_t$ then triggers one specific
subtask controller to generate the control signal for the humanoid joints. The
control signal is finally fed back to the simulation to generate the next state
$s^{meta}_{t+1}$ and a reward $r^{meta}_t$. Rather than evaluating the
similarity to a mocap reference, the reward now should be providing feedback on
the main task. We adopt a reward function that encourages the pelvis to move
towards and be in contact with the seat surface:
\begin{equation}
 r^{meta} =
 \left\{\begin{array}{cl}
  1                  & \text{if } z_{\text{contact}}=1 \\[1mm]
  0.5\cdot(-V^{sit}) & \text{otherwise}.
 \end{array}\right.
\end{equation}
$z_{\text{contact}}$ indicates whether the pelvis is in contact with the seat
surface, which can be detected by the physics simulator. $V^{sit}$ is defined
as in Eq.~\ref{eq:sit}.

\section{Training}

Since the subtasks and meta task are formulated as independent RL problems,
they can be trained independently using standard RL algorithms. We first train
each subtask controllers separately, and then train the meta controller using
the trained subtask controllers. All controllers are trained in a standard
actor-critic framework using the proximal policy optimization (PPO)
algorithm~\cite{schulman:arxiv2017}.

\vspace{-3mm}

\paragraph{1) Subtask Controller} The training of the subtasks is divided into
two stages. First, in each episode, we initialize the pose of the humanoid to
the first frame of the reference motion, and train the humanoid to execute the
subtask by imitating the following frames. This enables the humanoid to perform
the subtasks from the initial pose of the reference motion, but does not
guarantee successful transitions between subtasks (e.g.
\textit{walk}$\rightarrow$\textit{turn}), which is required for the main task.
Therefore in the second stage, we fine-tune the controllers by setting the
initial pose to a sampled ending pose of another subtask, similar to the policy
sequencing method in~\cite{clegg:siggraphasia2018}. For \textit{turn} and
\textit{sit}, the initial pose is sampled from the ending pose of \textit{walk}
and \textit{turn}, respectively.

\vspace{-3mm}

\begin{figure}[t]
 \centering
 \includegraphics[width=0.90\linewidth,trim={33 36 27 38},clip]{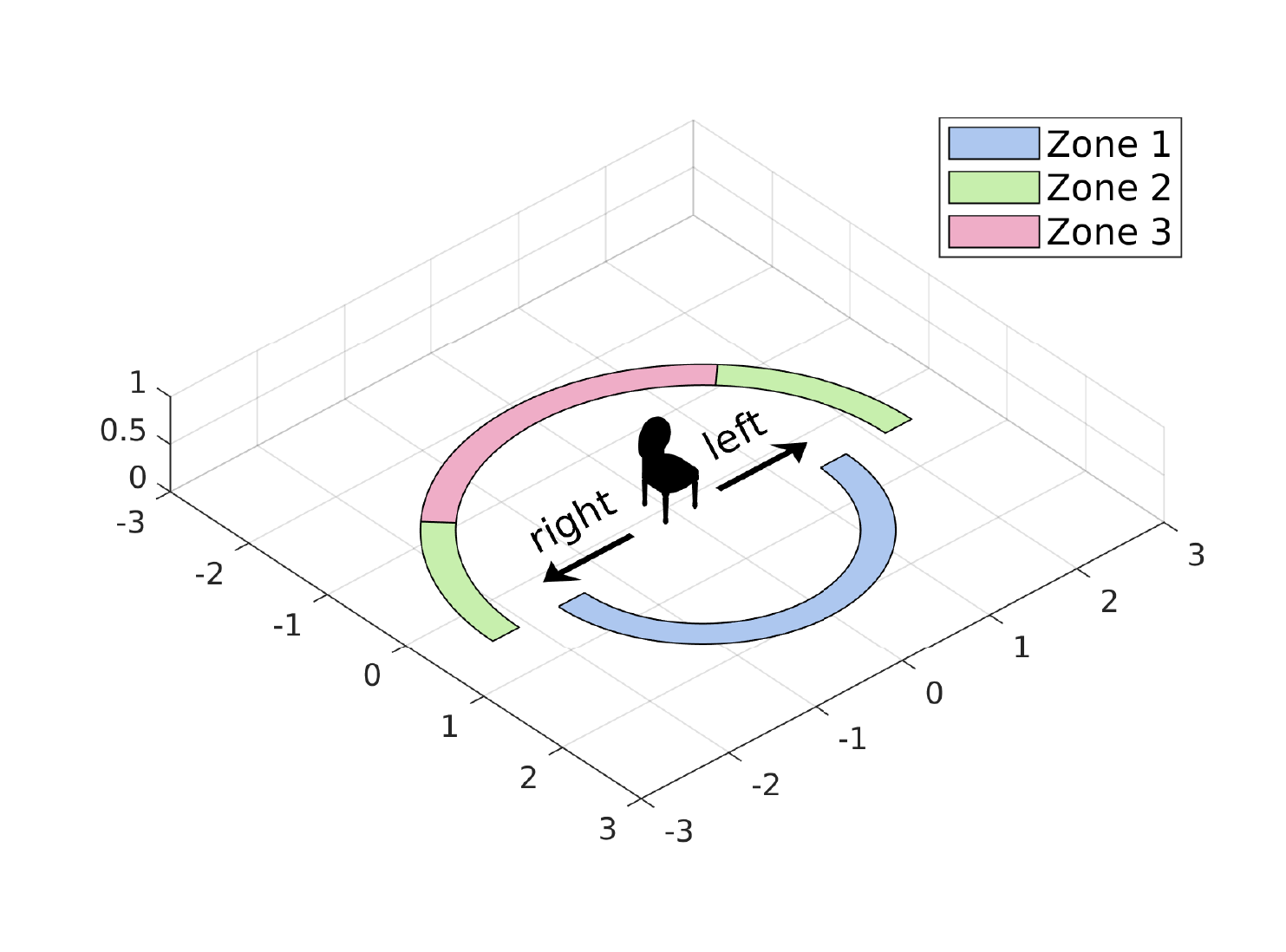}
 \caption{\small Curriculum learning for the meta controller. Humanoid spawn
location is initially set to less challenging states (Zone 1), and later moved
to more challenging states (Zone 2 and 3).}
 \label{fig:curriculum}
\end{figure}

\paragraph{2) Meta Controller} While our goal is to have the humanoid sit down
regardless of where it starts in the environment, the task's difficulty highly
depends on the initial state: if it is already facing the seat, it only needs
to turn and sit, while if it is behind the chair, it needs to first walk to the
front and then sit down. Training can be challenging when starting from a
difficult state, since the humanoid needs to by chance execute a long sequence
of correct actions to receive the reward for sitting down. To facilitate
training, we propose a multi-stage training strategy inspired by curriculum
learning~\cite{zaremba:arxiv2014}. The idea is to begin the training from
easier states, and progressively increase the difficulty when the training
converges. As illustrated in Fig.~\ref{fig:curriculum}, we begin by only
spawning the humanoid on the front side of the chair (Zone 1). Once trained, we
change the initial position to the lateral sides (Zone 2) and continue the
training. Finally, we train the humanoid to start from the rear side (Zone 3).

\section{Results}

\paragraph{Data and Implementation Details} The mocap references for each
subtask are collected from the CMU Graphics Lab Motion Capture
Database~\cite{cmu-mocap}. We extract relevant motion segments and retarget the
motion to our humanoid model. We use a 21-DoF humanoid model provided by the
Bullet Physics SDK~\cite{coumans:2019}. Motion retargeting is performed using a
Jacobian-based inverse kinematics method~\cite{holden:siggraph2016}. Our
simulation environment is based on OpenAI
Roboschool~\cite{schulman:arxiv2017,openai:roboschool}, which uses the Bullet
physics engine~\cite{coumans:2019}. We use a randomly selected chair model from
ShapeNet~\cite{chang:arxiv2015}. The PPO algorithm for training is based on the
implementation from OpenAI Baselines~\cite{openai:baselines}.

\vspace{-3mm}

\paragraph{Evaluation of Main Task} We adopt two different metrics to
quantitatively evaluate the main task: (1) \textit{success rate} and (2)
\textit{minimum distance}. We declare a success whenever the pelvis of the
humanoid has been continuously in contact with the seat surface for 3.0
seconds. We report the success rate over 10,000 trials by spawning the humanoid
at random locations. Note that the success rate evaluates task completion with
a hard constraint and does not reveal the progress when the humanoid fails.
Therefore we also compute the per-trial minimum distance (in meters) between
the pelvis and the center of the seat surface, and report the mean and standard
deviation over the 10,000 trials.

As noted in the last section, the task can be challenging when the initial
position of the humanoid is unconstrained. To better analyze the performance,
we consider two different initialization settings: (1) \textit{Easy} and (2)
\textit{Hard}. In the Easy setting, the humanoid is initialized from roughly 2
meters away on the front half plane of the chair (i.e. Zone 1 in
Fig.~\ref{fig:curriculum}), with an orientation roughly towards the chair. The
task is expected to be completed by simply walking forward, turning around, and
sitting down. In the Hard setting, humanoid is initialized again from roughly 2
meters away but on the lateral and rear sides of the chair (i.e. Zone 2 and 3
in Fig.~\ref{fig:curriculum}). It needs to walk around the chair to sit down
successfully.

\begin{table}[t]
 \centering
 \scriptsize
 \setlength{\tabcolsep}{5pt}
 \begin{tabular}{l||cc}
  \hline \TBstrut
                                                                                                       & Succ Rate (\%) & Min Dist (m)                 \\
  \hline \Tstrut
  Kinematics                                                                                           & --             & 1.2656 $\pm$ 0.0938          \\ \Bstrut
  Physics~\cite{peng:siggraph2018}                                                                     & ~~0.00         & 1.3316 $\pm$ 0.1966          \\
  \hline \Tstrut
  \textit{walk}$\rightarrow$\textit{left turn}$\rightarrow$\textit{sit}~\cite{clegg:siggraphasia2018}  & 25.16          & 0.3790 $\pm$ 0.2326          \\ \Bstrut
  \textit{walk}$\rightarrow$\textit{right turn}$\rightarrow$\textit{sit}~\cite{clegg:siggraphasia2018} & ~~0.92         & 0.7948 $\pm$ 0.2376          \\
  \hline \Tstrut
  \textit{walk} / \textit{left turn} / \textit{sit}                                                    & 29.38          & 0.3913 $\pm$ 0.2847          \\
  \textit{walk} / \textit{right turn} / \textit{sit}                                                   & 23.01          & 0.3620 $\pm$ 0.2378          \\ \Bstrut
  Full Model                                                                                           & \textbf{31.61} & \textbf{0.3303 $\pm$ 0.2393} \\
  \hline
 \end{tabular}
 \caption{\small Comparison of our approach with the non-hierarchical and
hierarchical baselines in the Easy setting.}
 \label{tab:baselines}
\end{table}

\begin{figure*}[t]
 \centering
 \begin{minipage}{0.12\textwidth} \centering \includegraphics[width=1.00\textwidth,trim={80 40 80 40},clip]{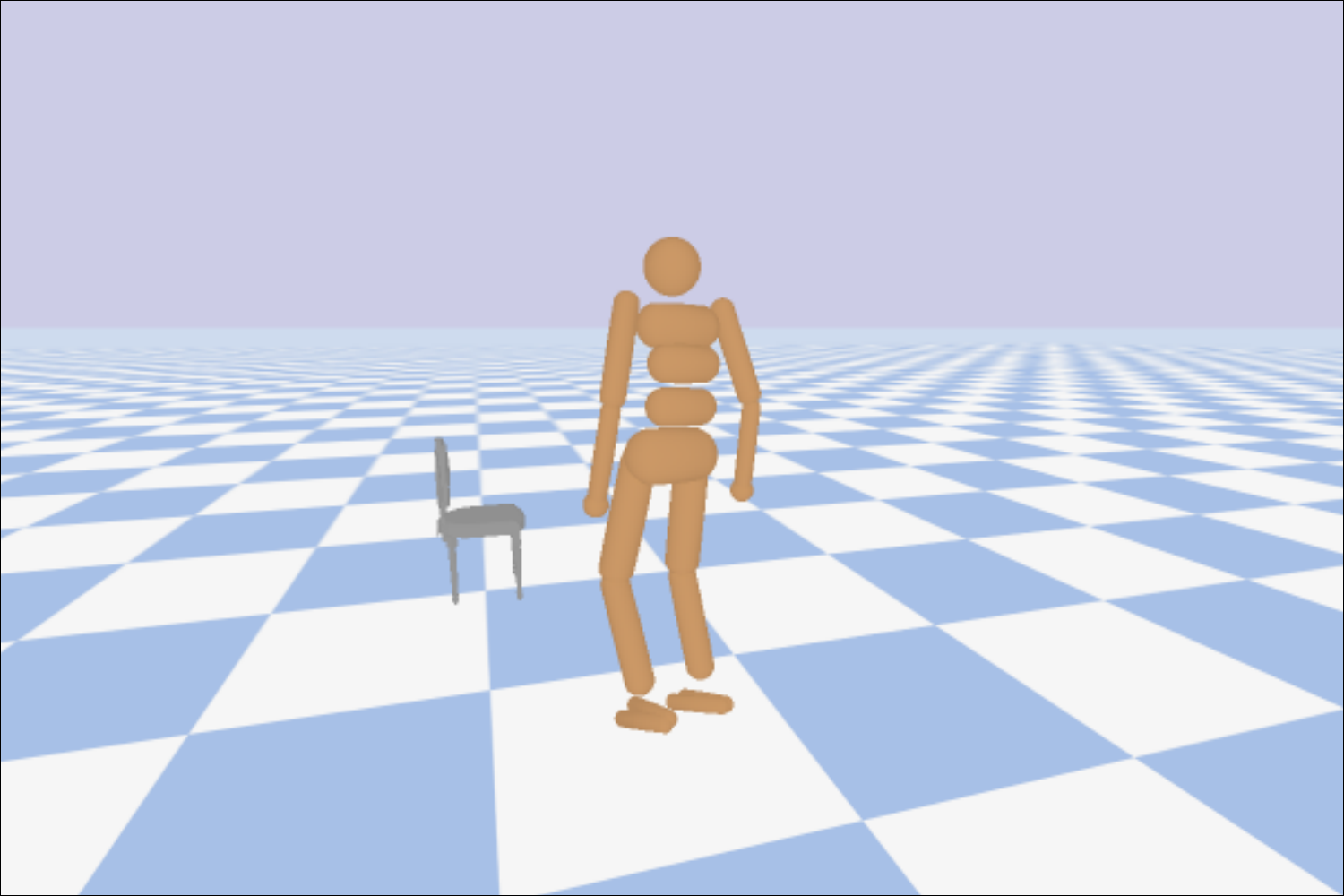} \end{minipage}
 \begin{minipage}{0.12\textwidth} \centering \includegraphics[width=1.00\textwidth,trim={80 40 80 40},clip]{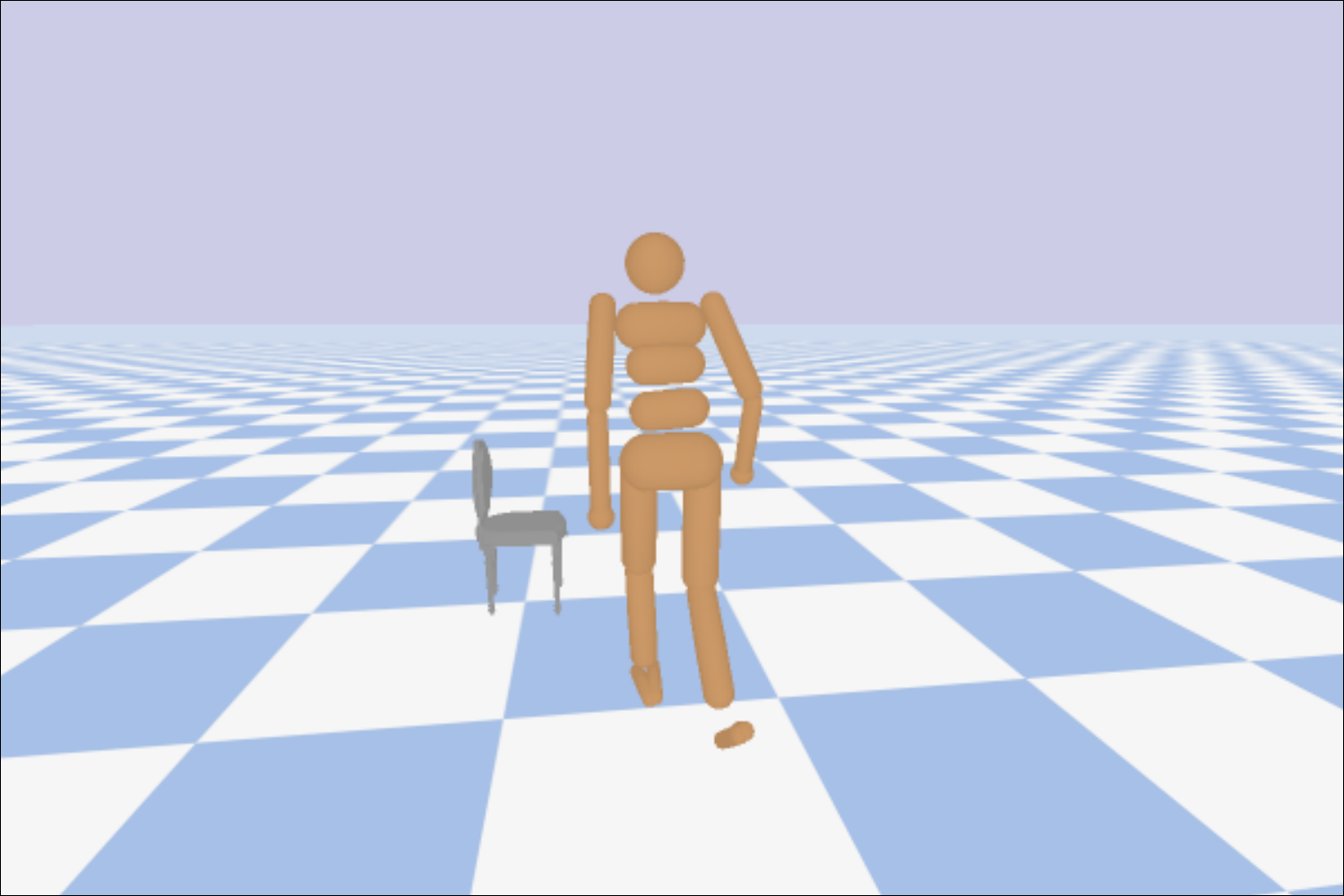} \end{minipage}
 \begin{minipage}{0.12\textwidth} \centering \includegraphics[width=1.00\textwidth,trim={80 40 80 40},clip]{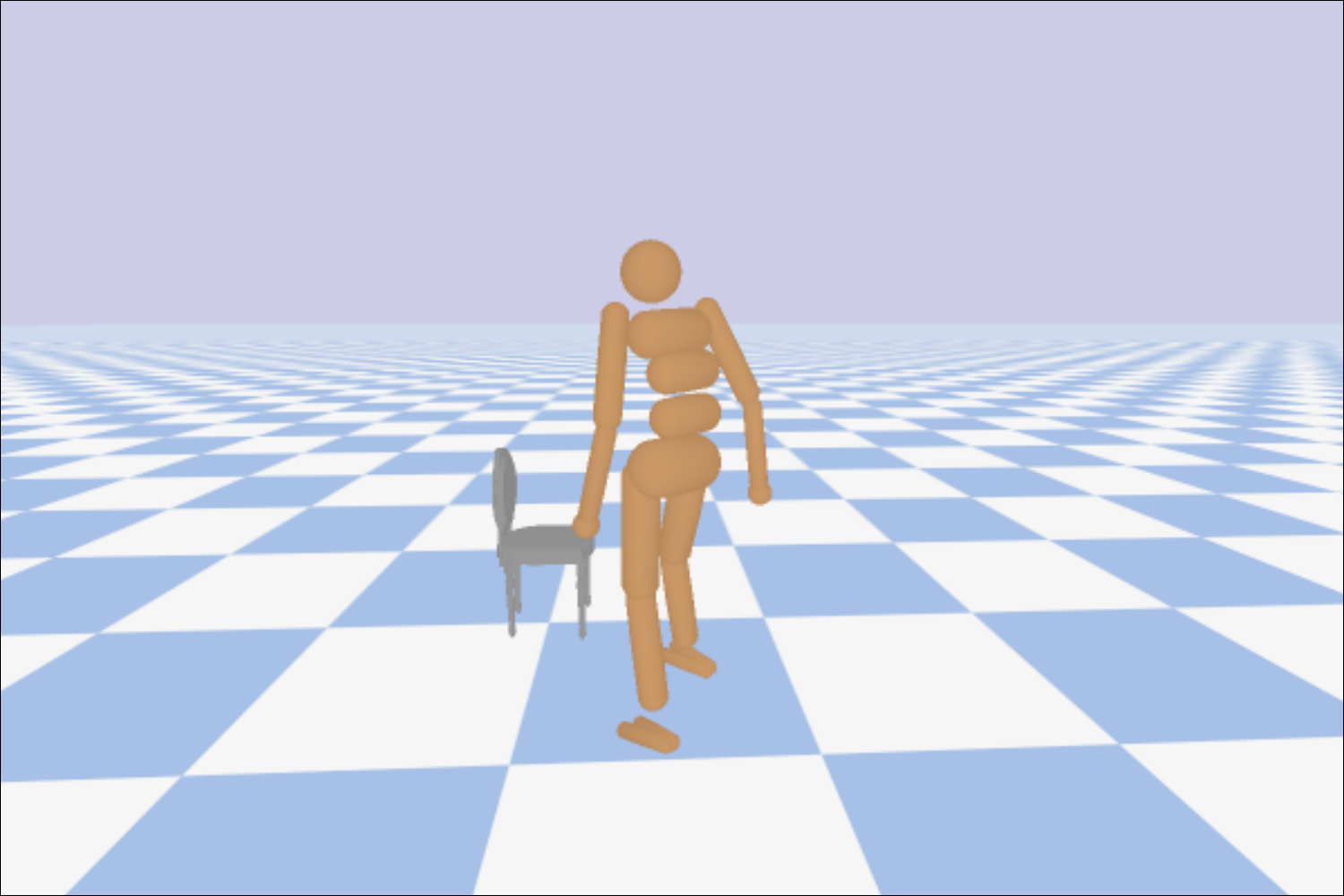} \end{minipage}
 \begin{minipage}{0.12\textwidth} \centering \includegraphics[width=1.00\textwidth,trim={80 40 80 40},clip]{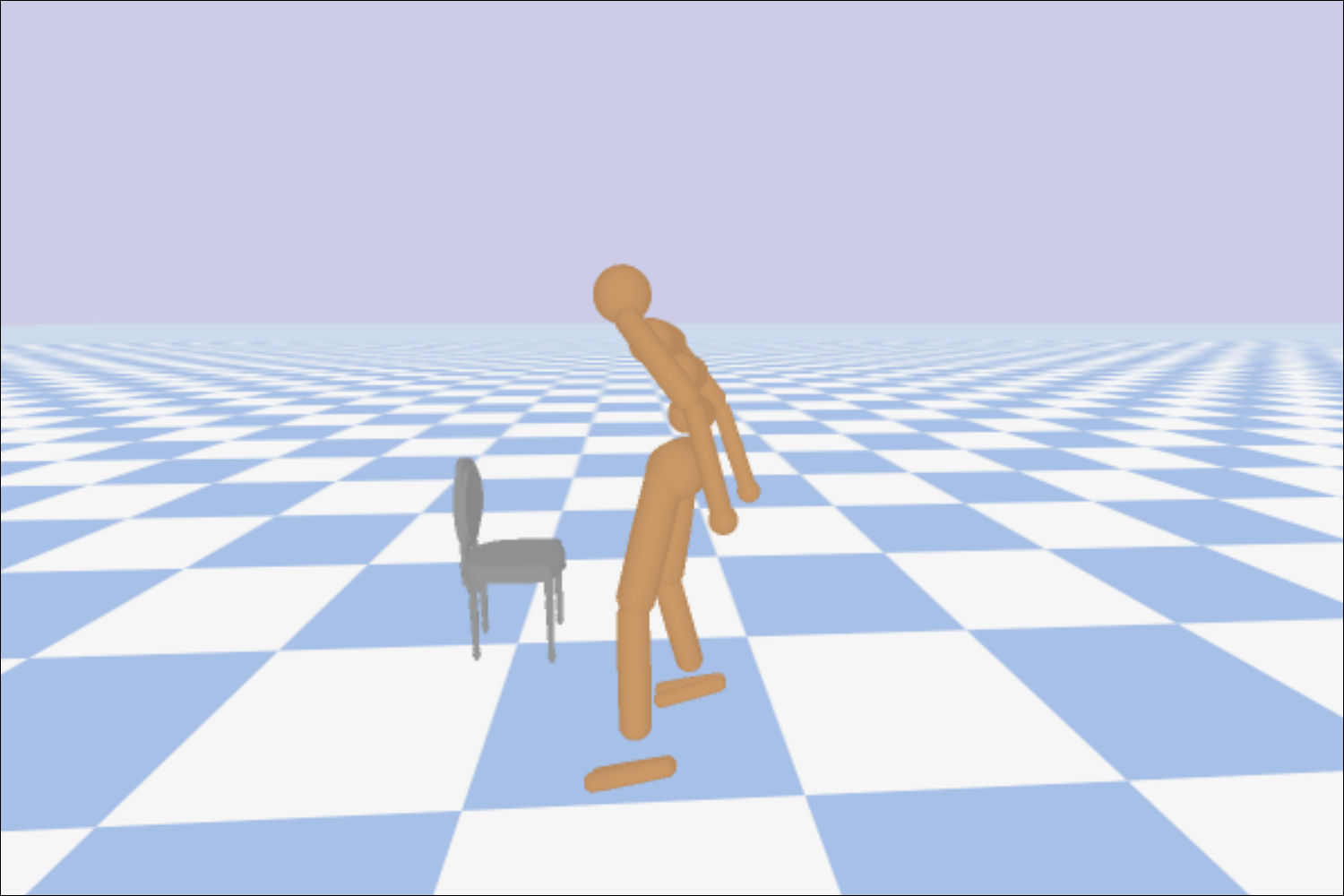} \end{minipage}
 \begin{minipage}{0.12\textwidth} \centering \includegraphics[width=1.00\textwidth,trim={80 40 80 40},clip]{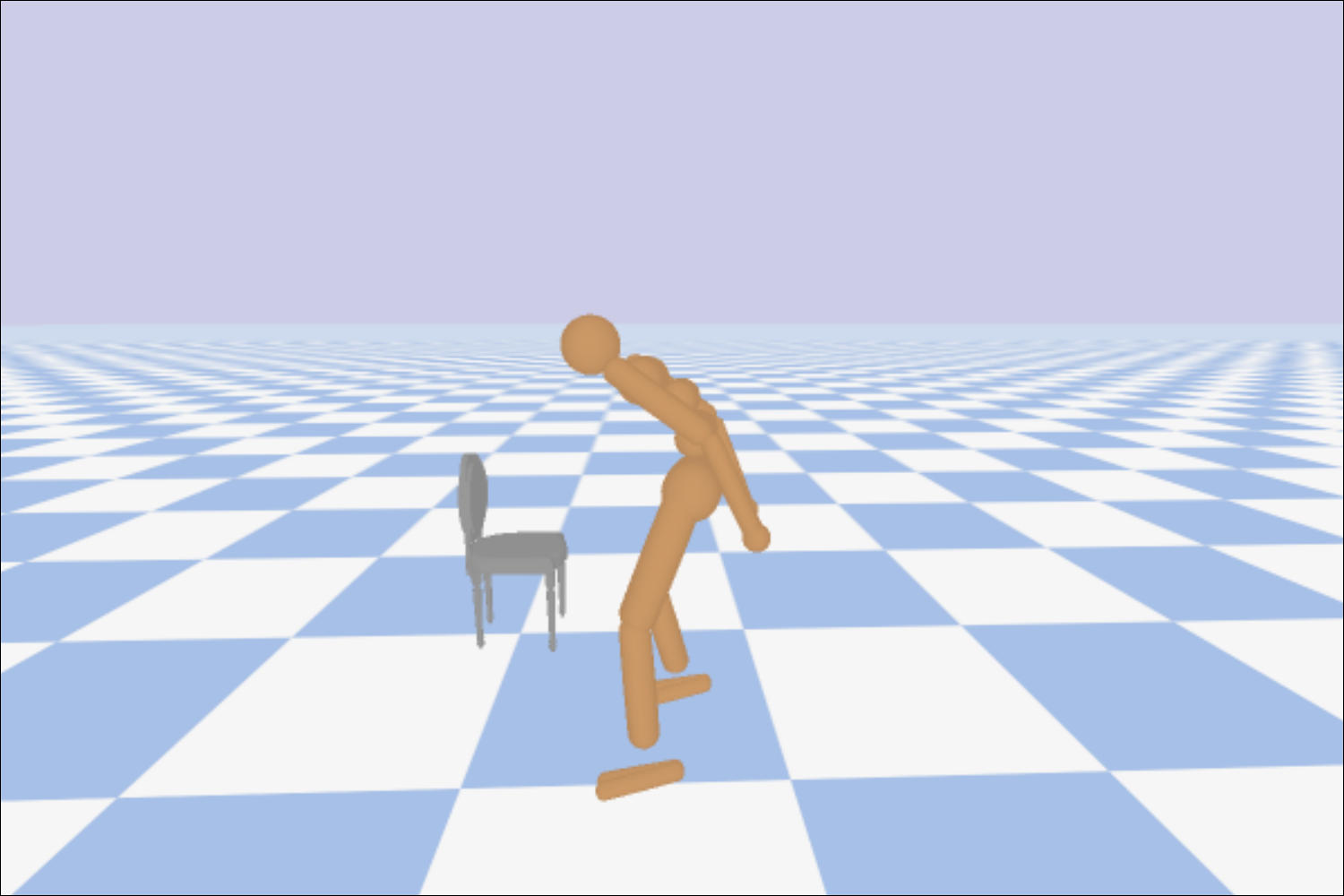} \end{minipage}
 \begin{minipage}{0.12\textwidth} \centering \includegraphics[width=1.00\textwidth,trim={80 40 80 40},clip]{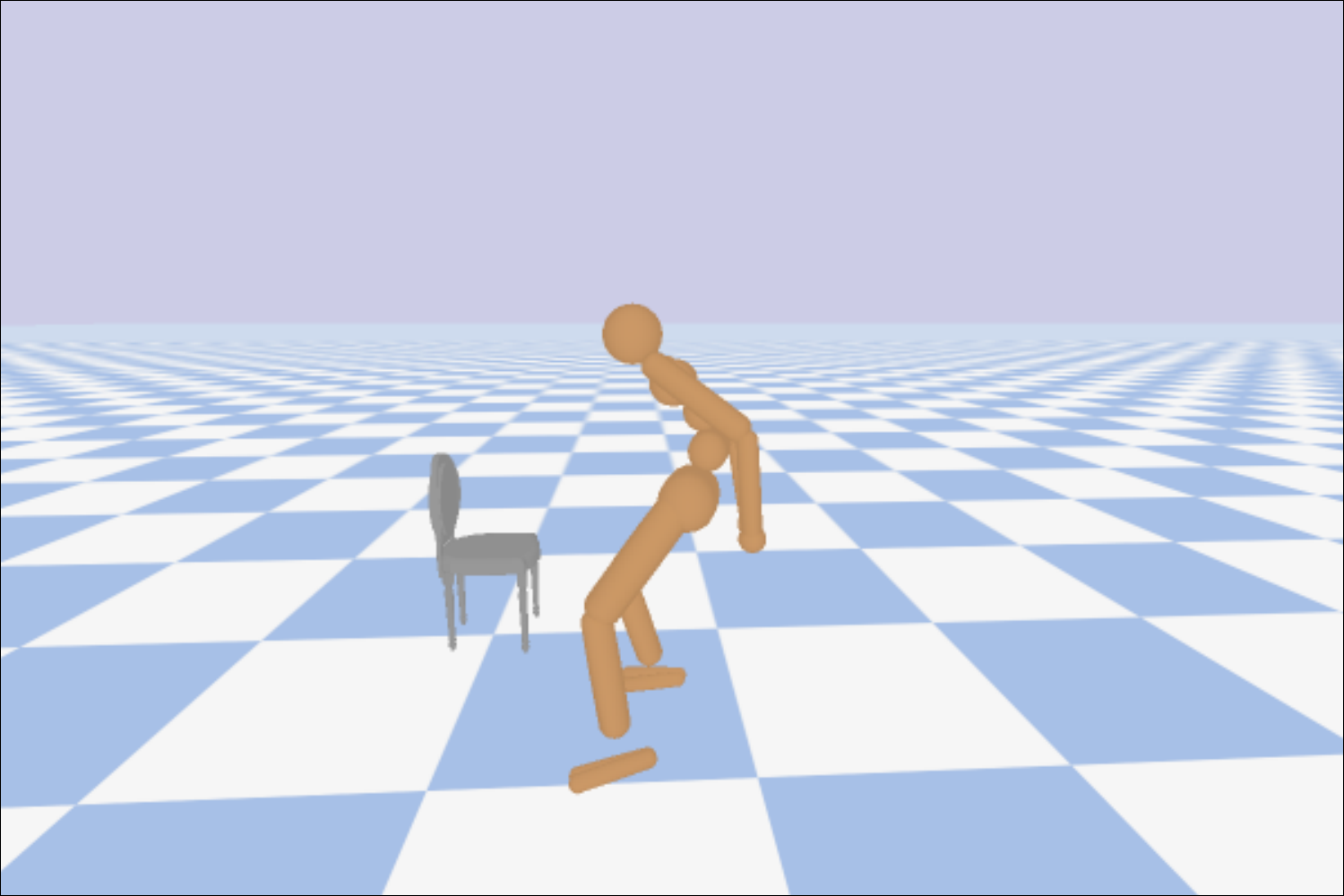} \end{minipage}
 \begin{minipage}{0.12\textwidth} \centering \includegraphics[width=1.00\textwidth,trim={80 40 80 40},clip]{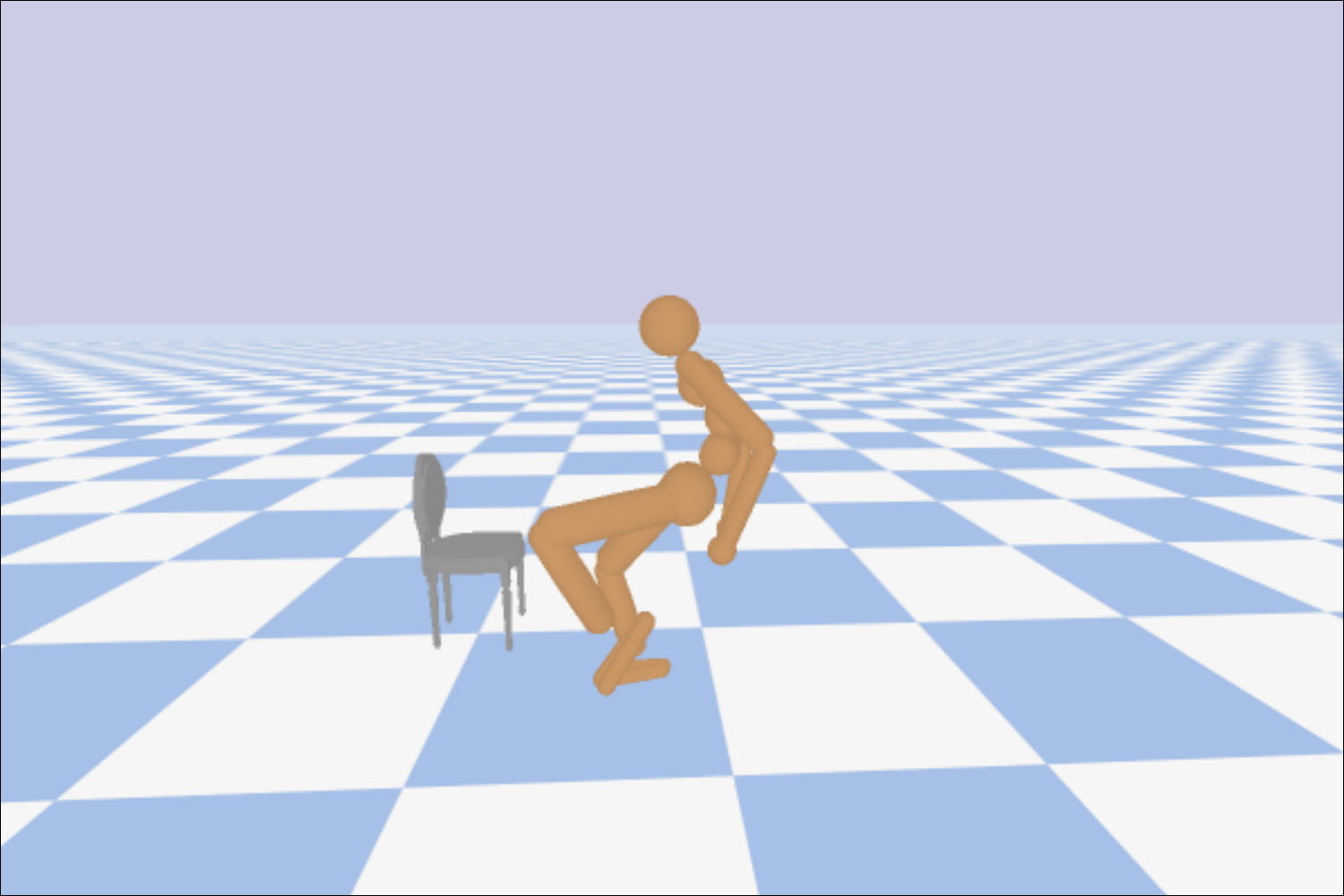} \end{minipage}
 \begin{minipage}{0.12\textwidth} \centering \includegraphics[width=1.00\textwidth,trim={80 40 80 40},clip]{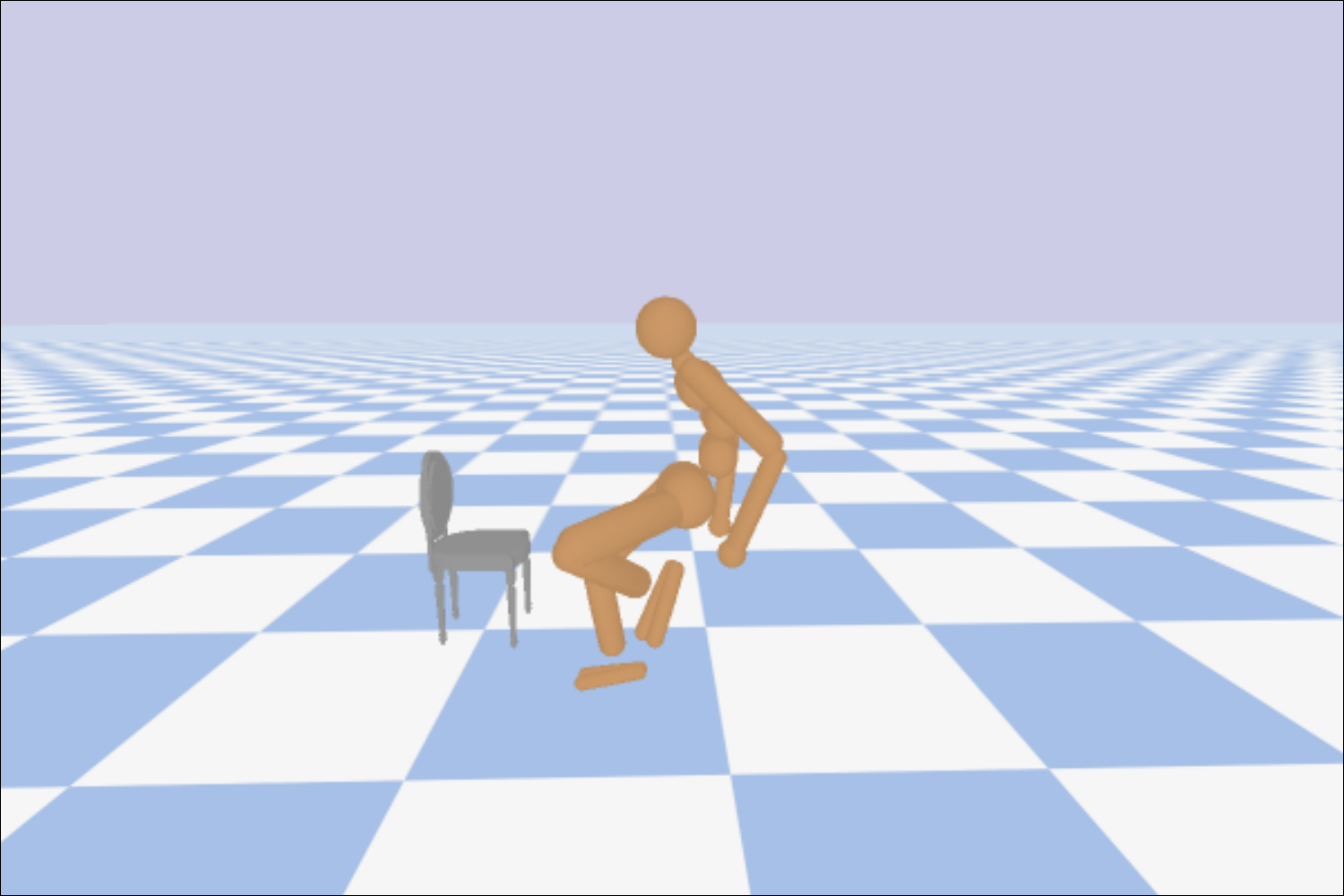} \end{minipage}
 \begin{minipage}{0.12\textwidth} \centering \includegraphics[width=1.00\textwidth,trim={80 40 80 40},clip]{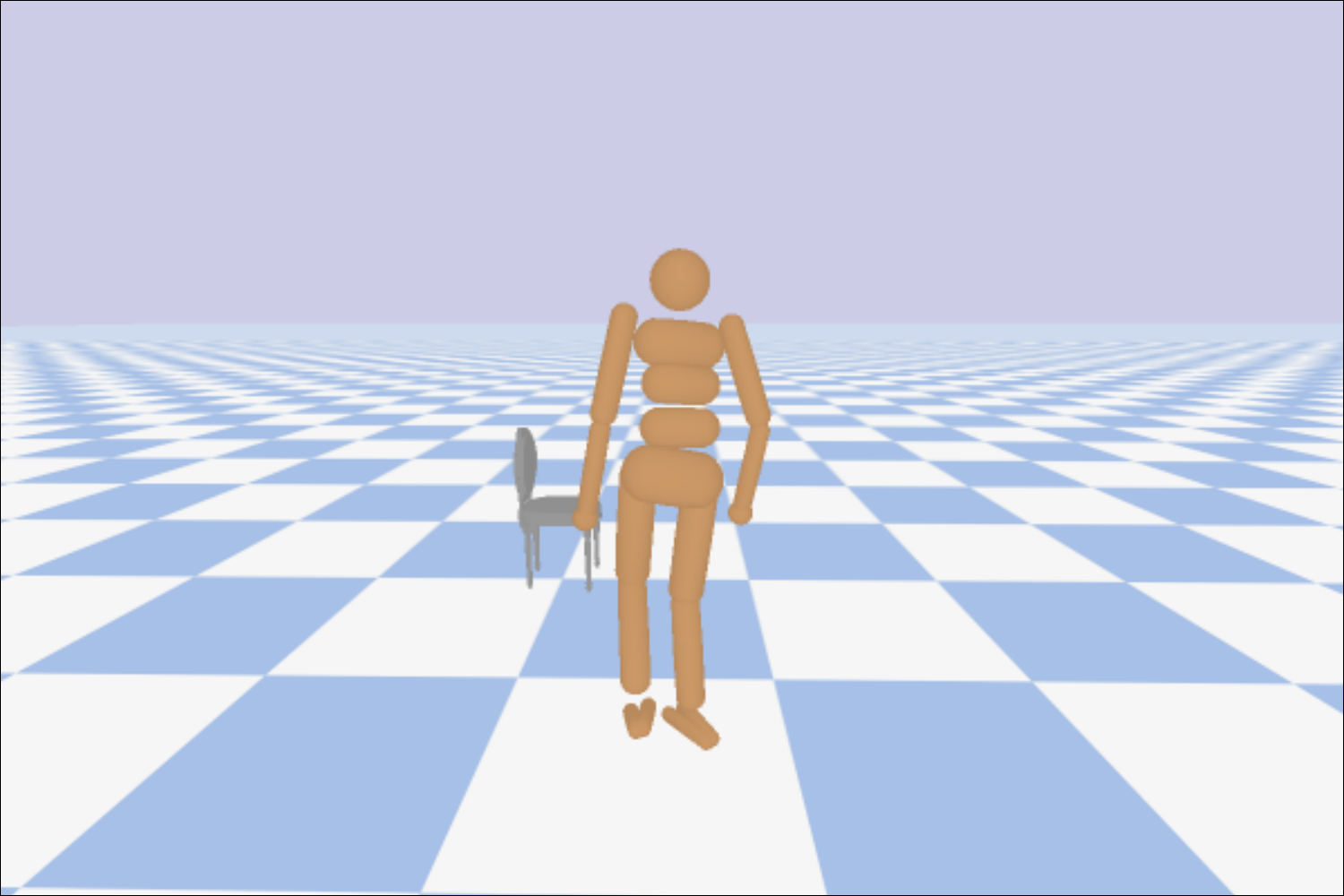} \end{minipage}
 \begin{minipage}{0.12\textwidth} \centering \includegraphics[width=1.00\textwidth,trim={80 40 80 40},clip]{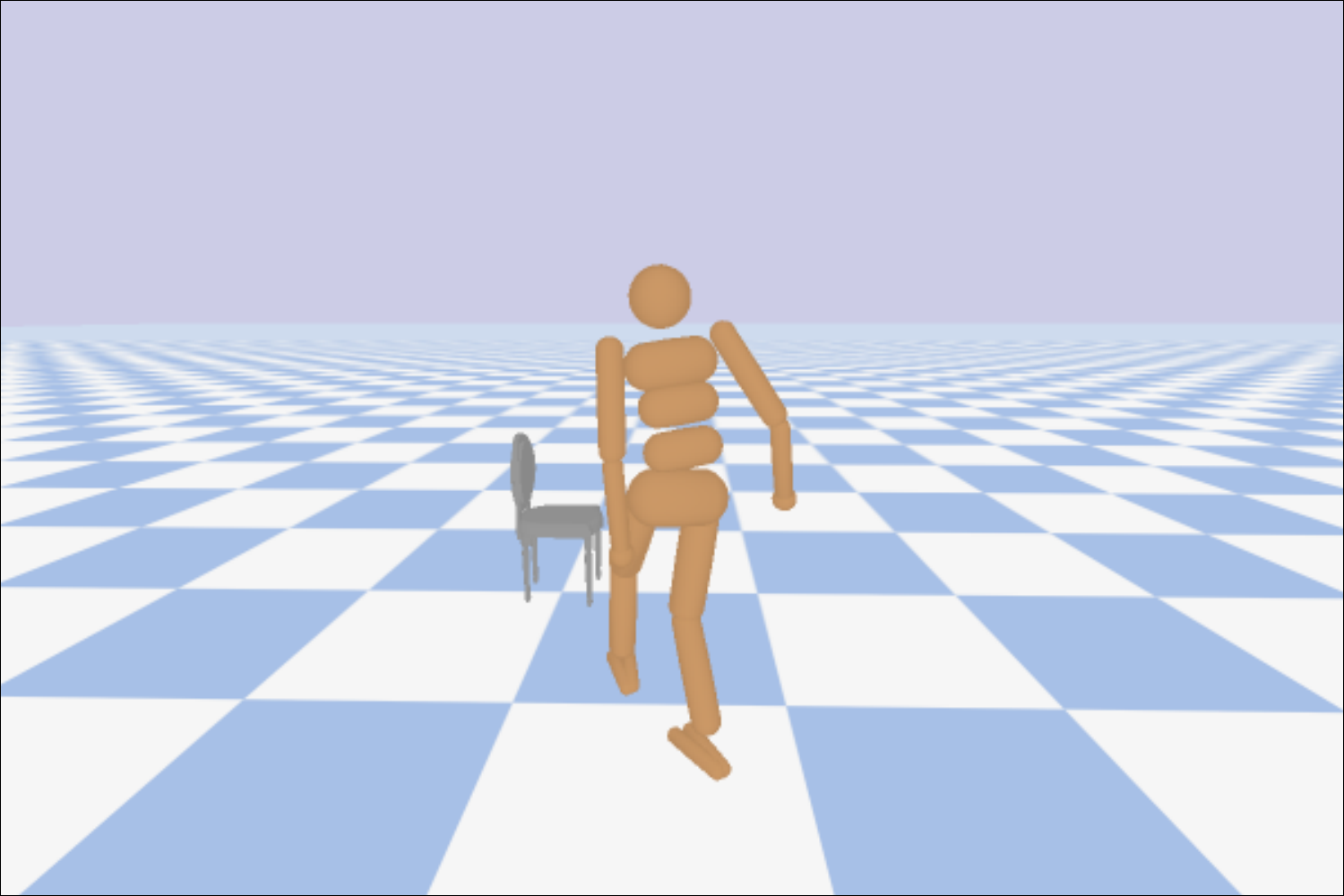} \end{minipage}
 \begin{minipage}{0.12\textwidth} \centering \includegraphics[width=1.00\textwidth,trim={80 40 80 40},clip]{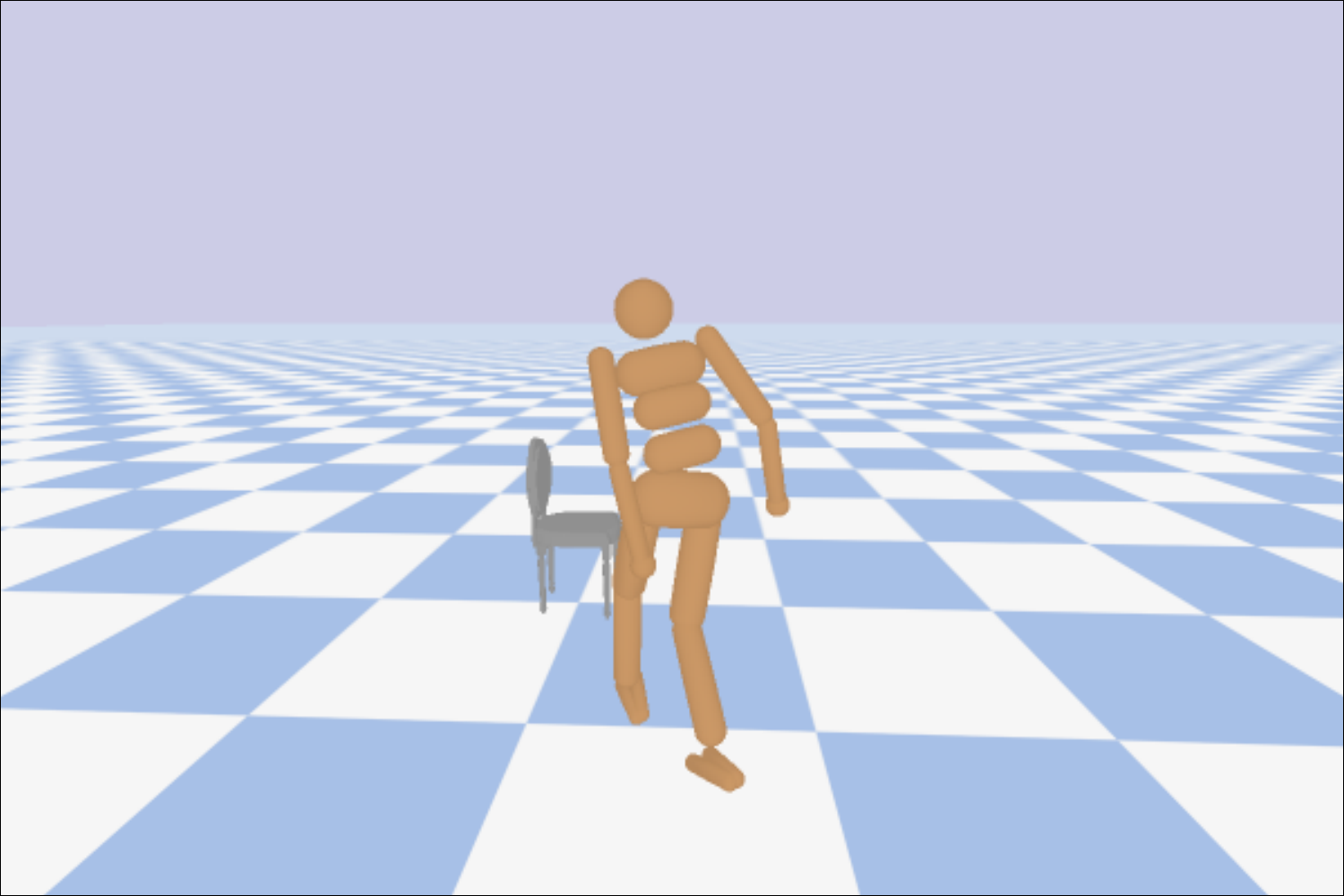} \end{minipage}
 \begin{minipage}{0.12\textwidth} \centering \includegraphics[width=1.00\textwidth,trim={80 40 80 40},clip]{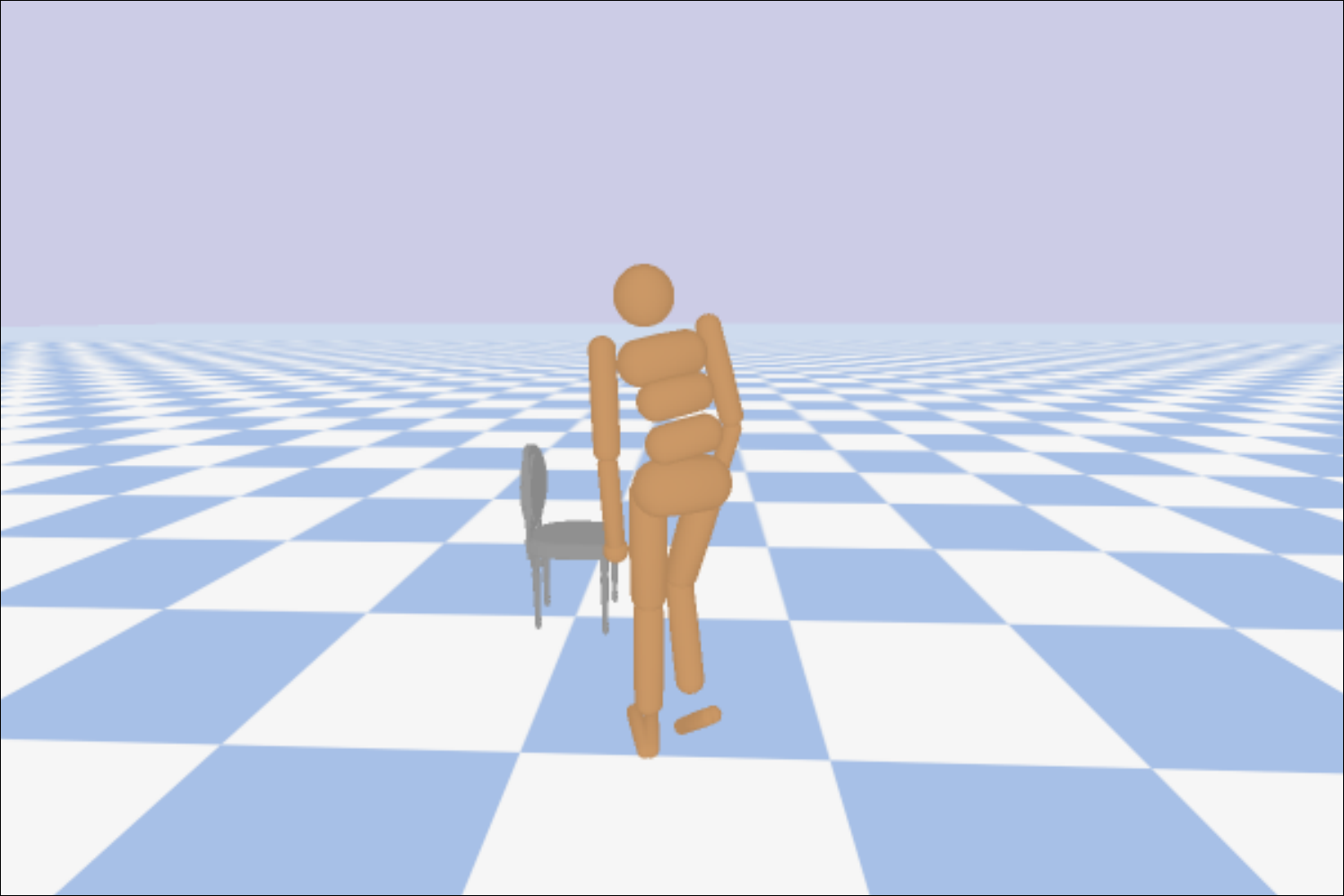} \end{minipage}
 \begin{minipage}{0.12\textwidth} \centering \includegraphics[width=1.00\textwidth,trim={80 40 80 40},clip]{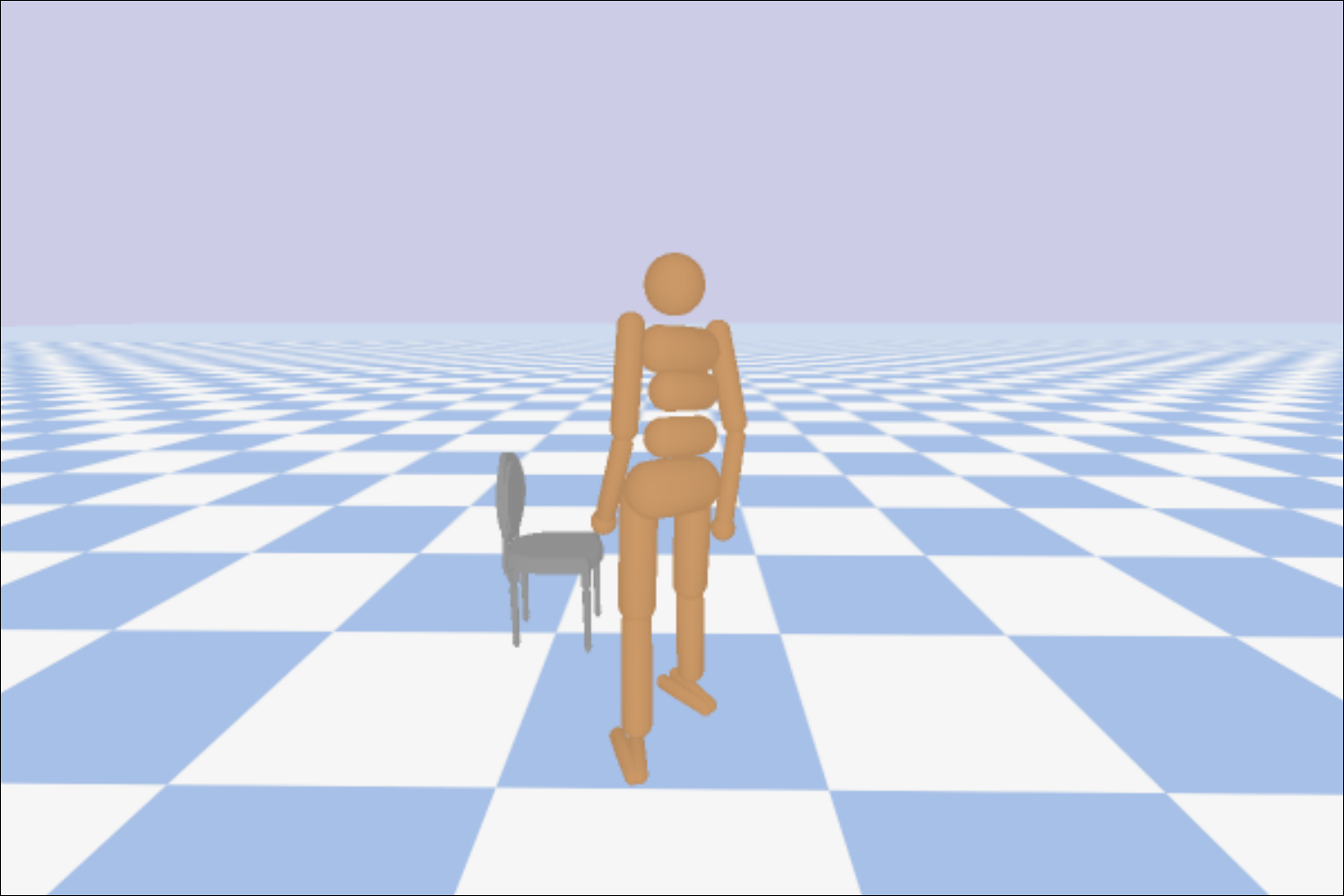} \end{minipage}
 \begin{minipage}{0.12\textwidth} \centering \includegraphics[width=1.00\textwidth,trim={80 40 80 40},clip]{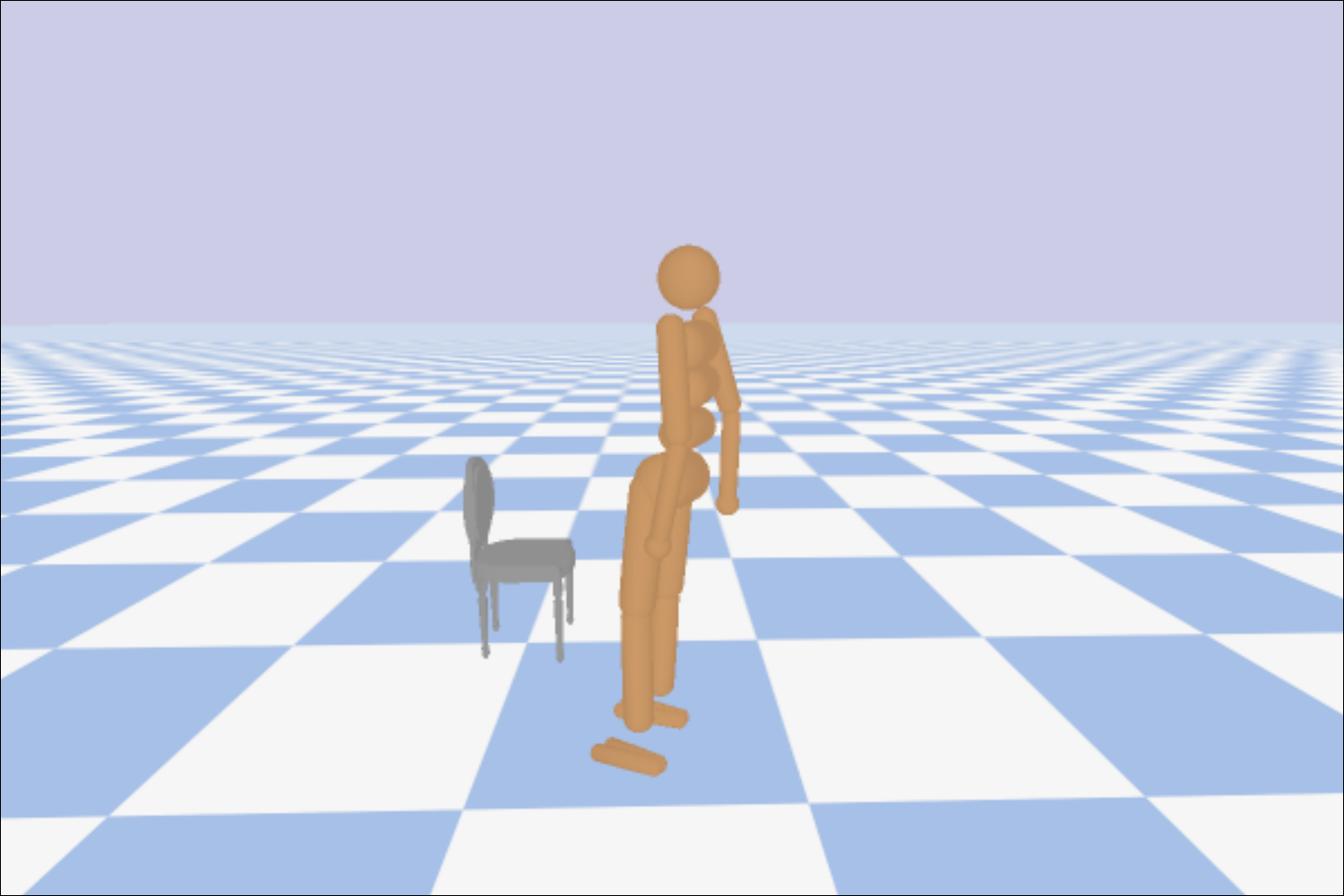} \end{minipage}
 \begin{minipage}{0.12\textwidth} \centering \includegraphics[width=1.00\textwidth,trim={80 40 80 40},clip]{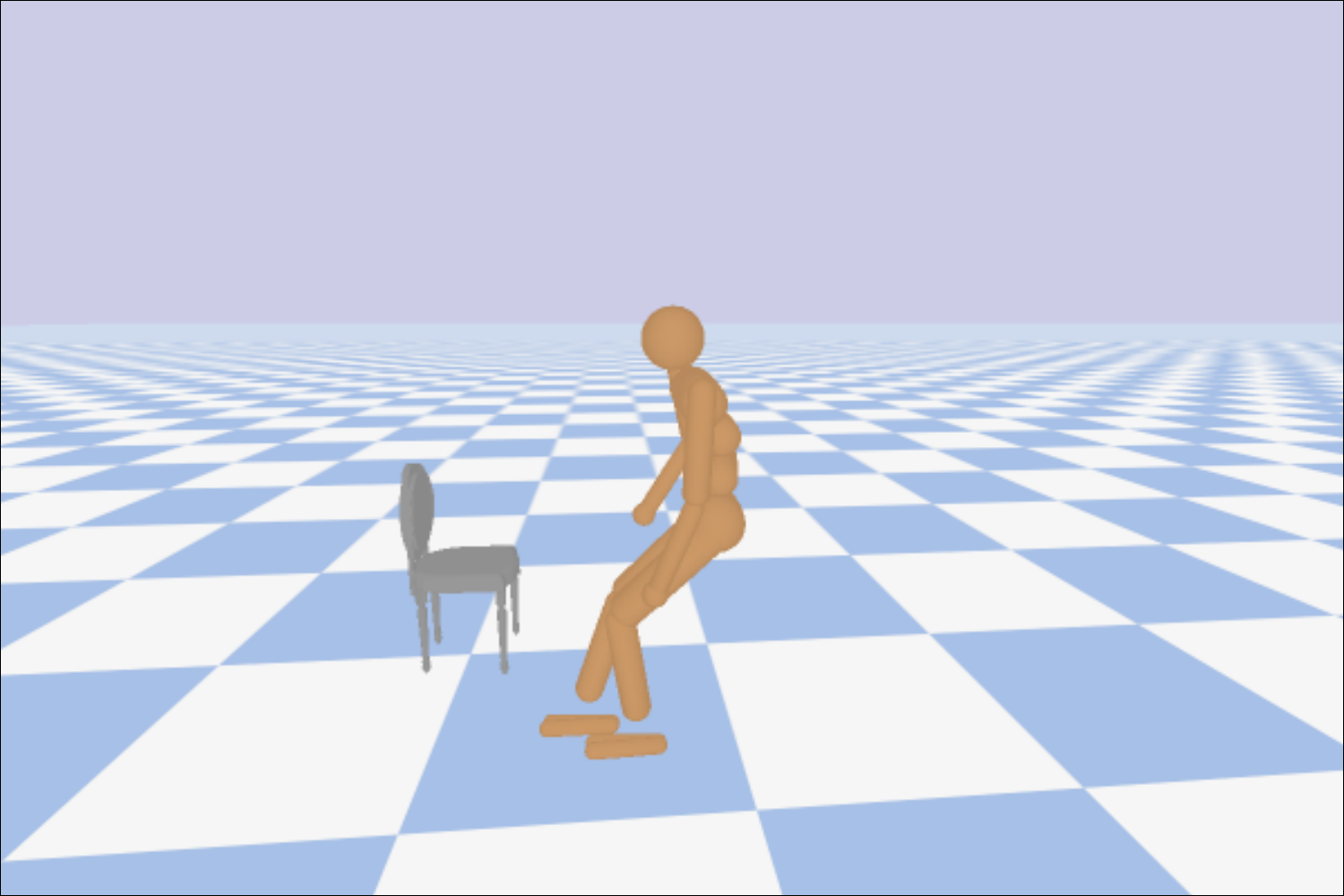} \end{minipage}
 \begin{minipage}{0.12\textwidth} \centering \includegraphics[width=1.00\textwidth,trim={80 40 80 40},clip]{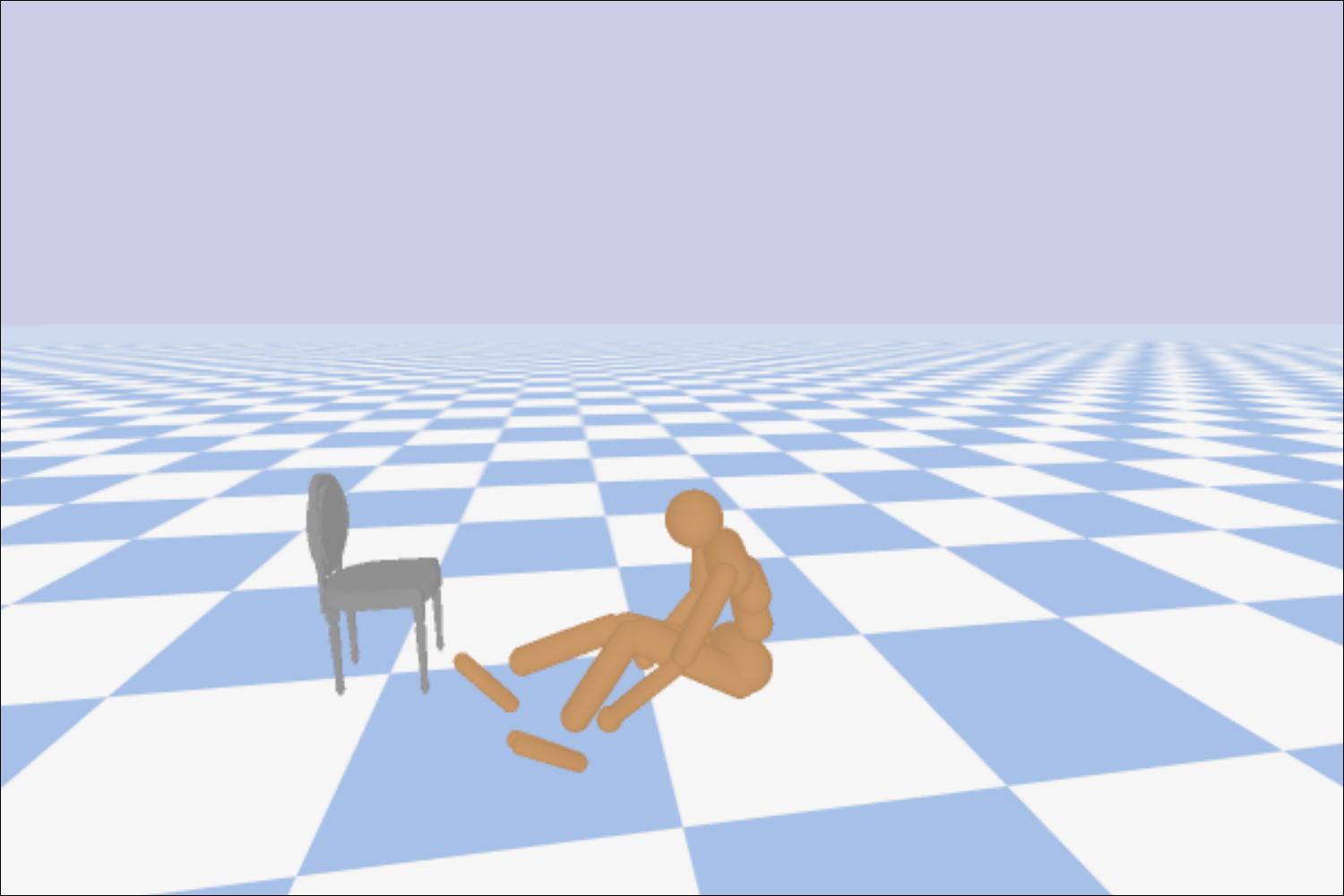} \end{minipage}
 \\ \vspace{2mm}
 \begin{minipage}{0.12\textwidth} \centering \includegraphics[width=1.00\textwidth,trim={80 40 80 40},clip]{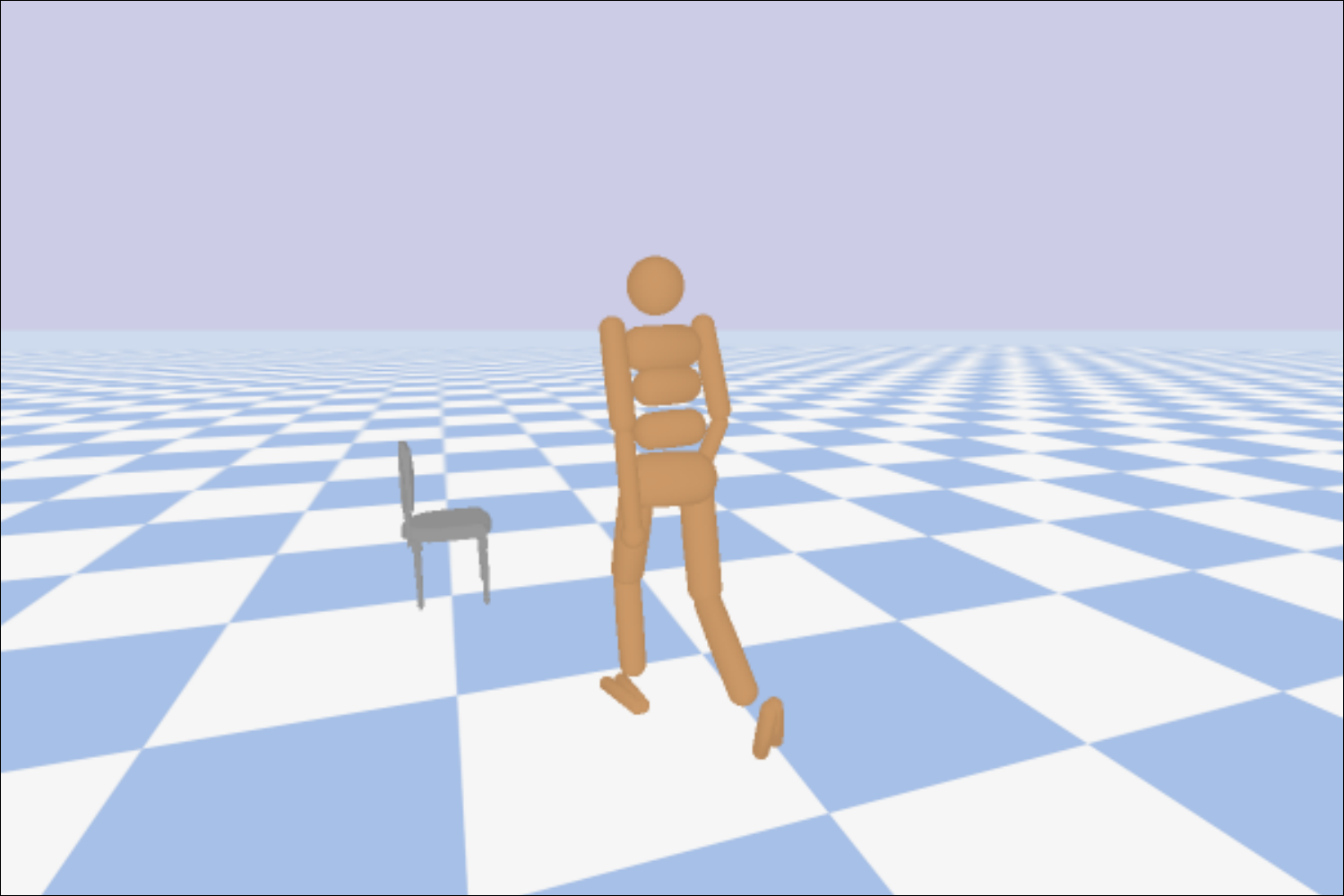} \end{minipage}
 \begin{minipage}{0.12\textwidth} \centering \includegraphics[width=1.00\textwidth,trim={80 40 80 40},clip]{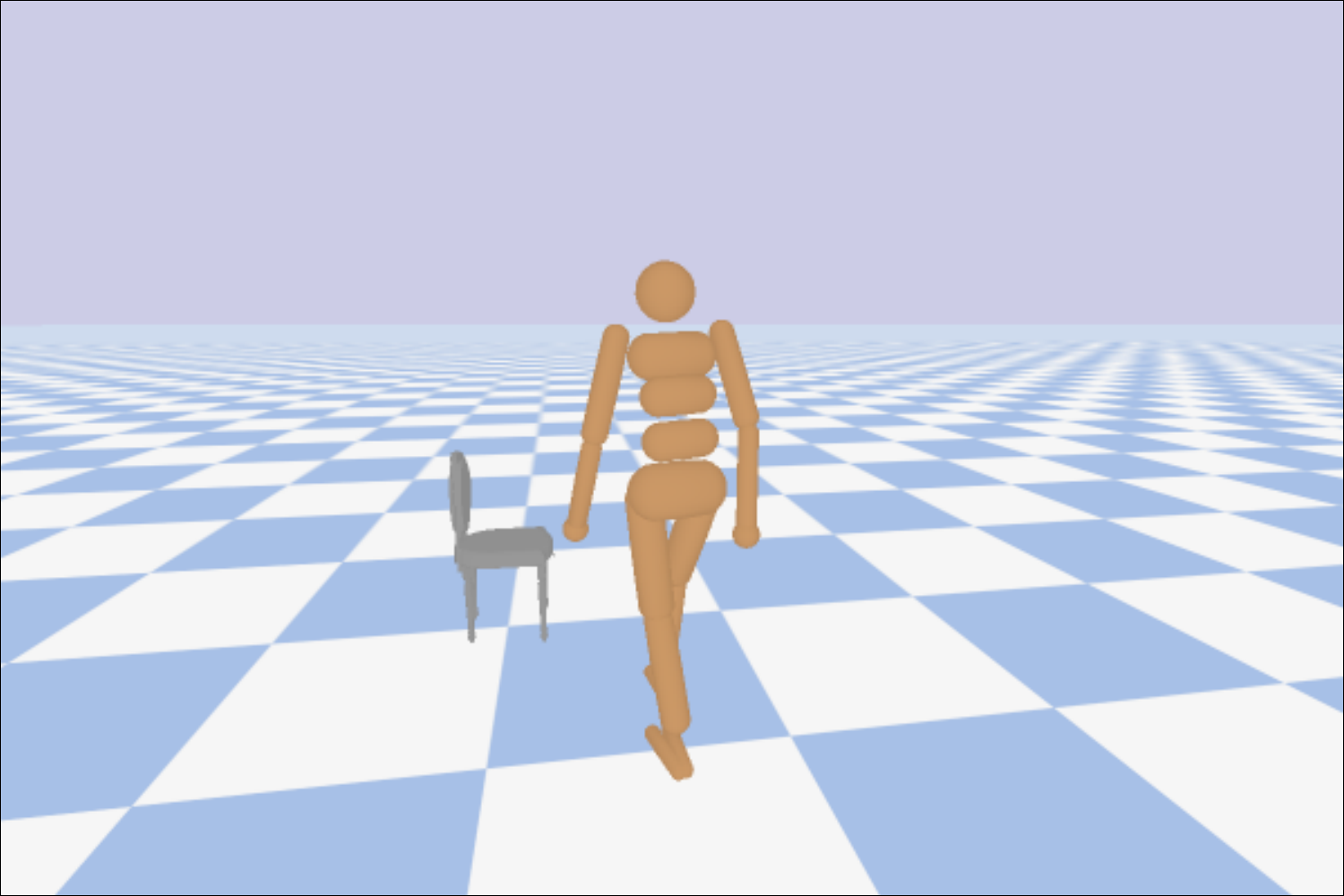} \end{minipage}
 \begin{minipage}{0.12\textwidth} \centering \includegraphics[width=1.00\textwidth,trim={80 40 80 40},clip]{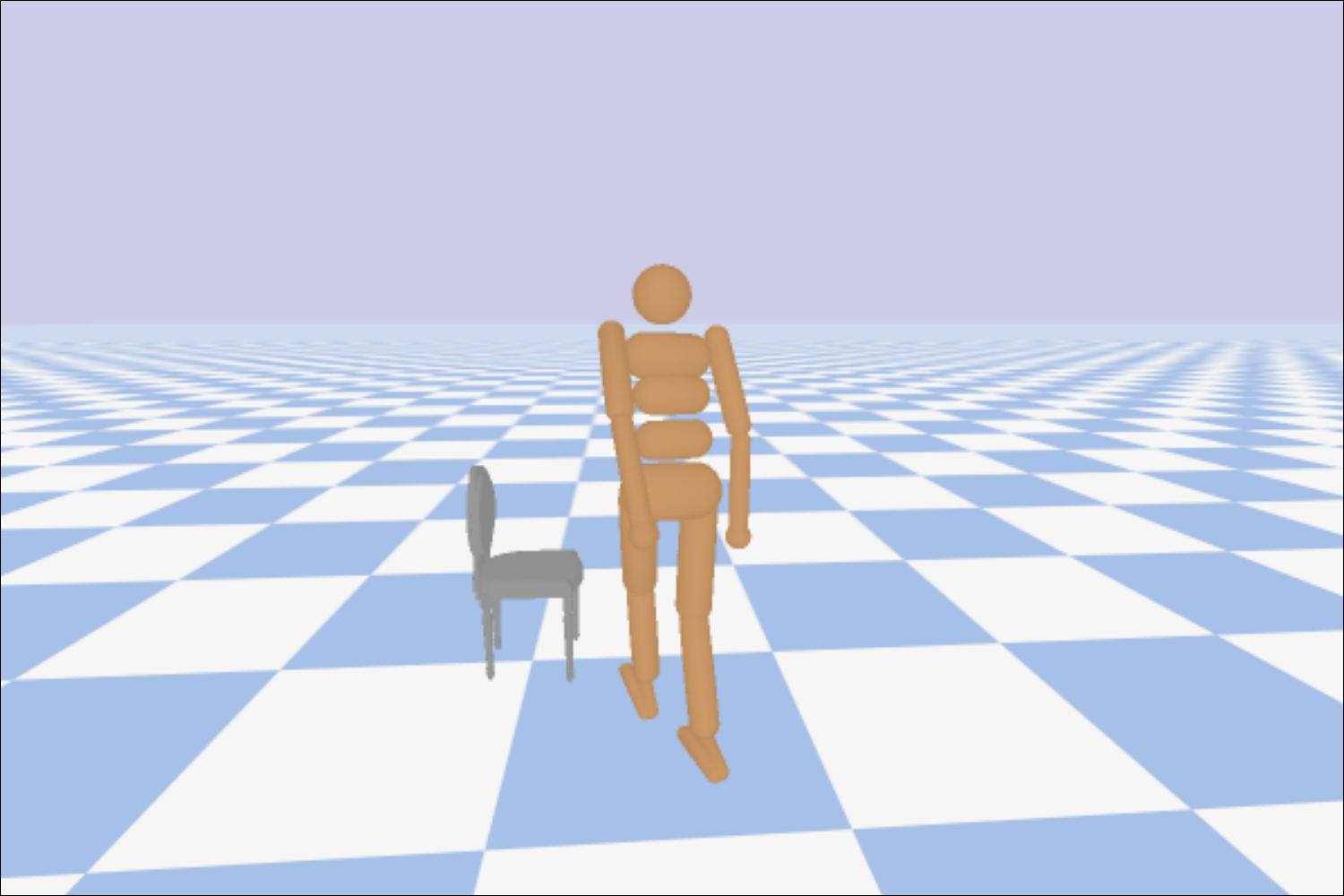} \end{minipage}
 \begin{minipage}{0.12\textwidth} \centering \includegraphics[width=1.00\textwidth,trim={80 40 80 40},clip]{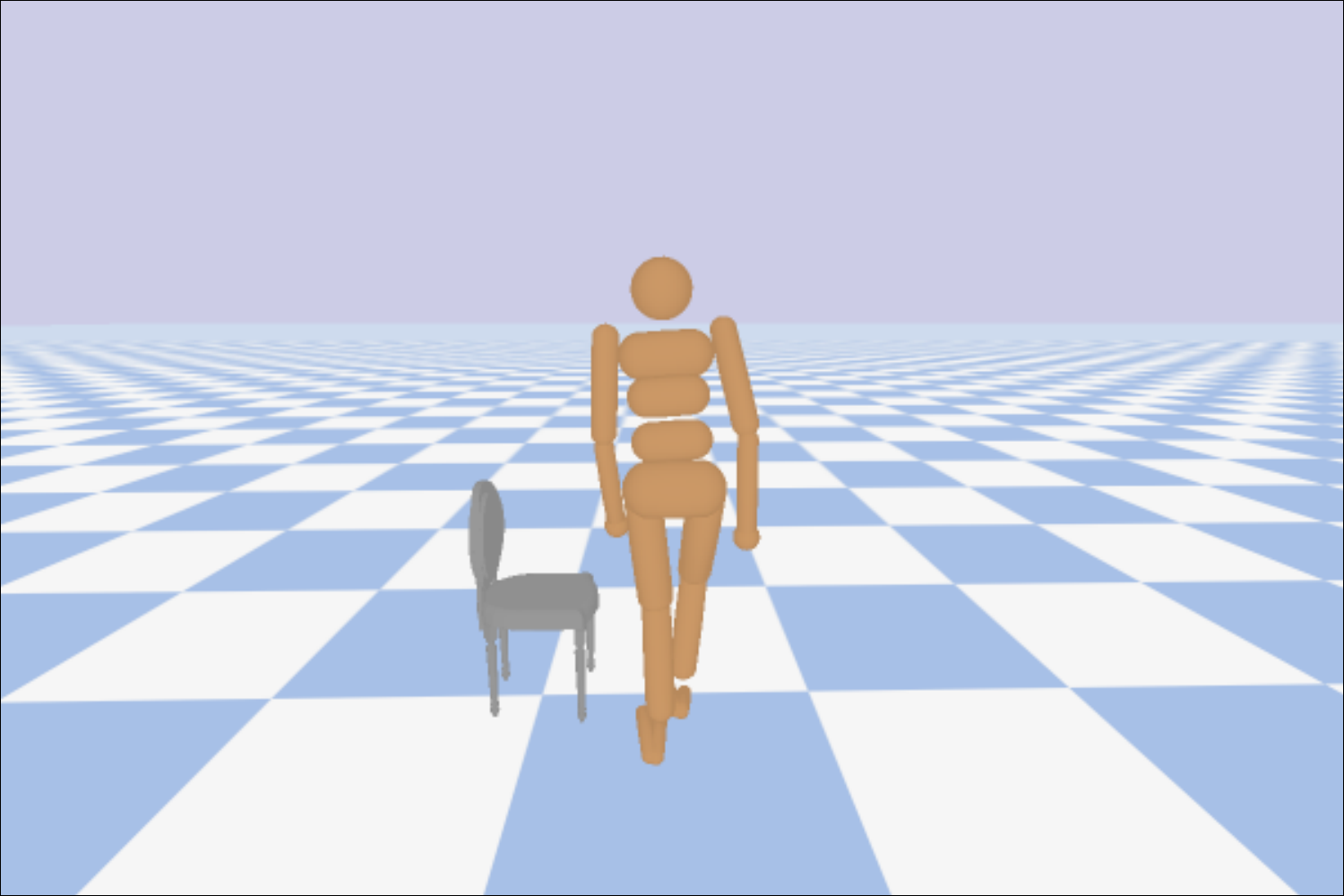} \end{minipage}
 \begin{minipage}{0.12\textwidth} \centering \includegraphics[width=1.00\textwidth,trim={80 40 80 40},clip]{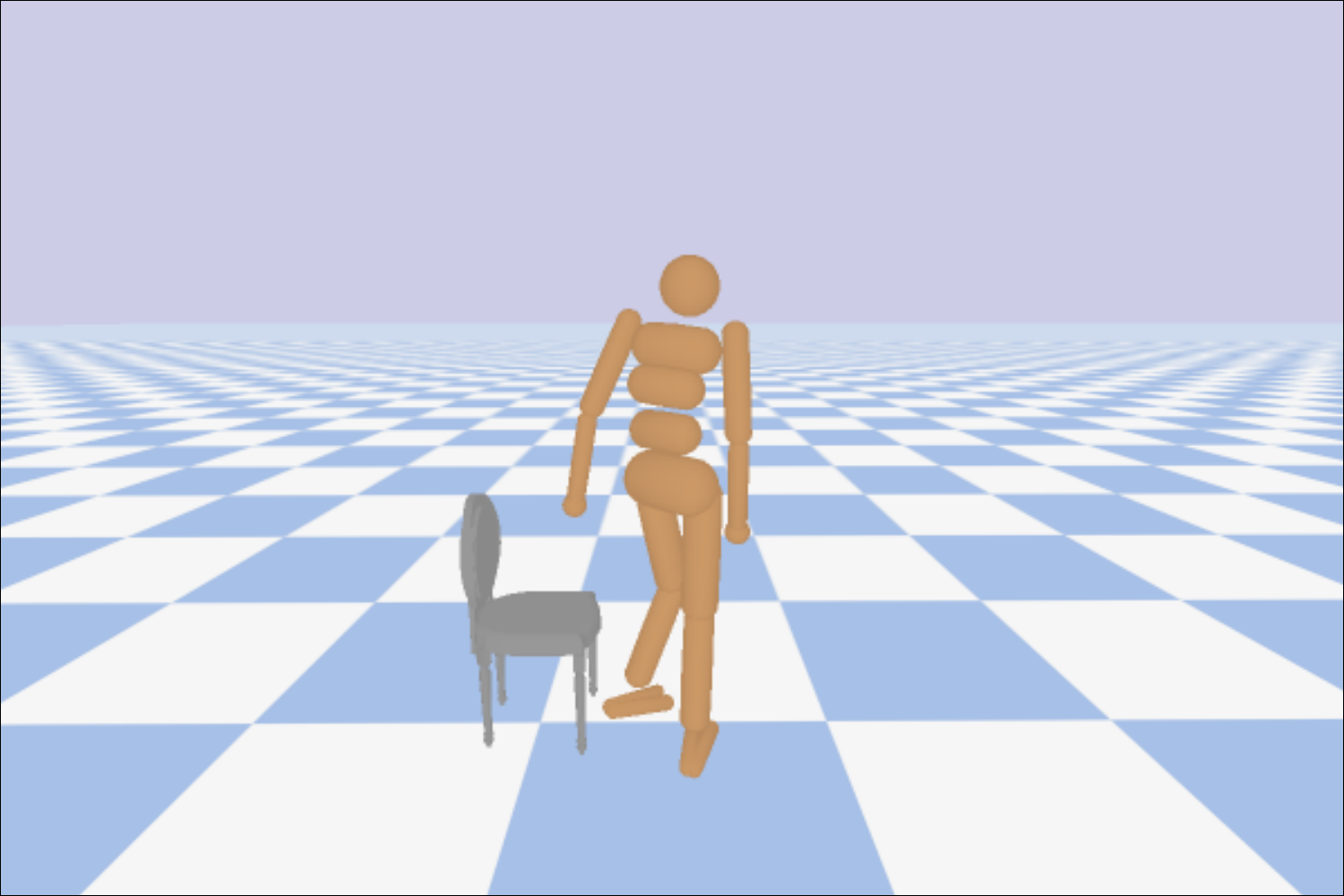} \end{minipage}
 \begin{minipage}{0.12\textwidth} \centering \includegraphics[width=1.00\textwidth,trim={80 40 80 40},clip]{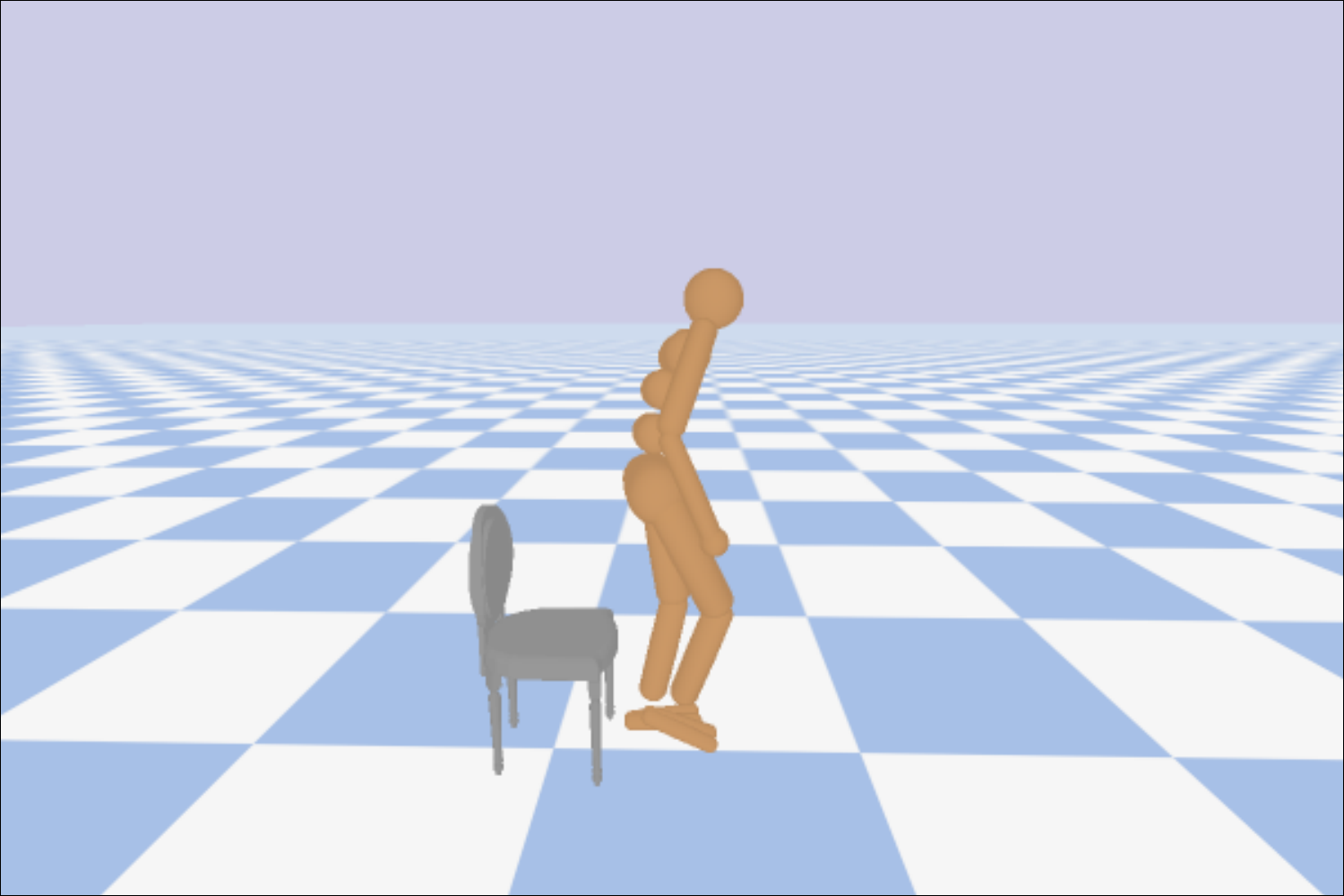} \end{minipage}
 \begin{minipage}{0.12\textwidth} \centering \includegraphics[width=1.00\textwidth,trim={80 40 80 40},clip]{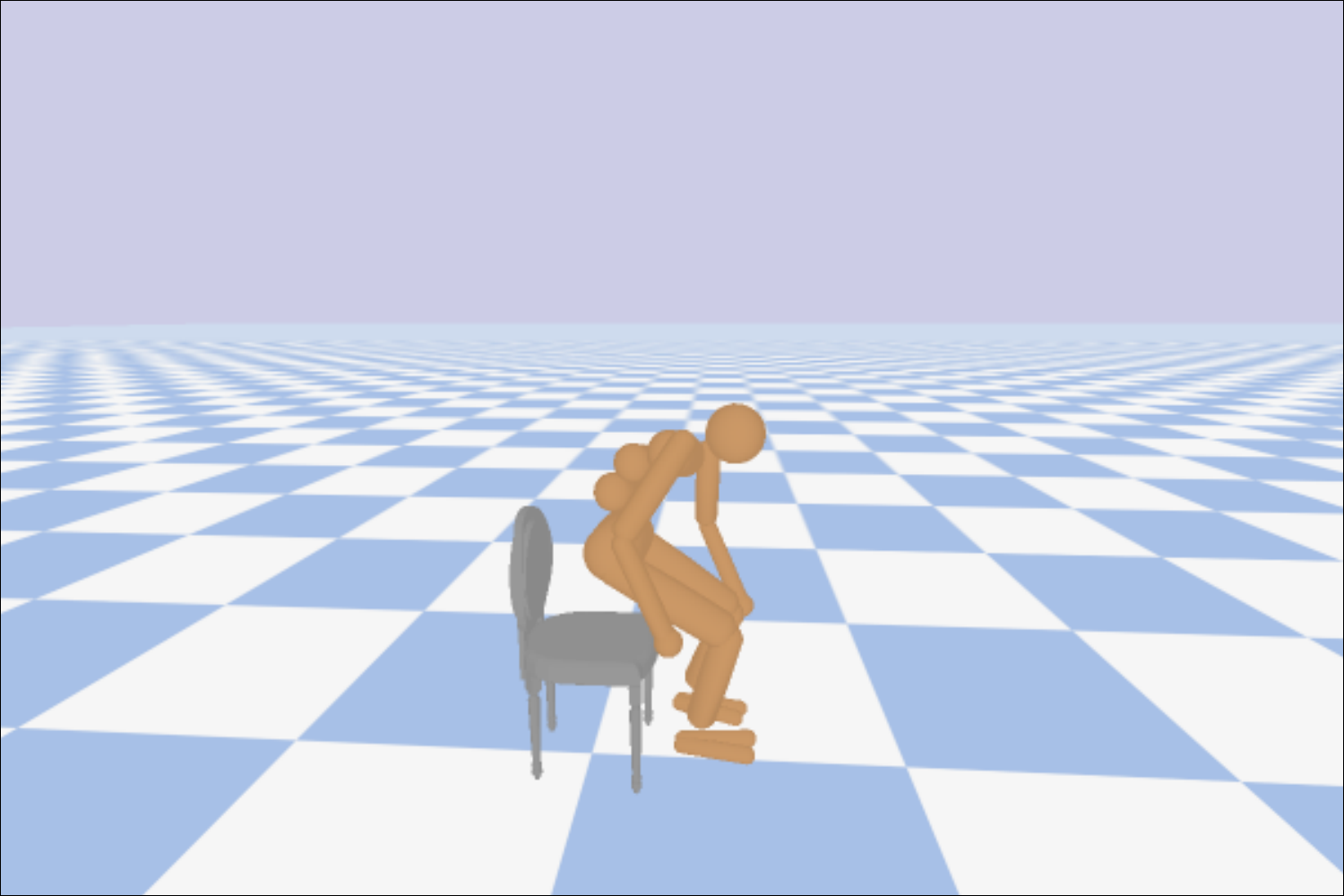} \end{minipage}
 \begin{minipage}{0.12\textwidth} \centering \includegraphics[width=1.00\textwidth,trim={80 40 80 40},clip]{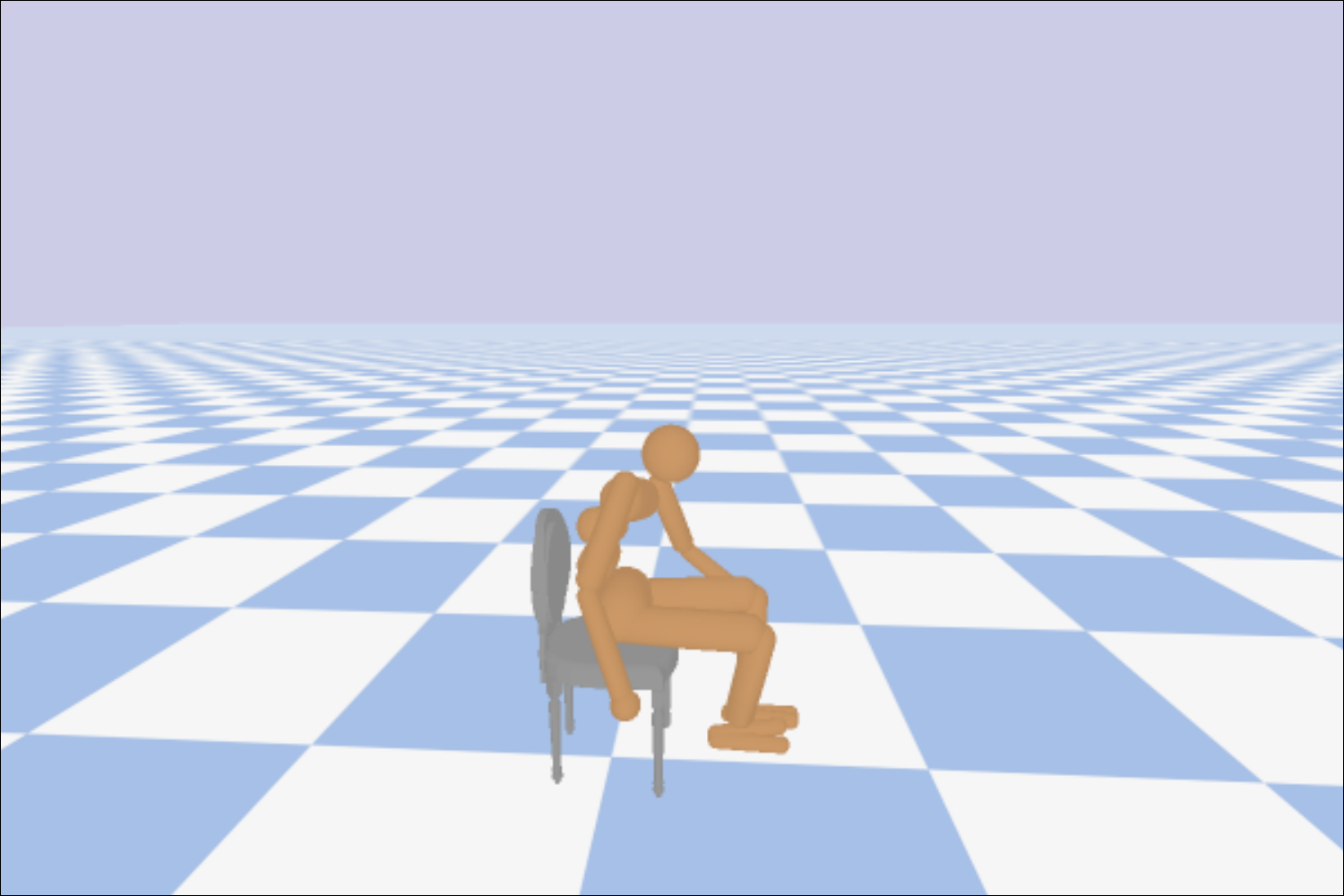} \end{minipage}
 \begin{minipage}{0.12\textwidth} \centering \includegraphics[width=1.00\textwidth,trim={80 40 80 40},clip]{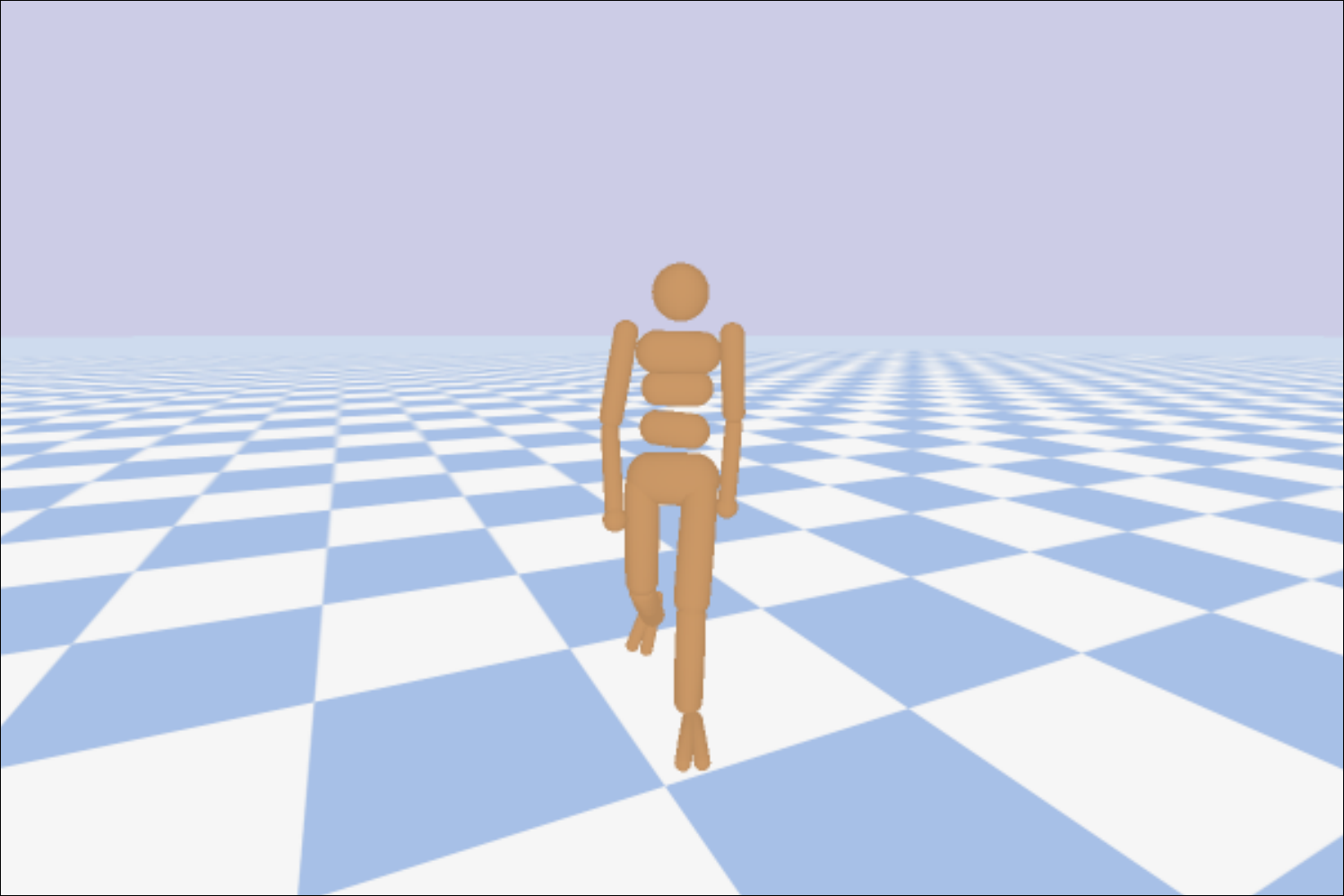} \end{minipage}
 \begin{minipage}{0.12\textwidth} \centering \includegraphics[width=1.00\textwidth,trim={80 40 80 40},clip]{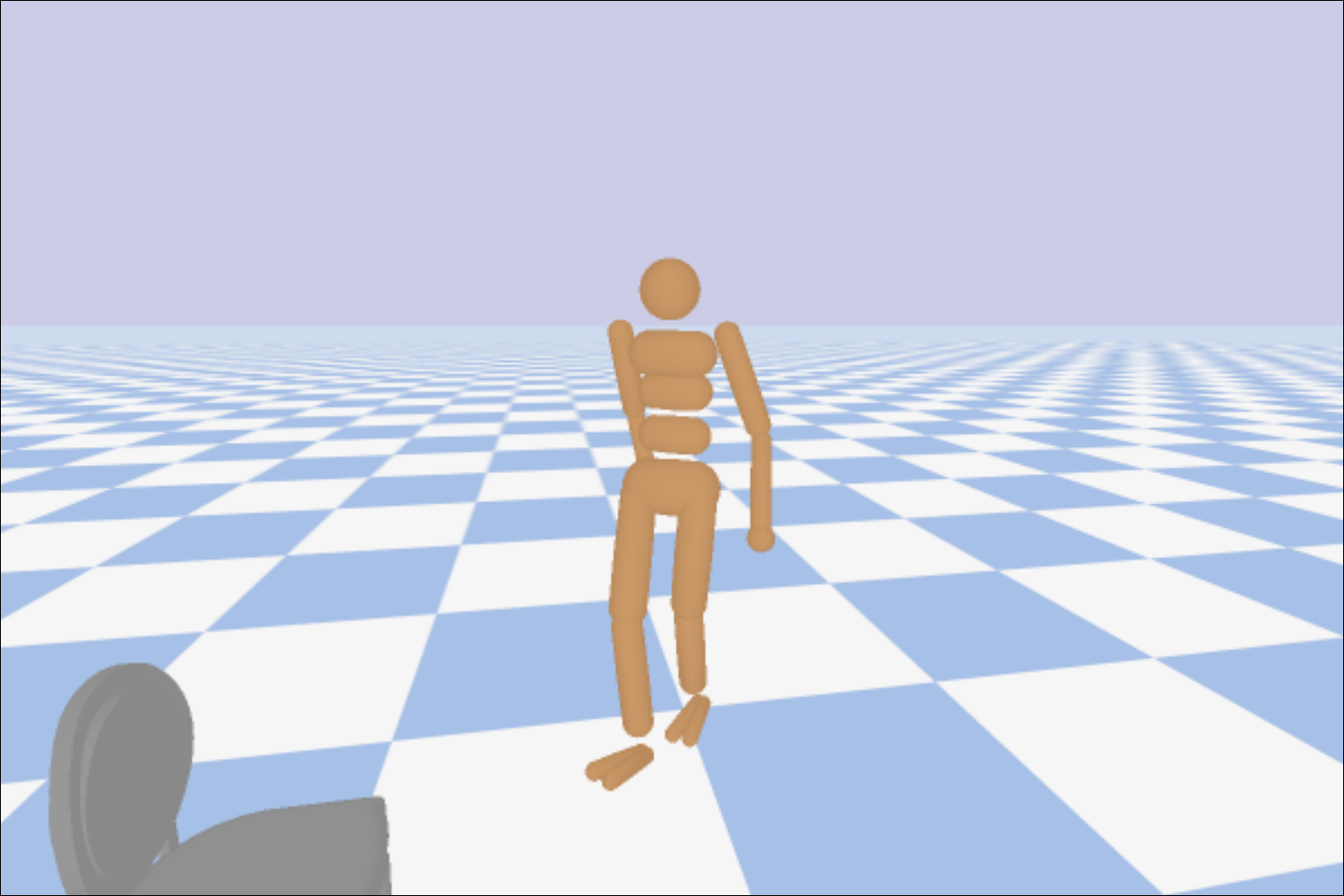} \end{minipage}
 \begin{minipage}{0.12\textwidth} \centering \includegraphics[width=1.00\textwidth,trim={80 40 80 40},clip]{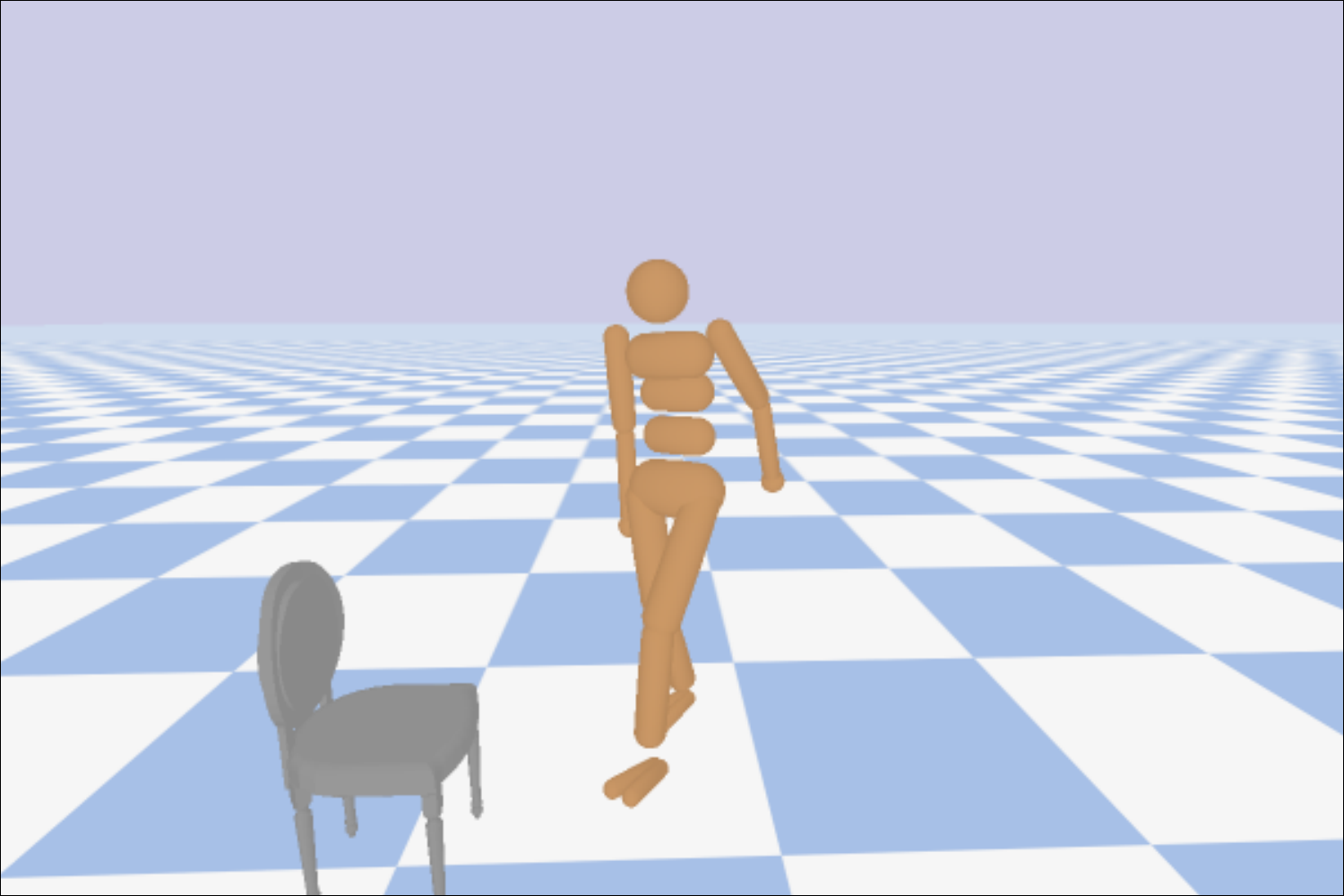} \end{minipage}
 \begin{minipage}{0.12\textwidth} \centering \includegraphics[width=1.00\textwidth,trim={80 40 80 40},clip]{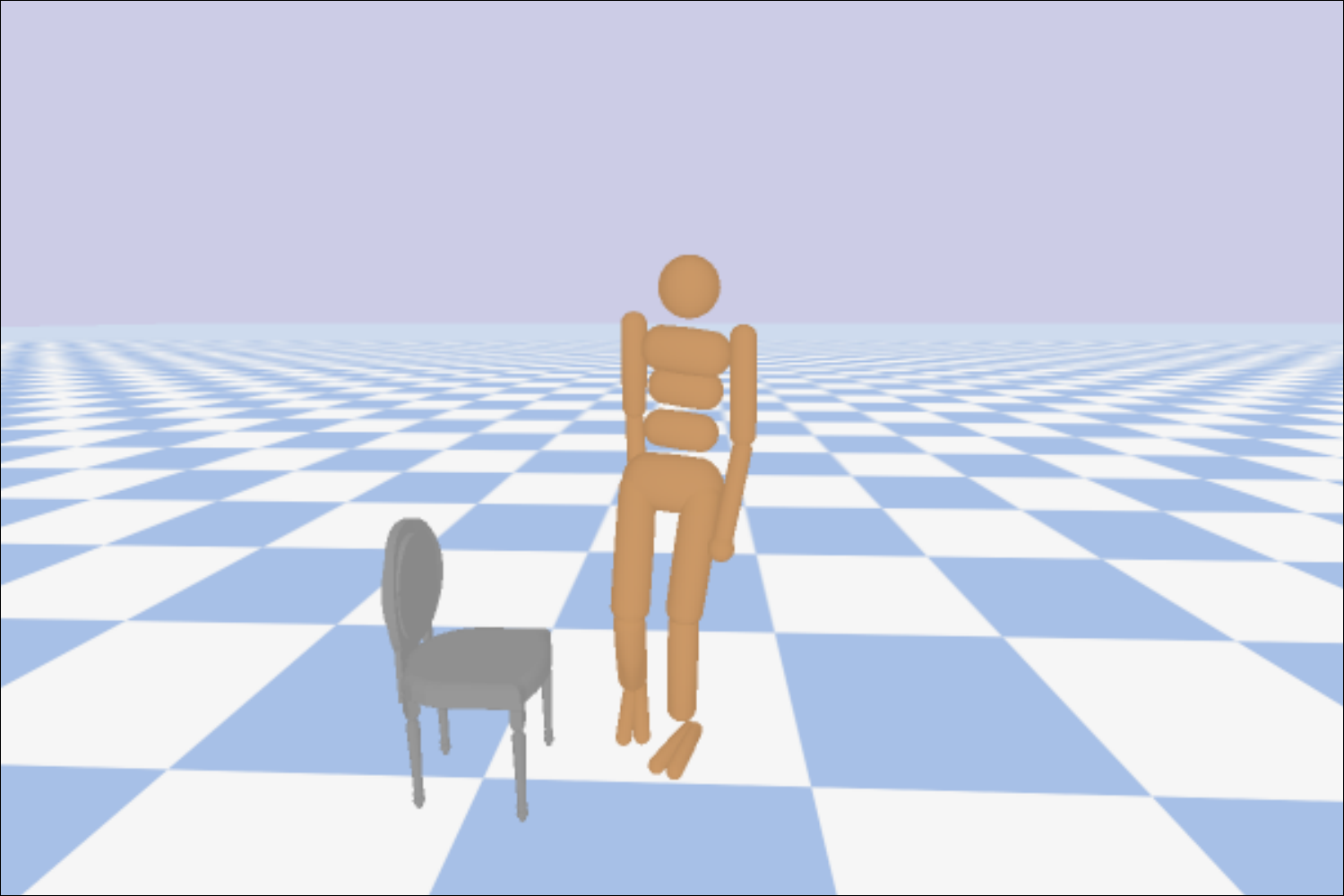} \end{minipage}
 \begin{minipage}{0.12\textwidth} \centering \includegraphics[width=1.00\textwidth,trim={80 40 80 40},clip]{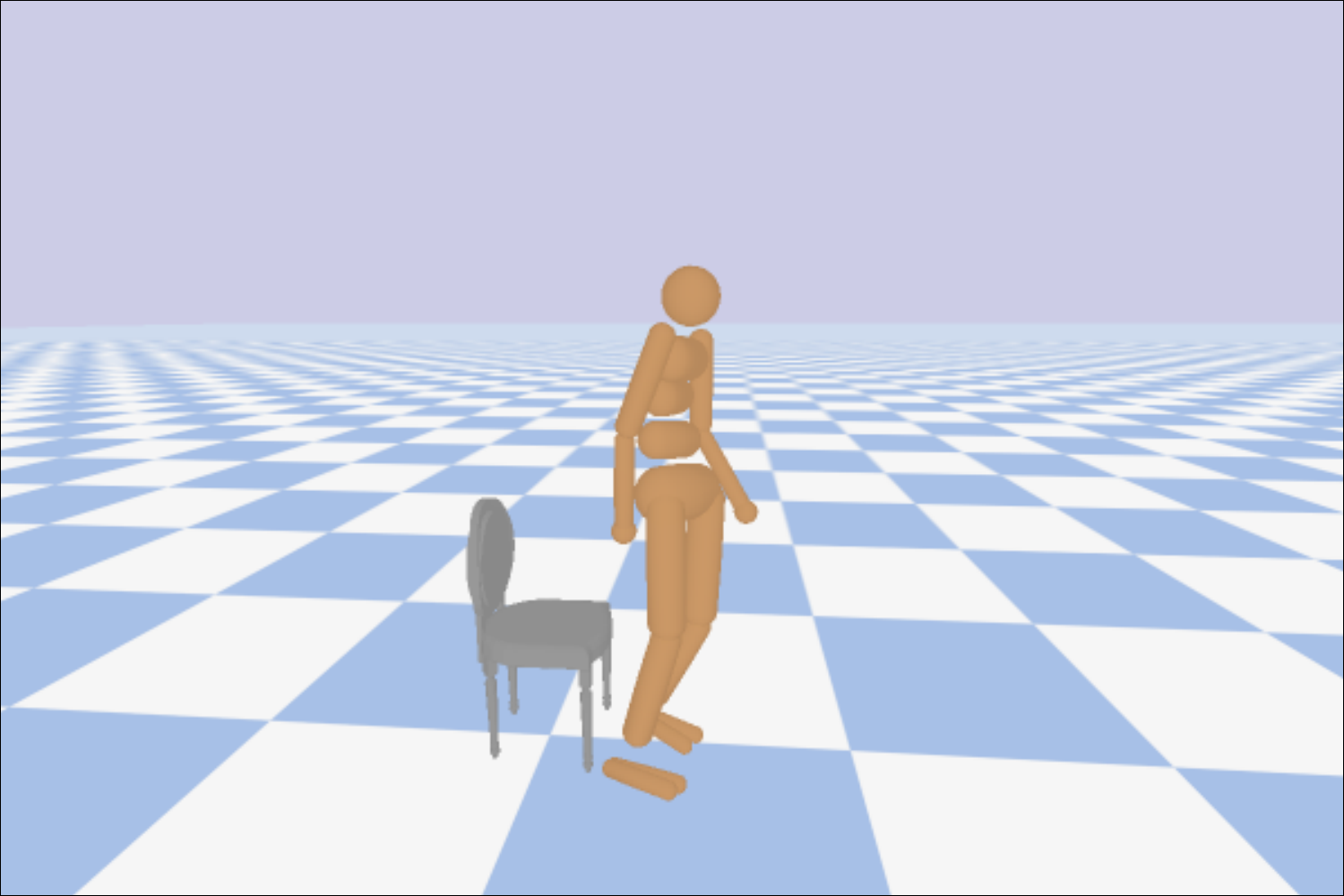} \end{minipage}
 \begin{minipage}{0.12\textwidth} \centering \includegraphics[width=1.00\textwidth,trim={80 40 80 40},clip]{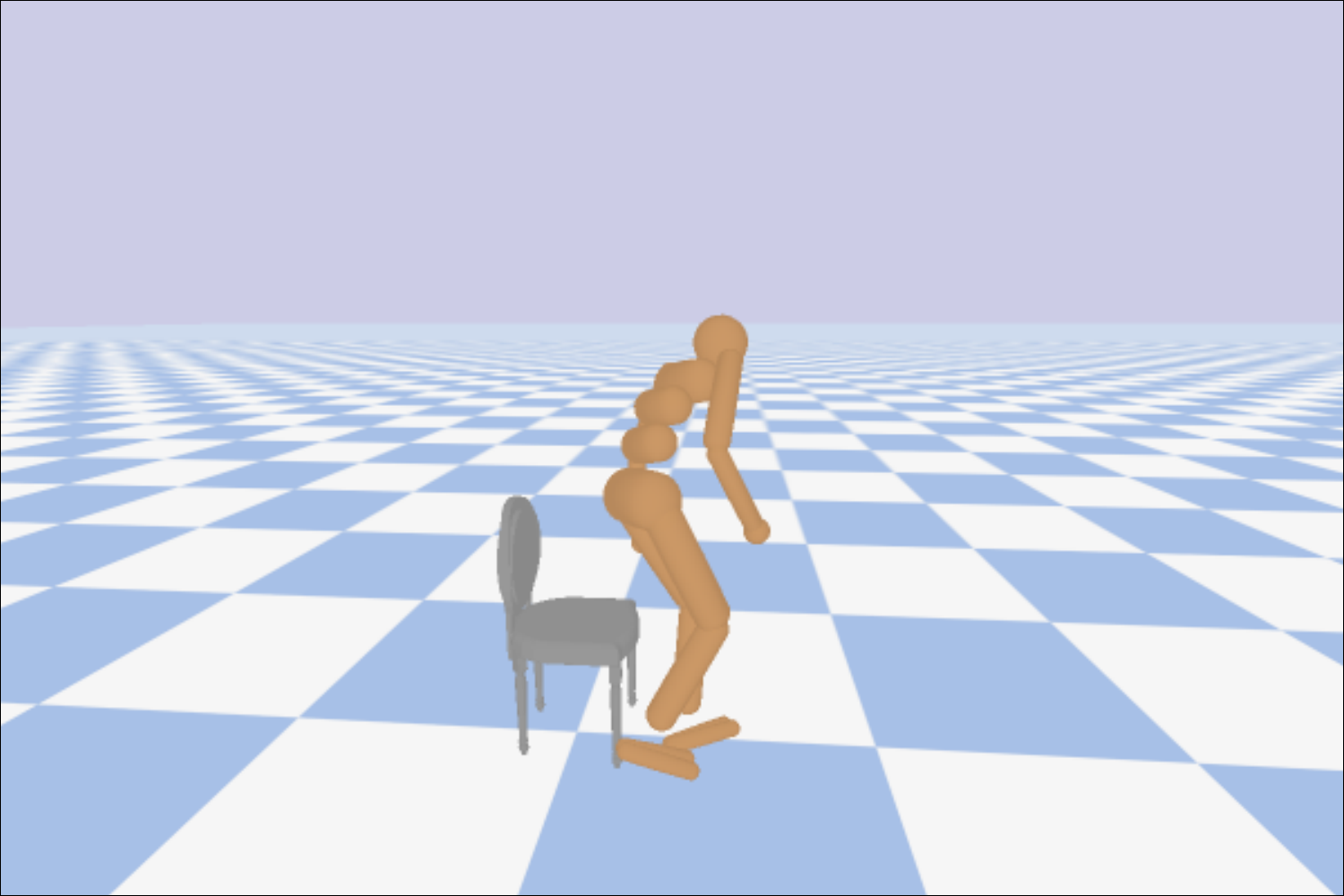} \end{minipage}
 \begin{minipage}{0.12\textwidth} \centering \includegraphics[width=1.00\textwidth,trim={80 40 80 40},clip]{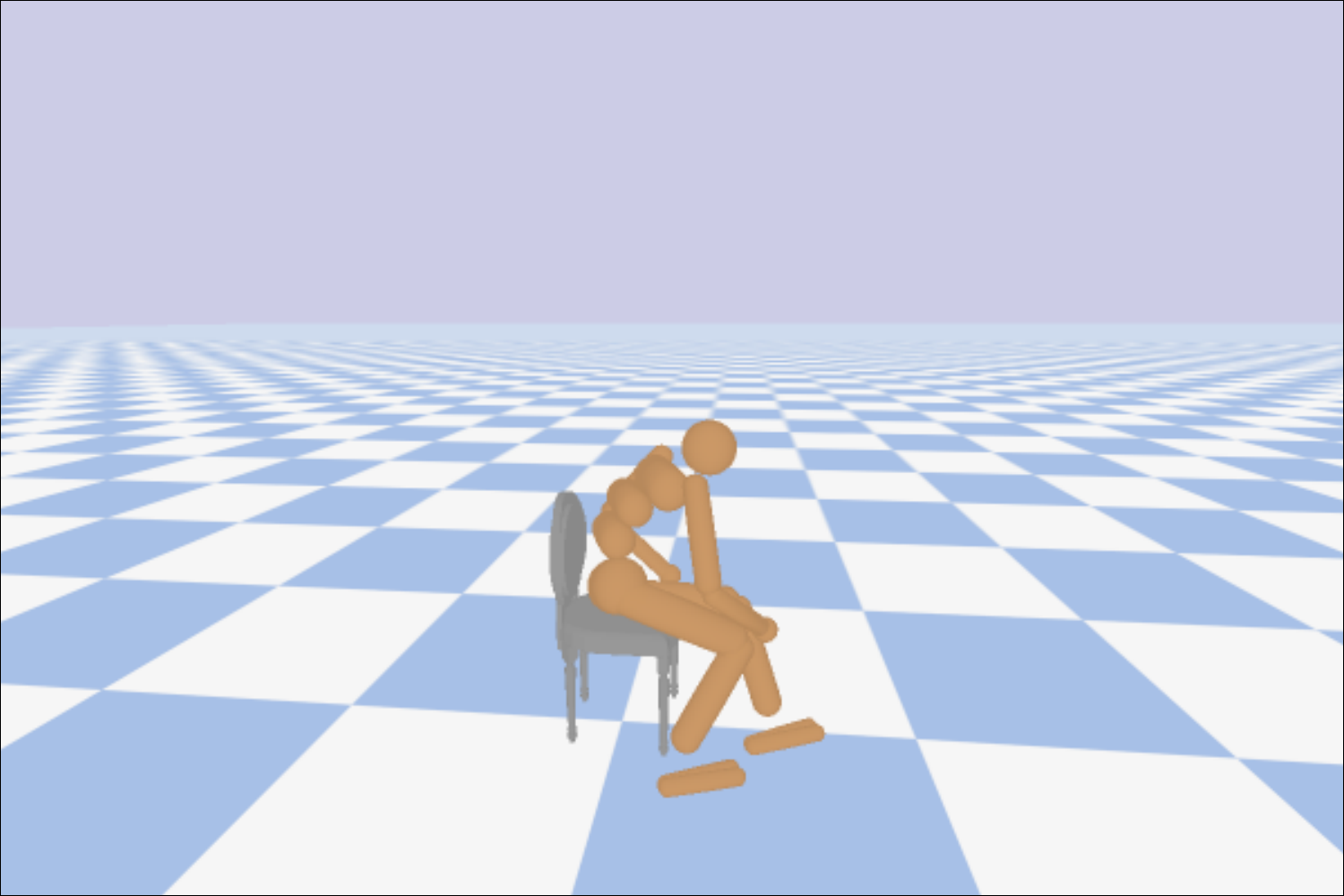} \end{minipage}
 \begin{minipage}{0.12\textwidth} \centering \includegraphics[width=1.00\textwidth,trim={80 40 80 40},clip]{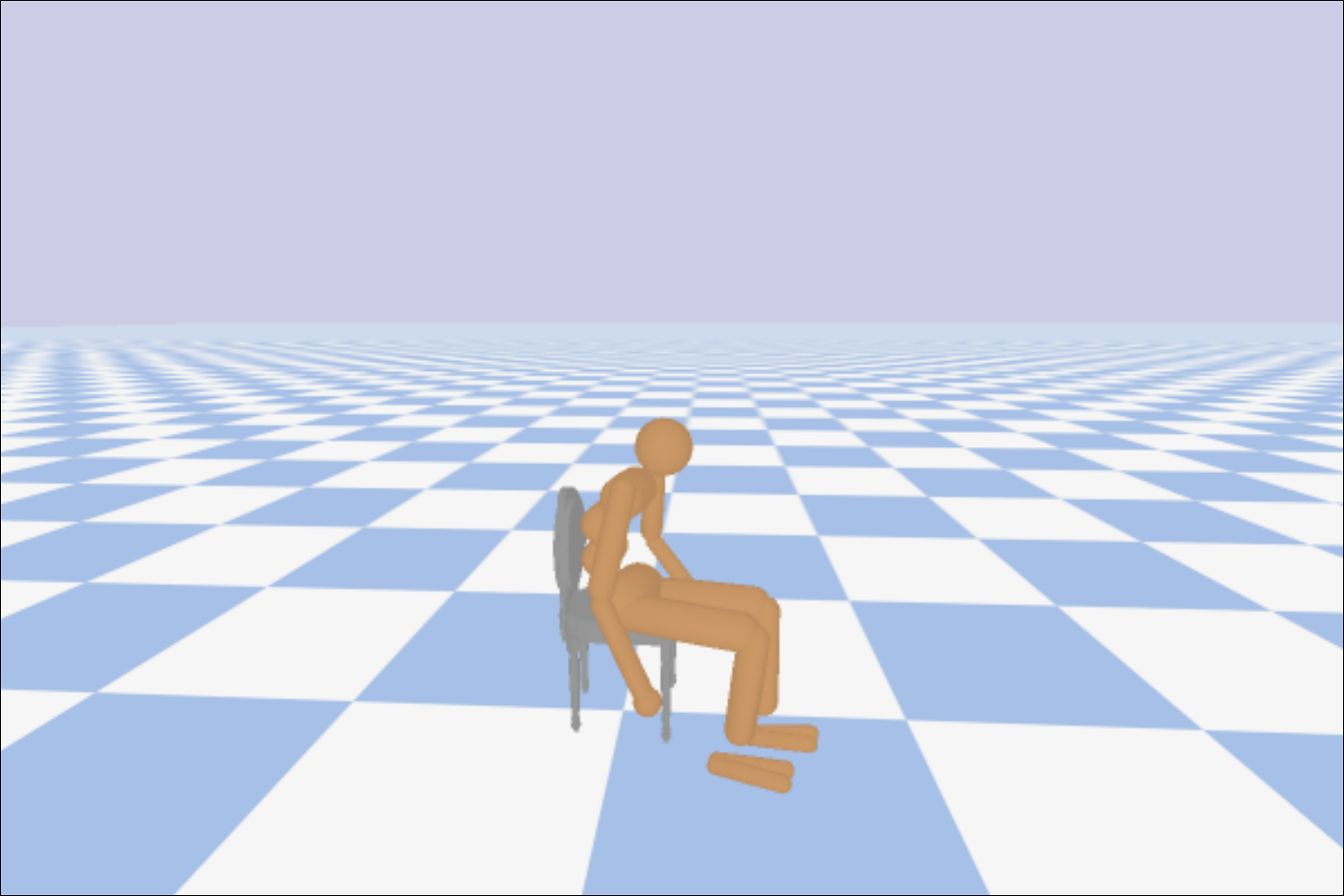} \end{minipage}
 \begin{minipage}{0.12\textwidth} \centering \includegraphics[width=1.00\textwidth,trim={80 40 80 40},clip]{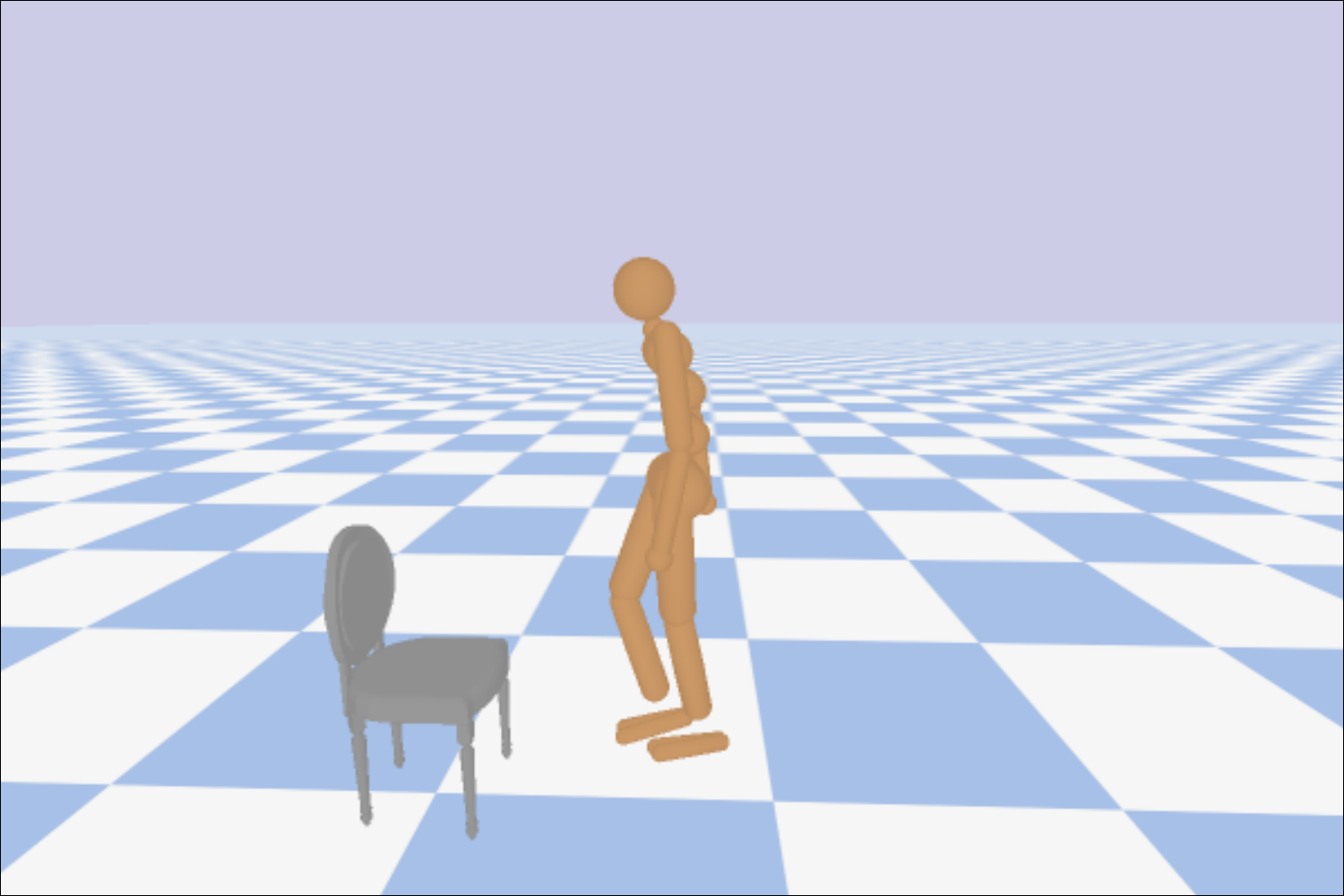} \end{minipage}
 \begin{minipage}{0.12\textwidth} \centering \includegraphics[width=1.00\textwidth,trim={80 40 80 40},clip]{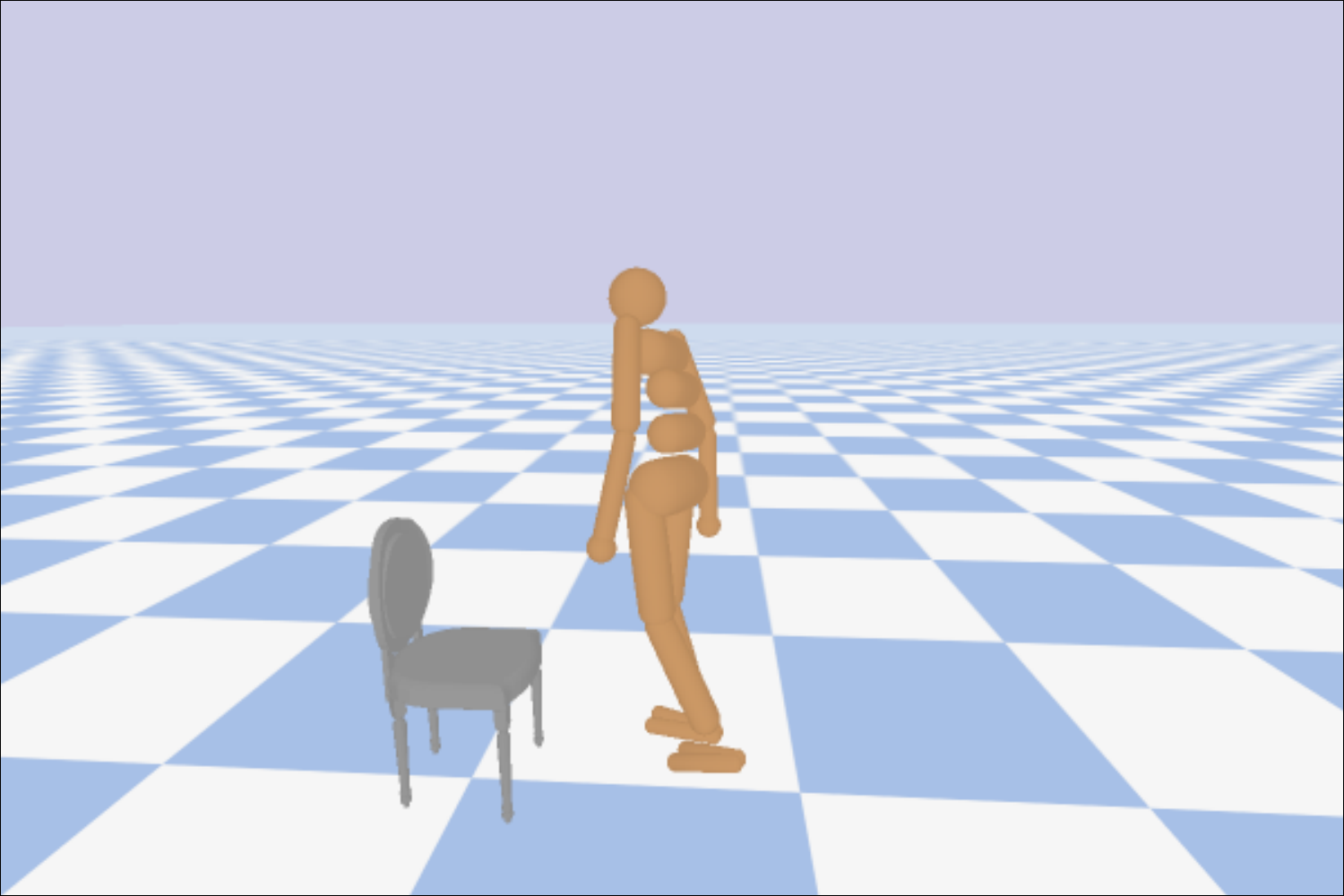} \end{minipage}
 \begin{minipage}{0.12\textwidth} \centering \includegraphics[width=1.00\textwidth,trim={80 40 80 40},clip]{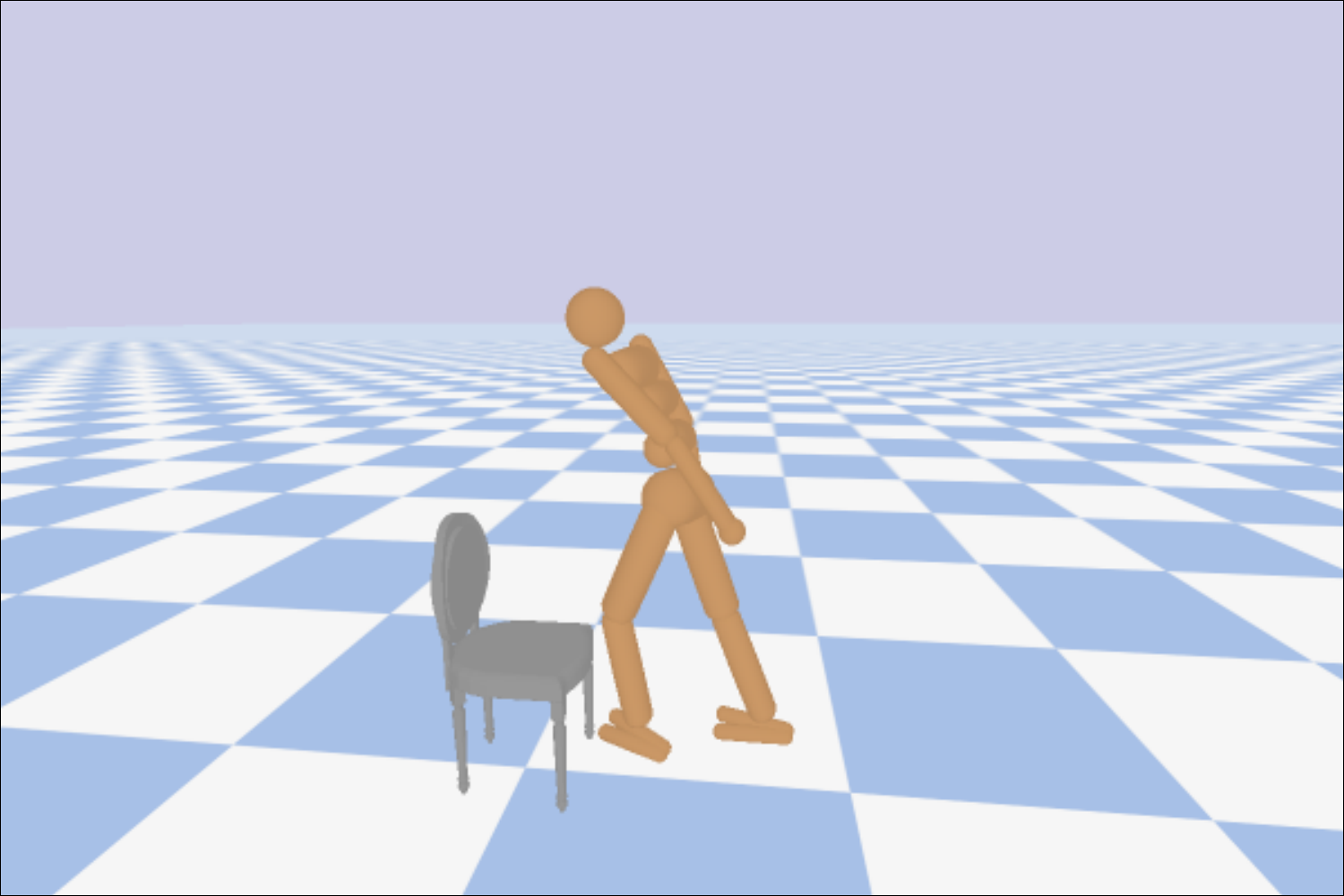} \end{minipage}
 \begin{minipage}{0.12\textwidth} \centering \includegraphics[width=1.00\textwidth,trim={80 40 80 40},clip]{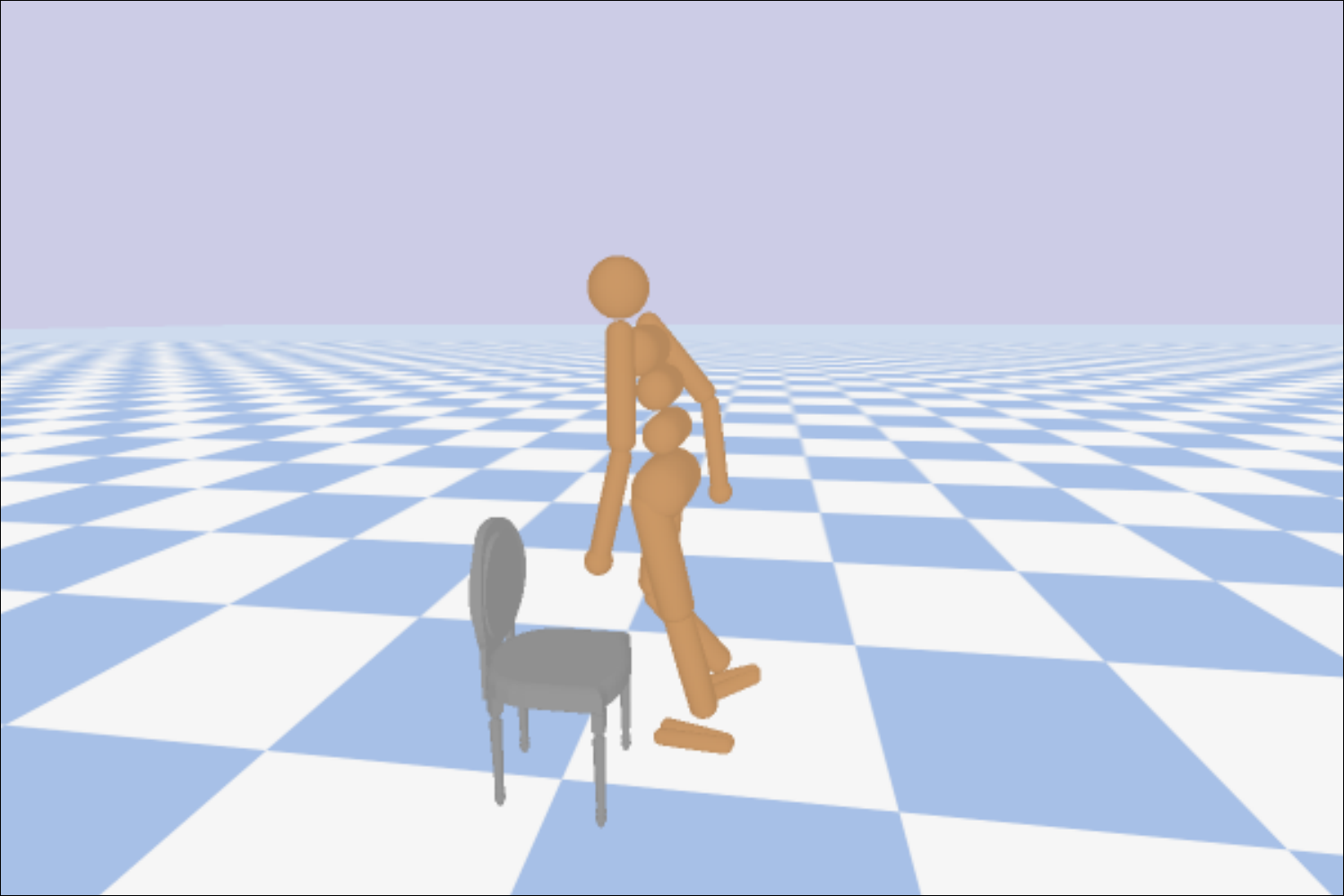} \end{minipage}
 \begin{minipage}{0.12\textwidth} \centering \includegraphics[width=1.00\textwidth,trim={80 40 80 40},clip]{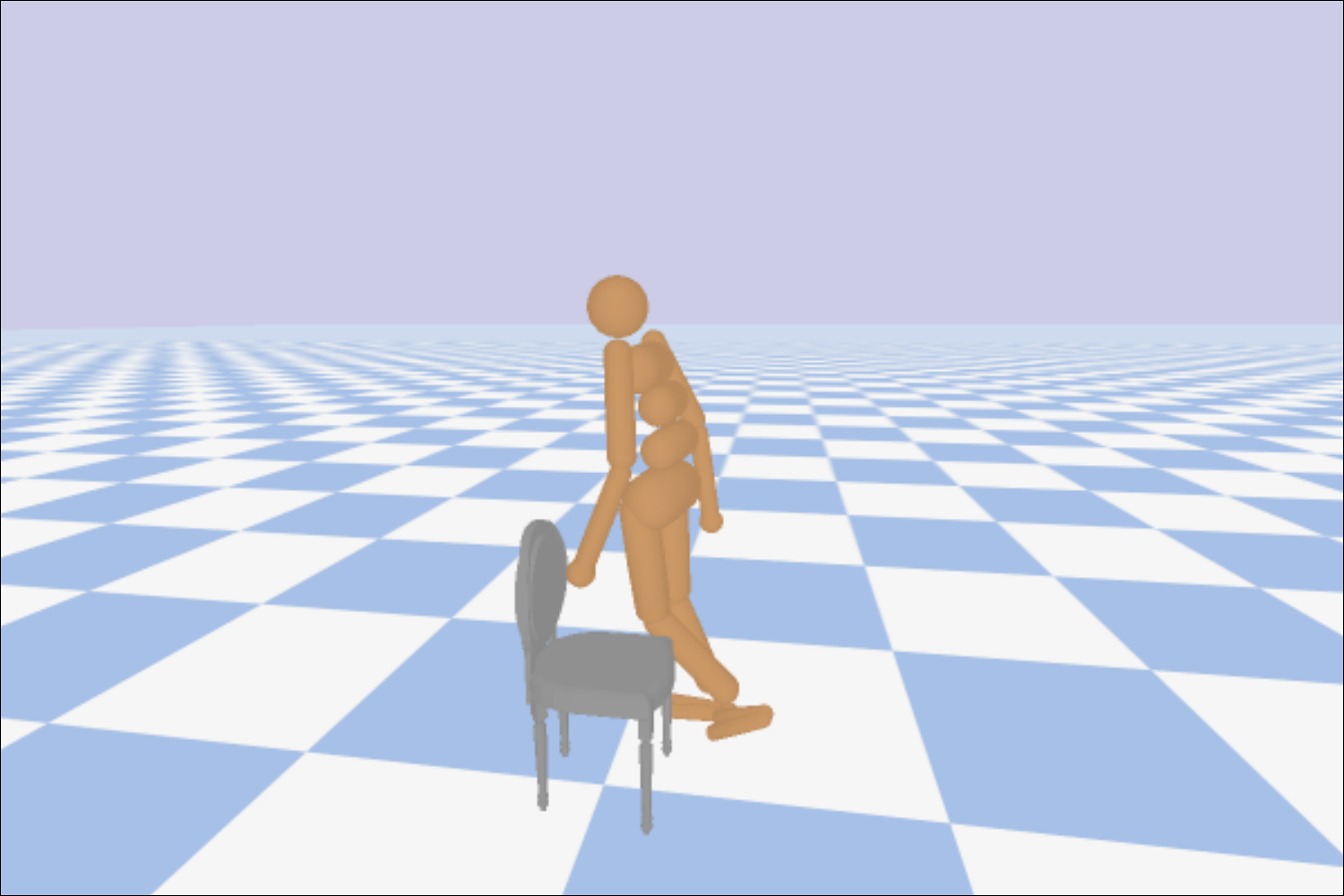} \end{minipage}
 \begin{minipage}{0.12\textwidth} \centering \includegraphics[width=1.00\textwidth,trim={80 40 80 40},clip]{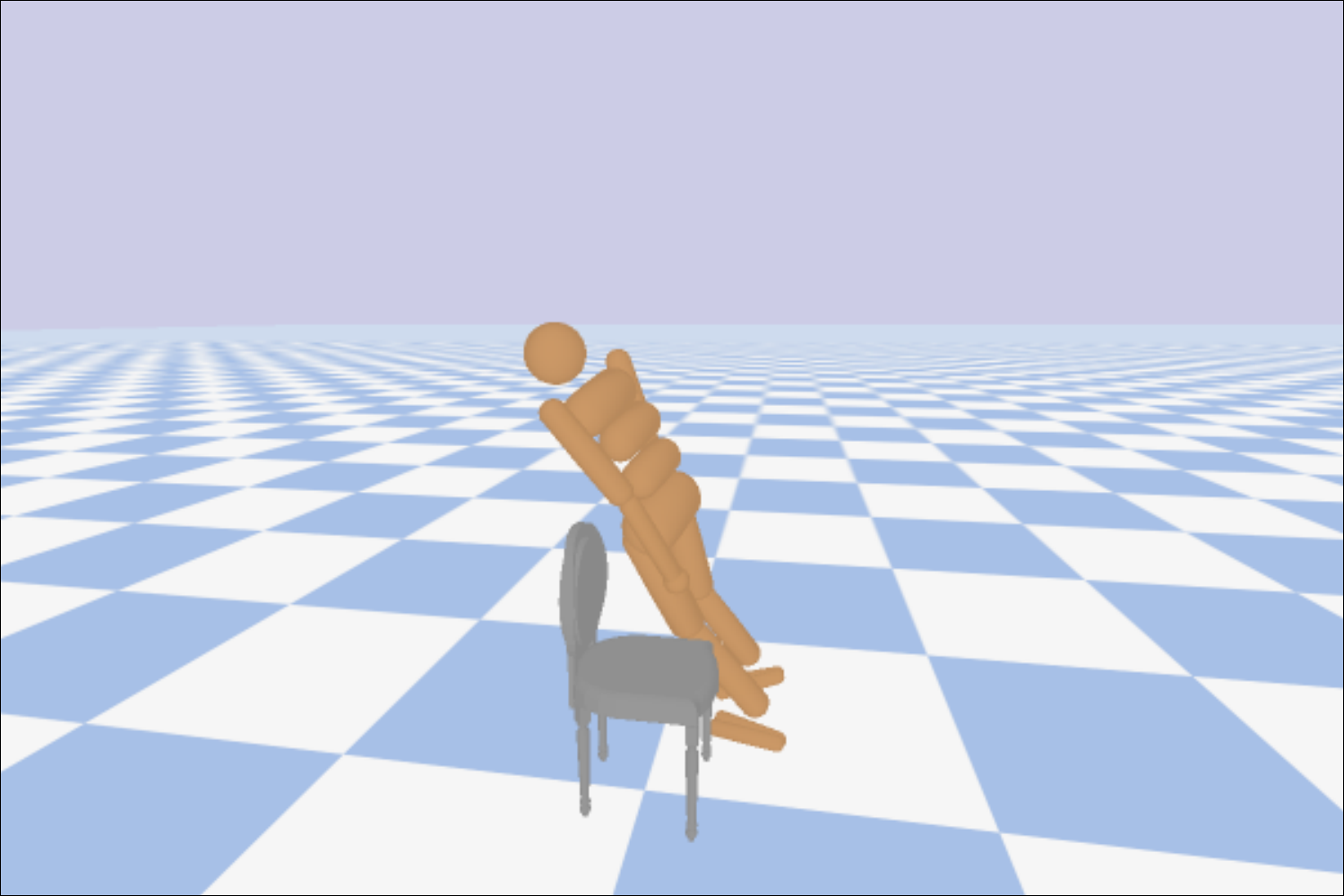} \end{minipage}
 \begin{minipage}{0.12\textwidth} \centering \includegraphics[width=1.00\textwidth,trim={80 40 80 40},clip]{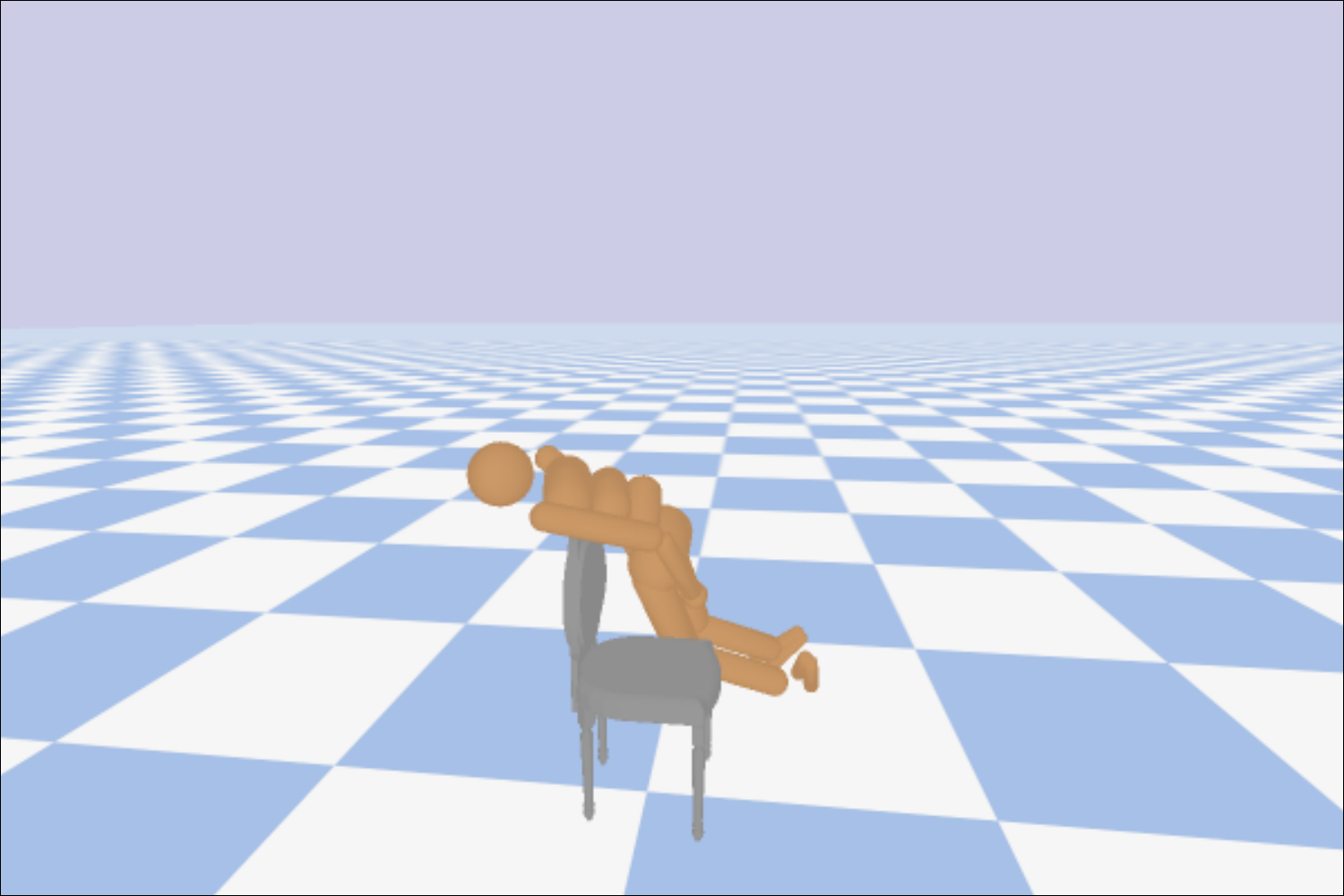} \end{minipage}
 \begin{minipage}{0.12\textwidth} \centering \includegraphics[width=1.00\textwidth,trim={80 40 80 40},clip]{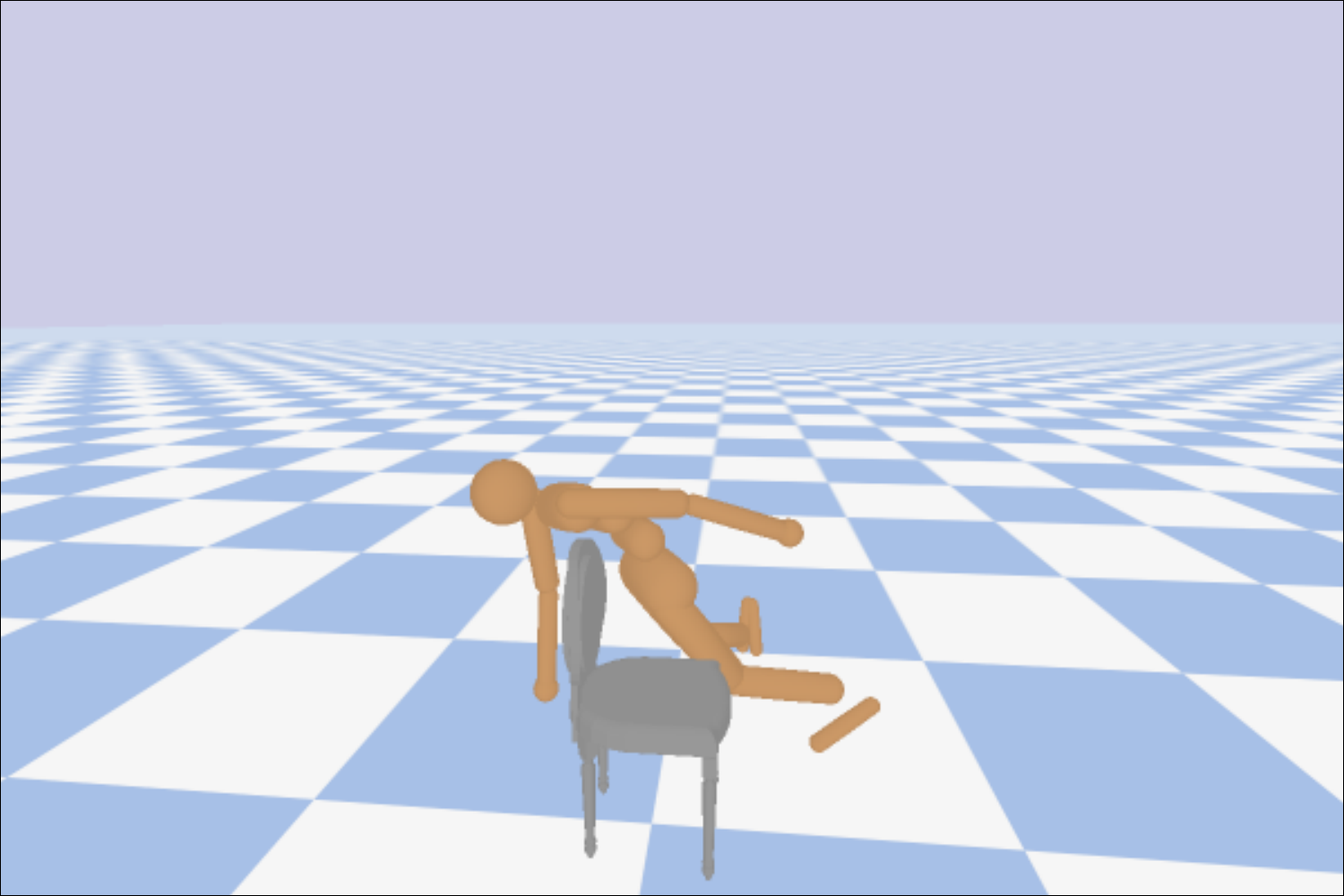} \end{minipage}
 \caption{\small Qualitative results of our approach and the baselines. Row 1
and 2 show failure cases from the kinematics and physics baselines,
respectively. The former violates physics rules (e.g. sitting in air), and both
do not generalize to new human-chair configurations. Row 3 to 4 show two
successful cases and row 5 shows one failure cases from our approach.}
 \label{fig:qual-easy}
\end{figure*}

\vspace{-3mm}

\paragraph{Easy Setting} We benchmark our approach against various baselines in
this setting. We start with two \textit{non-hierarchical} baselines. The first
is a \textit{kinematics}-based method: we select a mocap clip with a holistic
motion sequence that successively performs walking, turning, and sitting on a
chair. When a trial begins, we align the first frame of the sequence to the
humanoid's initial pose by aligning the yaw of the root. Once aligned, we
simply use the following frames of the sequence as the kinematic trajectory of
the trial. Note that this method is purely kinematic and cannot reflect any
physical interactions between the humanoid and chair. The second method extends
the first one to a \textit{physics}-based approach: we use the same kinematic
sequence but now train a controller to imitate the motion. This is similar
to~\cite{peng:siggraph2018} and equivalent to training a subtask controller
except the subtask is holistic (i.e. containing walk, turn, and sit in one
reference motion). Both methods are considered non-hierarchical as neither
performs task decomposition.

\begin{table}[t]
 \centering
 \small
 \begin{tabular}{cc||cc}
  \hline \TBstrut
  Subtask                  & Initial Pose                         & \multicolumn{2}{c}{Succ Rate (\%)} \\
  \hline \Tstrut
  \textit{left turn}       & mocap                                & \multicolumn{2}{c}{87.02}          \\
  \textit{right turn}      & mocap                                & \multicolumn{2}{c}{67.59}          \\ \Bstrut
  \textit{sit}             & mocap                                & \multicolumn{2}{c}{99.25}          \\
  \hline
  \hline \Tstrut
  \multirow{2}{*}{Subtask} & \multirow{2}{*}{Initial Pose}        & \multicolumn{2}{c}{Succ Rate (\%)} \\ \Bstrut
                           &                                      & w/o FT & w/ FT                     \\
  \hline \Tstrut
  \textit{left turn}       & \textit{walk}                        & ~~0.09 & 51.12                     \\
  \textit{right turn}      & \textit{walk}                        & ~~1.96 & 58.31                     \\ \Bstrut
  \textit{sit}             & \textit{left} or \textit{right turn} & 32.94  & 87.41                     \\
  \hline
 \end{tabular}
 \caption{\small Evaluation of individual subtasks.}
 \label{tab:subtasks}
\end{table}

Tab.~\ref{tab:baselines} shows the quantitative results. For the kinematics
baseline, the success rate is not reported since we are unable to detect
physical contact between the pelvis and chair. However, the 1.2656 mean minimum
distance suggests that the humanoid on average remains far from the chair. For
the physics baseline, we observe a similar mean minimum distance (i.e. 1.3316).
The zero success rate is unsurprising given that the humanoid is unable to get
close to the chair in most trials. As shown in the qualitative examples
(Fig.~\ref{fig:qual-easy}), the motion generated by the kinematics baseline
(row 1) is not physics realistic (e.g. sitting in air). The physics baseline
(row 2), while following physics rules (e.g. falling on the ground eventually),
still fails in approaching the chair. These holistic baselines perform poorly
since they simply imitate the mocap example and repeat the same motion pattern
regardless of their starting position.

We now turn to a set of \textit{hierarchical} baselines and our approach. We
also consider two baselines. The first one always executes the subtasks in a
pre-defined order, and the meta controller is only used to trigger transitions
(i.e. a binary classification). Note that this is in similar spirit to
\cite{clegg:siggraphasia2018}. We consider two particular orders:
\textit{walk}$\rightarrow$\textit{left turn}$\rightarrow$\textit{sit} and
\textit{walk}$\rightarrow$\textit{right turn}$\rightarrow$\textit{sit}. The
second one is a degenerated version of our approach that uses either only
\textit{left turn} or \textit{right turn}: \textit{walk} / \textit{left turn} /
\textit{sit} and \textit{walk} / \textit{right turn} / \textit{sit}.

\begin{figure}[t]
 \centering
 \begin{subfigure}[c]{0.415\linewidth}
  \centering
  \includegraphics[width=1.00\textwidth]{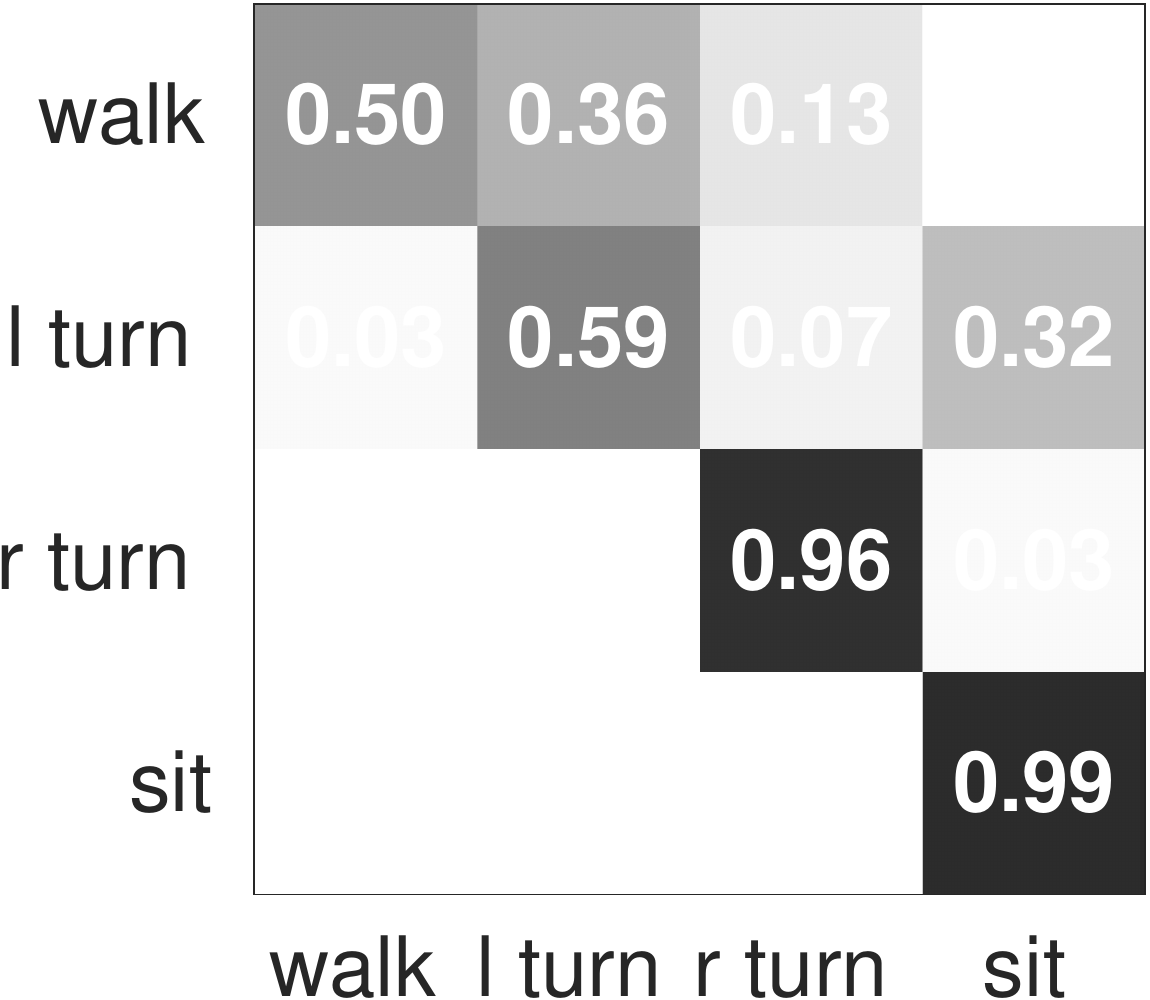}
  \caption{left side}
 \end{subfigure}
 ~~~~~~~
 \begin{subfigure}[c]{0.415\linewidth}
  \centering
  \includegraphics[width=1.00\textwidth]{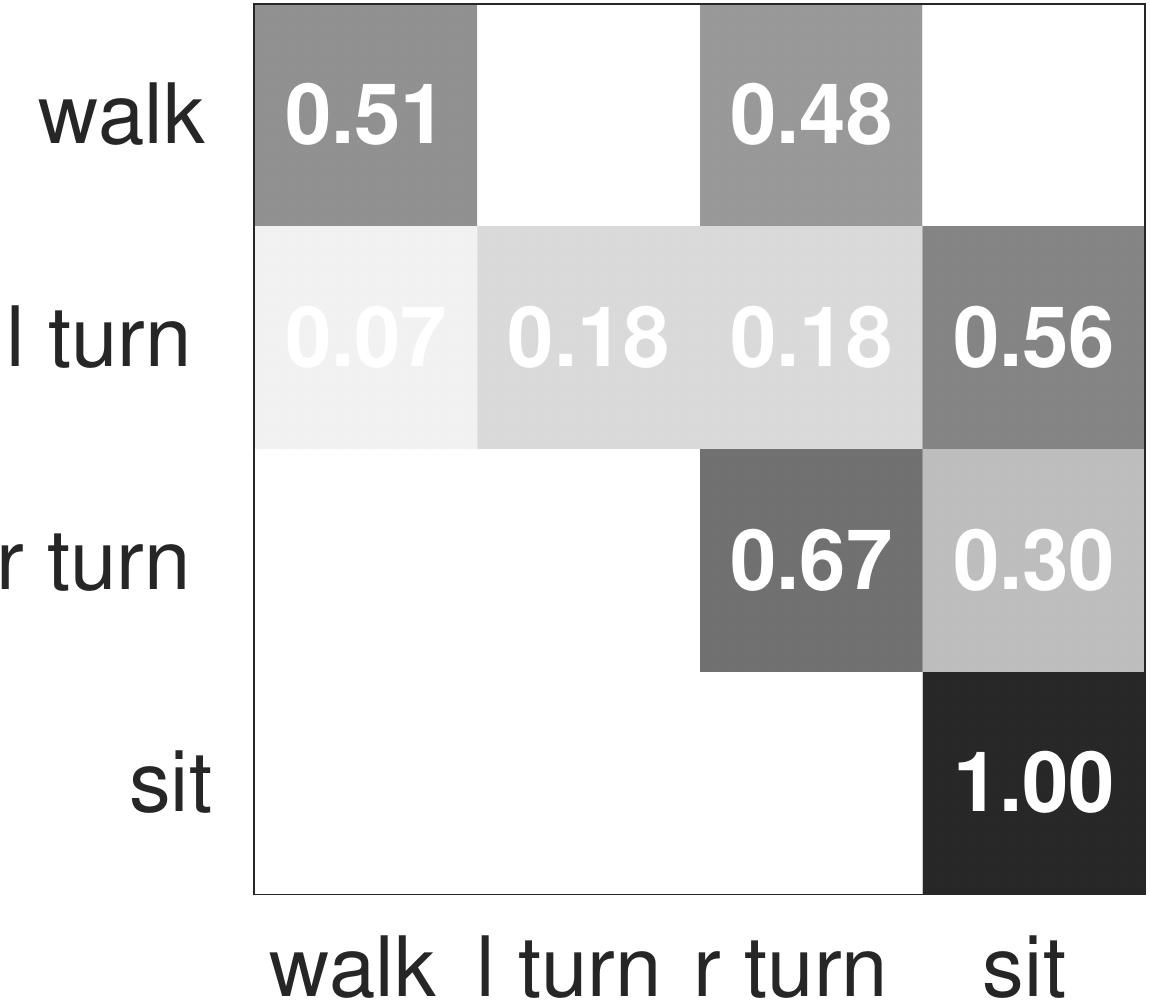}
  \caption{right side}
 \end{subfigure}
 \caption{\small Transition matrices from different sides of the chair.}
 \label{fig:transition}
\end{figure}

\begin{table}[t]
 \centering
 \small
 \begin{tabular}{l||cc}
  \hline \TBstrut
                & Succ Rate (\%)  & Min Dist (m)                 \\
  \hline\hline \TBstrut
  Zone 1        & 31.61           & 0.3303 $\pm$ 0.2393          \\ \hline\hline \Tstrut
  Zone 2 w/o CL & ~~0.00          & 0.5549 $\pm$ 0.2549          \\ \Bstrut
  Zone 2        & \textbf{10.01}  & \textbf{0.5526 $\pm$ 0.3303} \\ \hline\hline \Tstrut
  Zone 3 w/o CL & ~~4.05          & 0.5636 $\pm$ 0.2263          \\ \Bstrut
  Zone 3 w/ CL  & \textbf{~~7.05} & \textbf{0.5262 $\pm$ 0.2602} \\
  \hline
 \end{tabular}
 \caption{\small Comparison of the Easy and Hard settings. Applying the
curriculum learning strategy improves the performance.}
 \label{tab:curriculum}
\end{table}

\begin{figure*}[t]
 \centering
 \begin{minipage}{0.12\textwidth} \centering \includegraphics[width=1.00\textwidth]{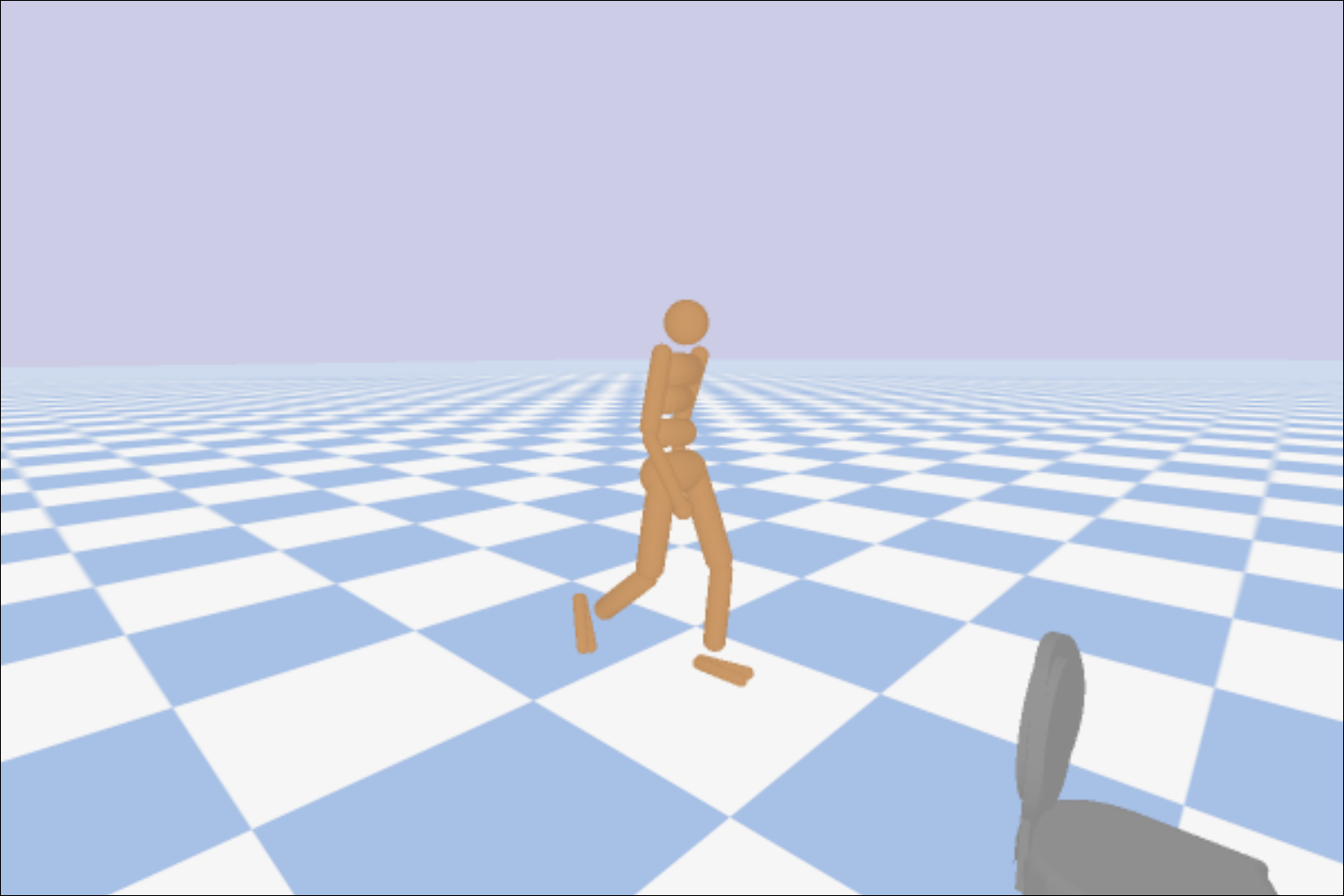} \end{minipage}
 \begin{minipage}{0.12\textwidth} \centering \includegraphics[width=1.00\textwidth]{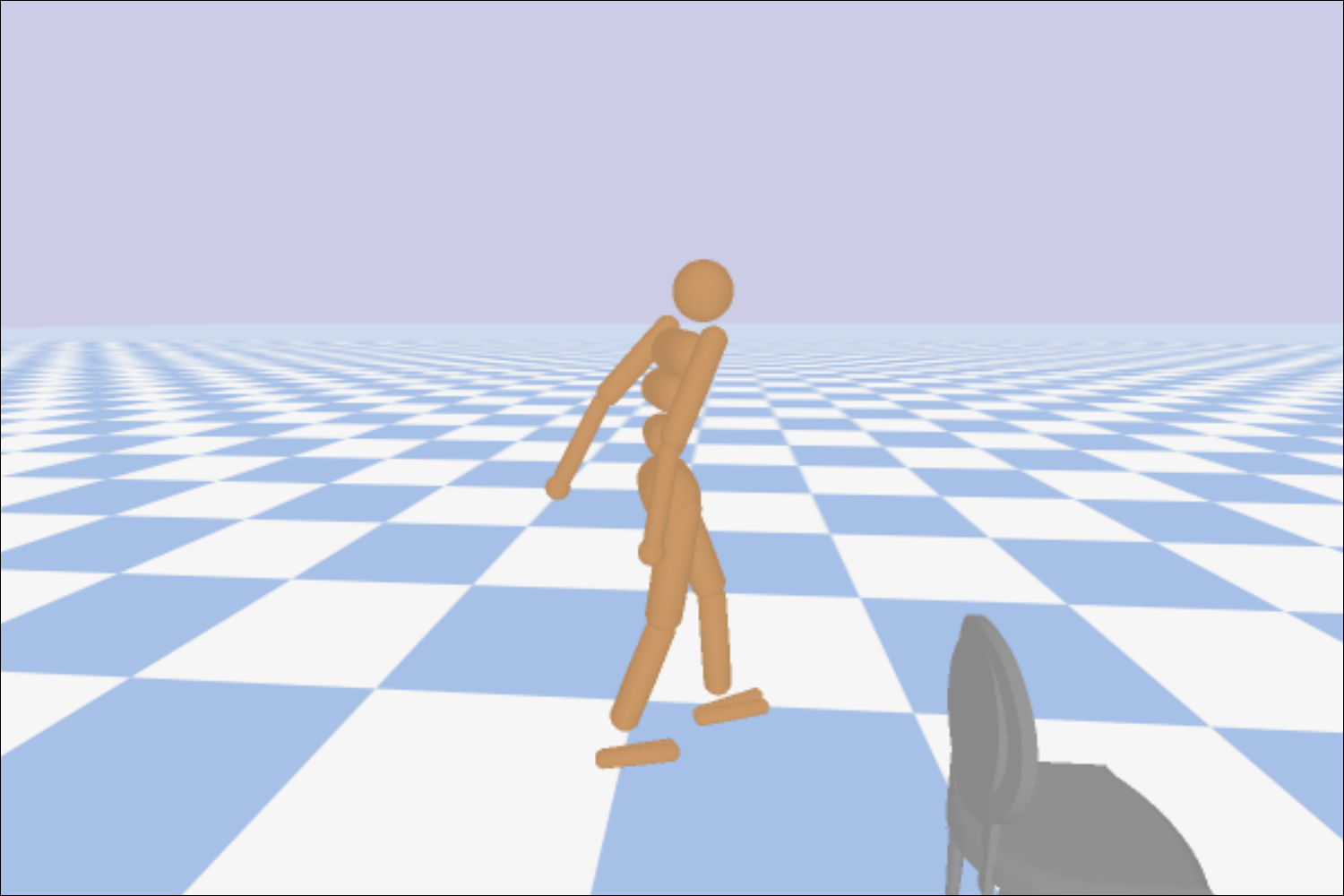} \end{minipage}
 \begin{minipage}{0.12\textwidth} \centering \includegraphics[width=1.00\textwidth]{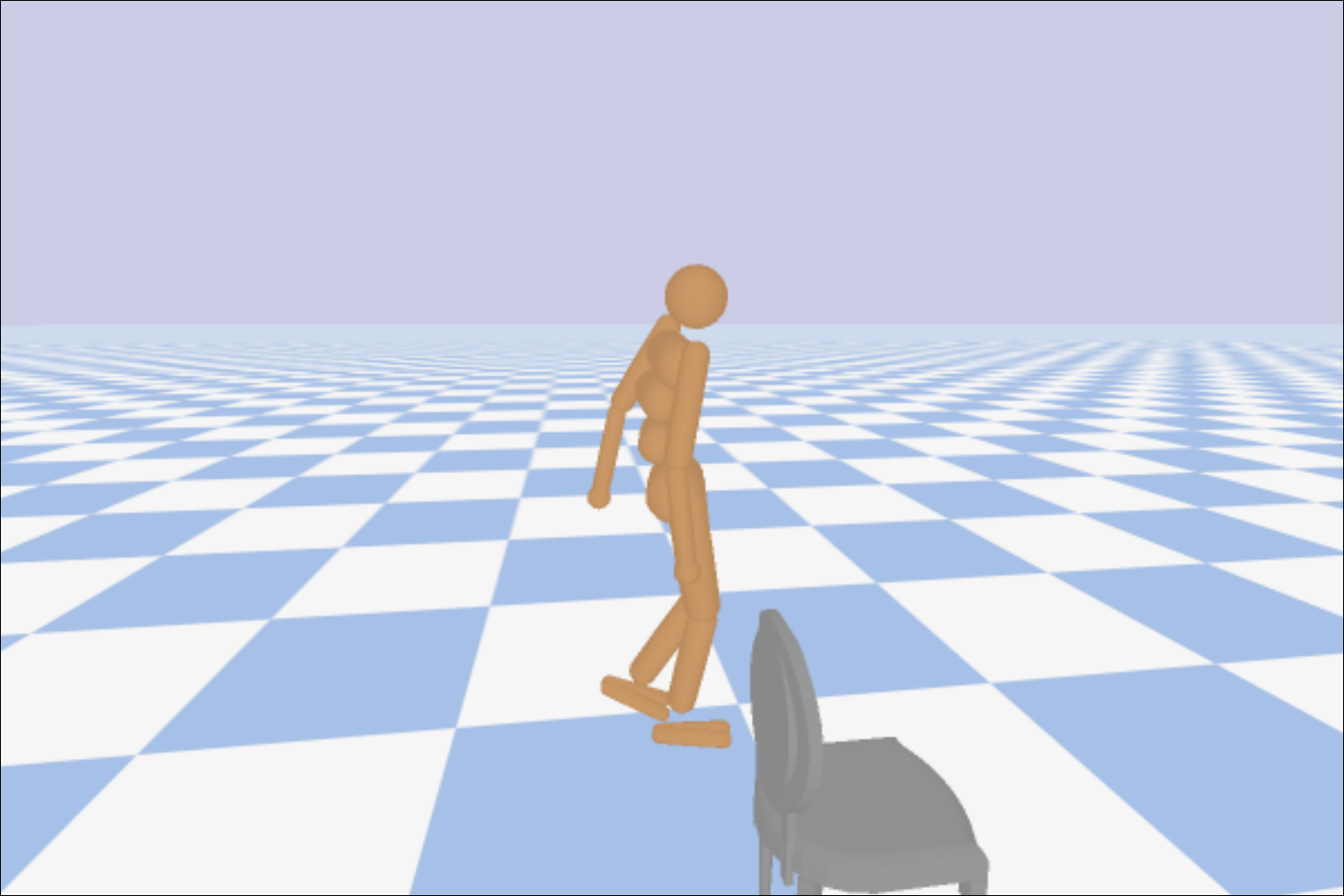} \end{minipage}
 \begin{minipage}{0.12\textwidth} \centering \includegraphics[width=1.00\textwidth]{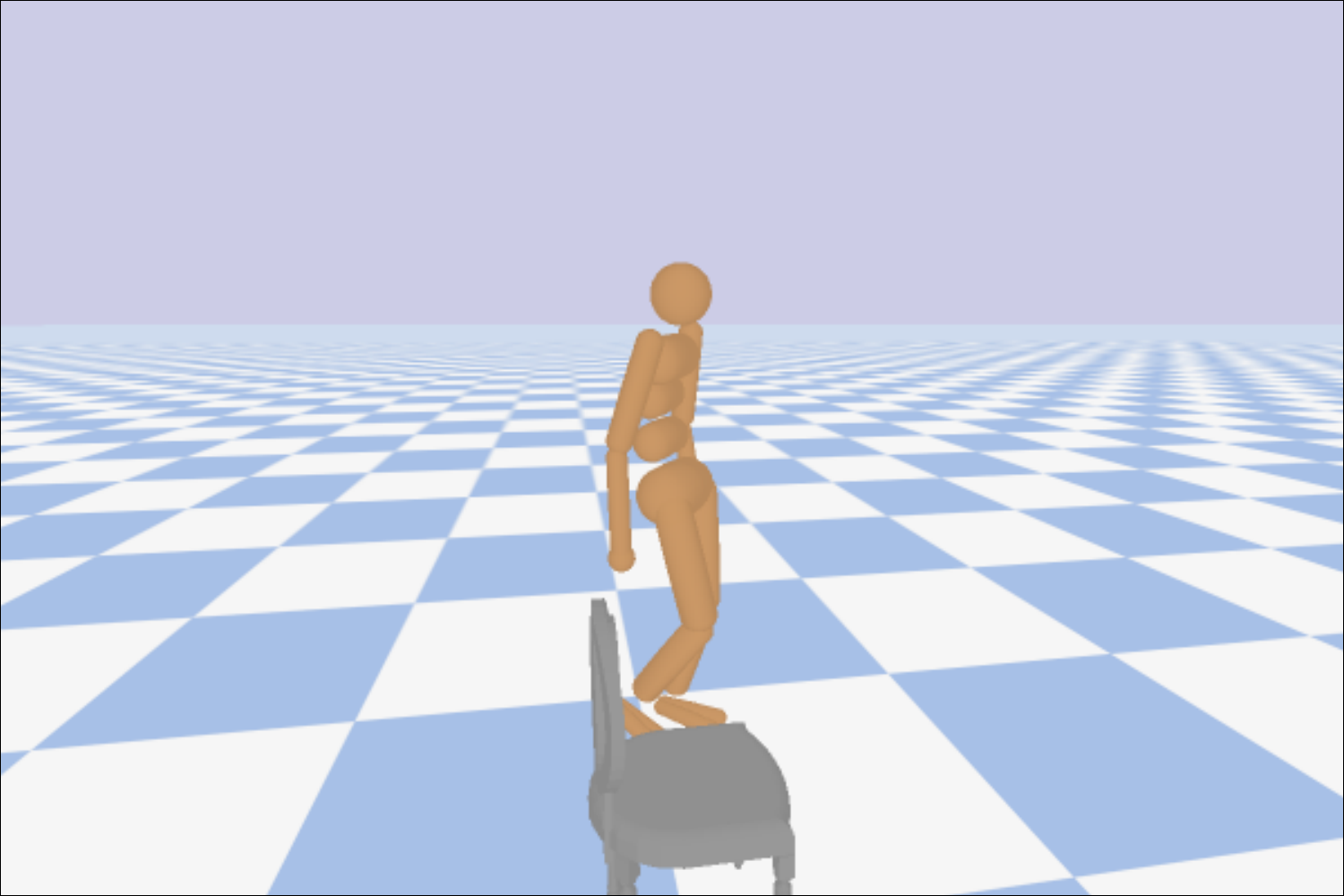} \end{minipage}
 \begin{minipage}{0.12\textwidth} \centering \includegraphics[width=1.00\textwidth]{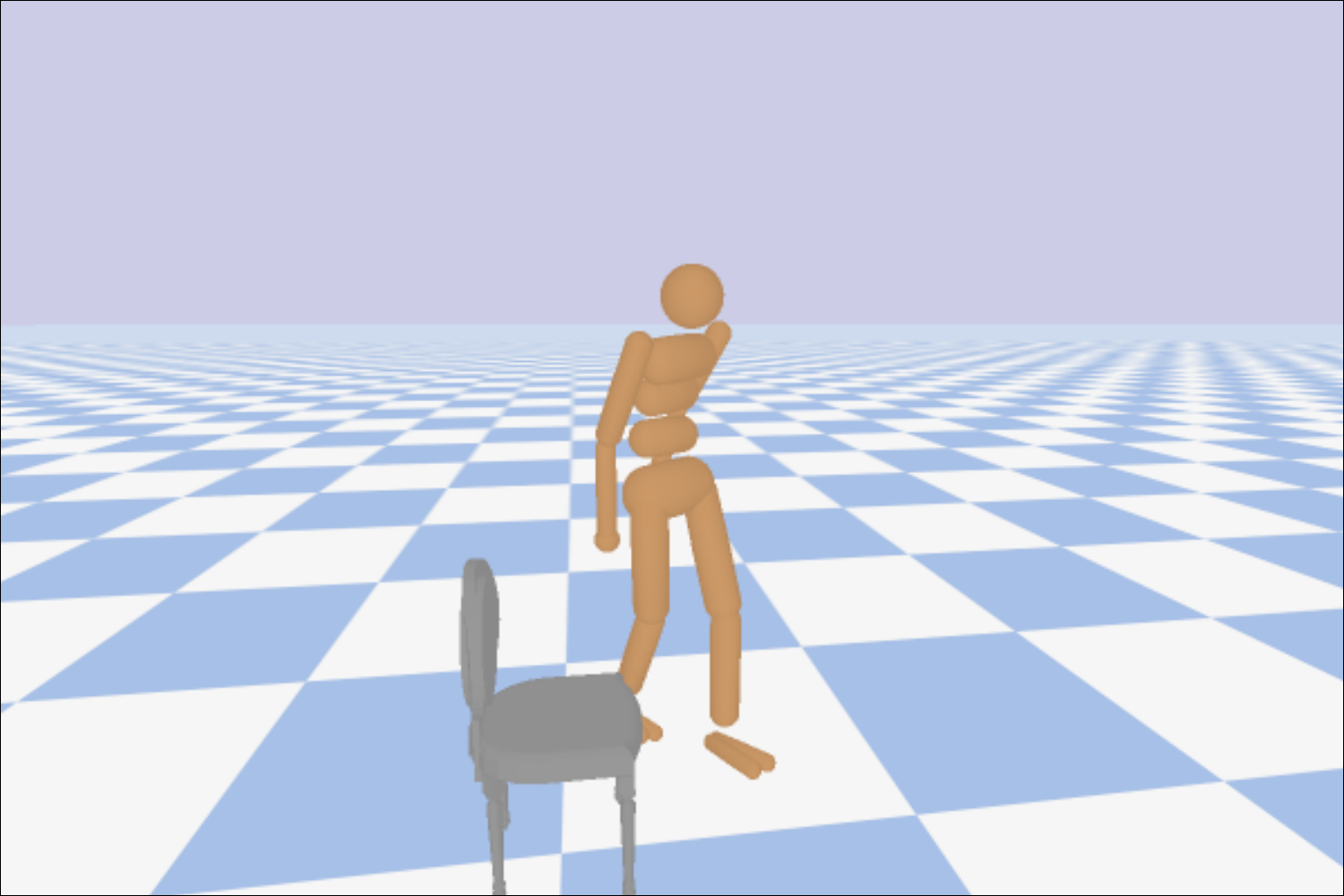} \end{minipage}
 \begin{minipage}{0.12\textwidth} \centering \includegraphics[width=1.00\textwidth]{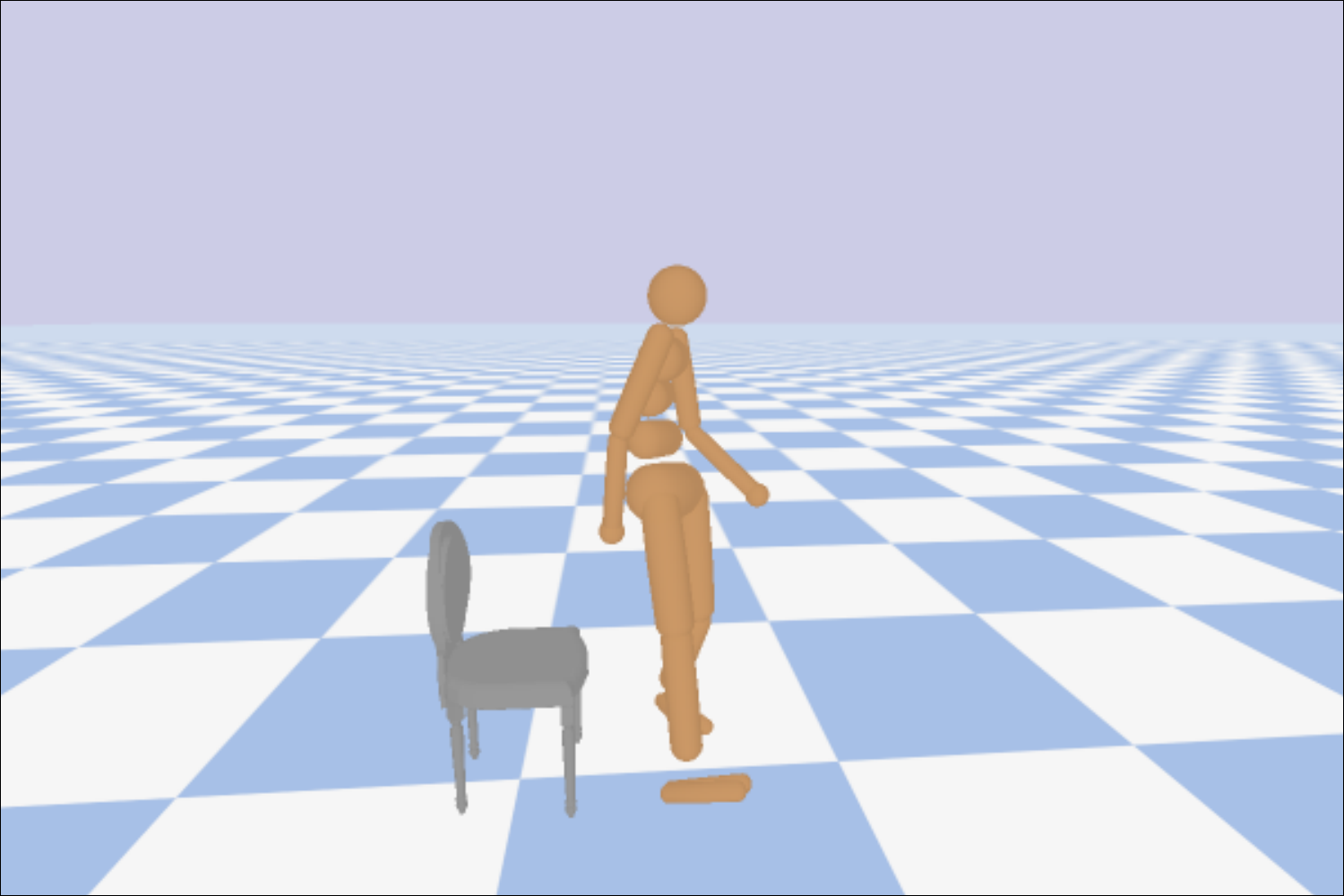} \end{minipage}
 \begin{minipage}{0.12\textwidth} \centering \includegraphics[width=1.00\textwidth]{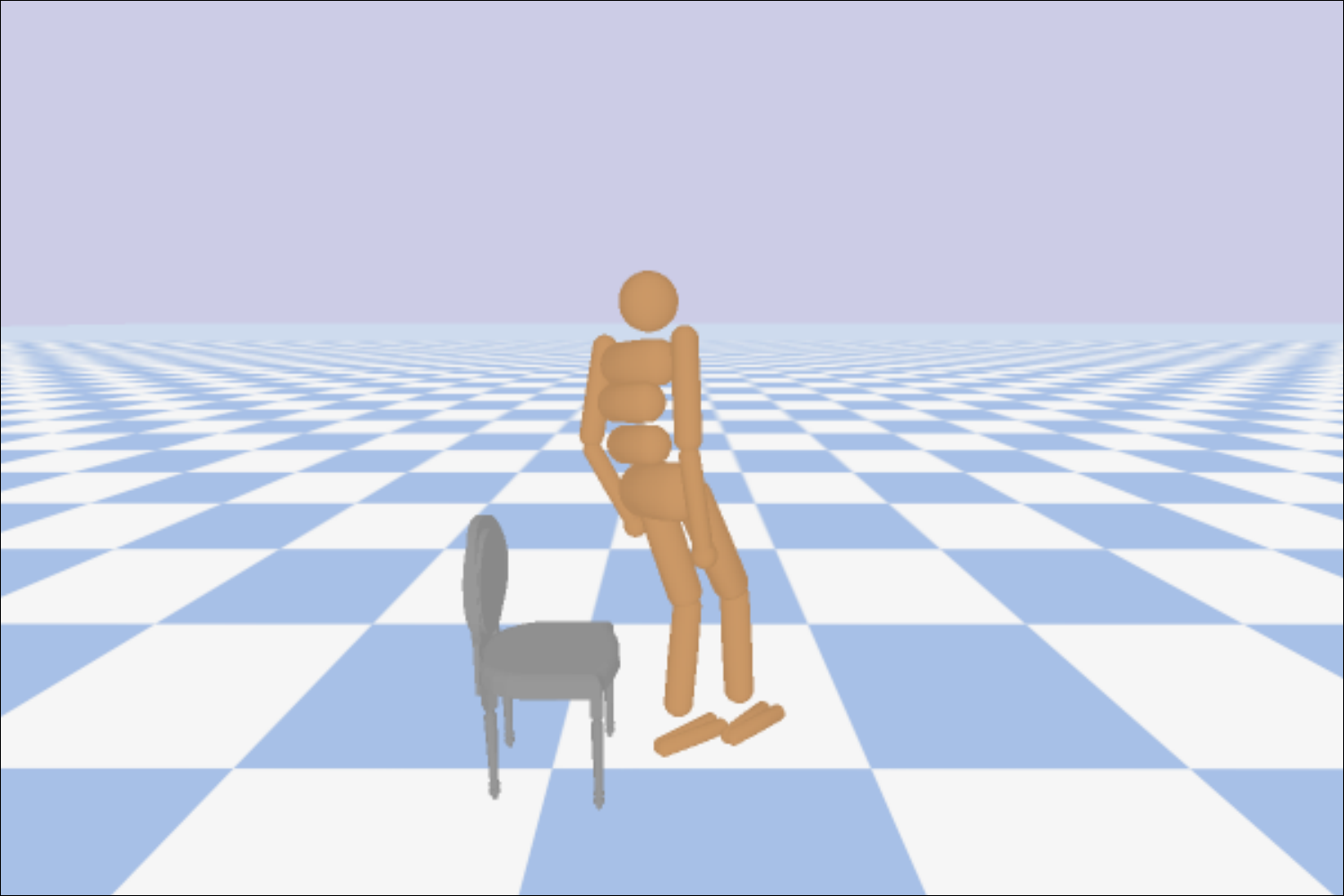} \end{minipage}
 \begin{minipage}{0.12\textwidth} \centering \includegraphics[width=1.00\textwidth]{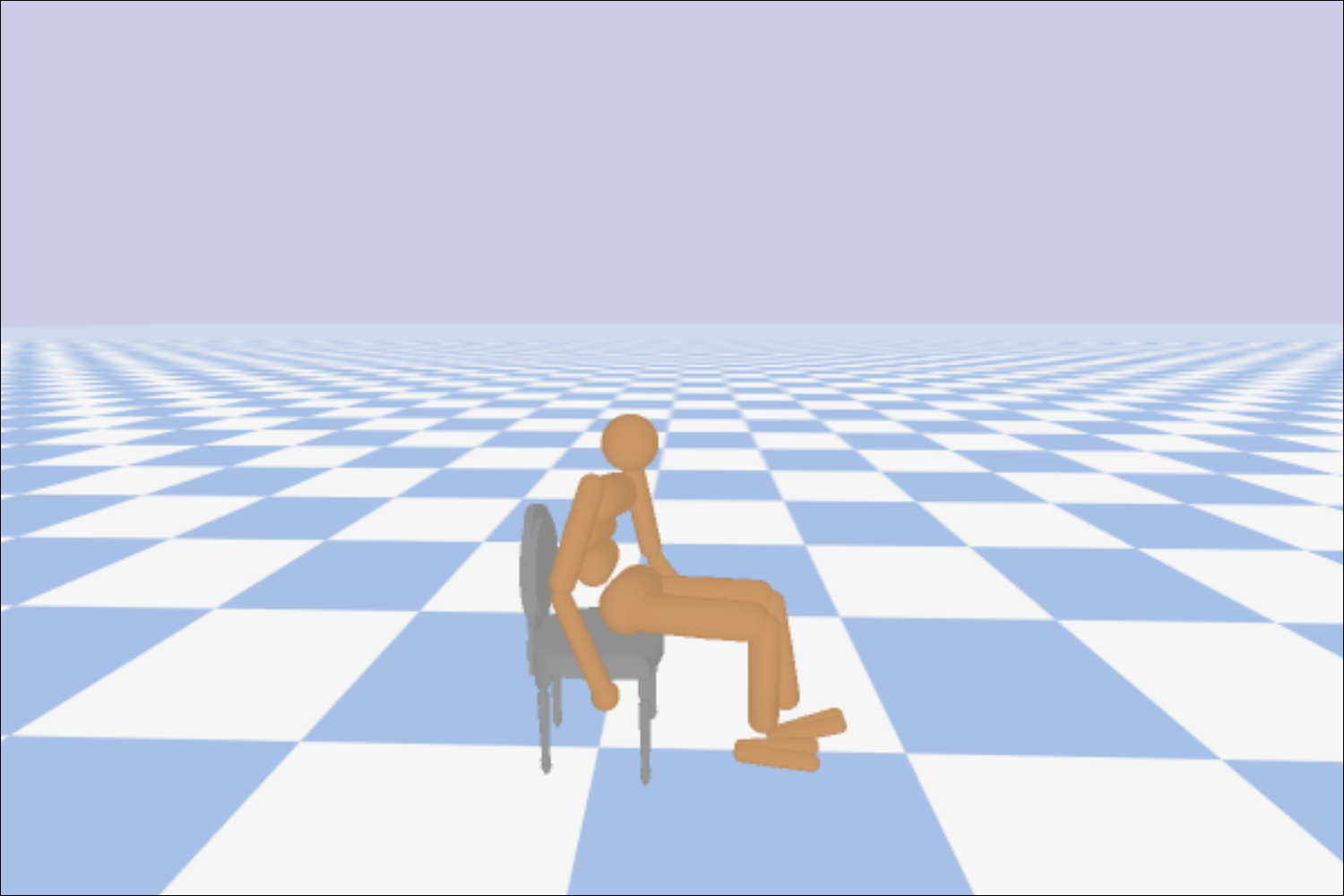} \end{minipage}
 \\ \vspace{1mm}
 \begin{minipage}{0.12\textwidth} \centering \includegraphics[width=1.00\textwidth]{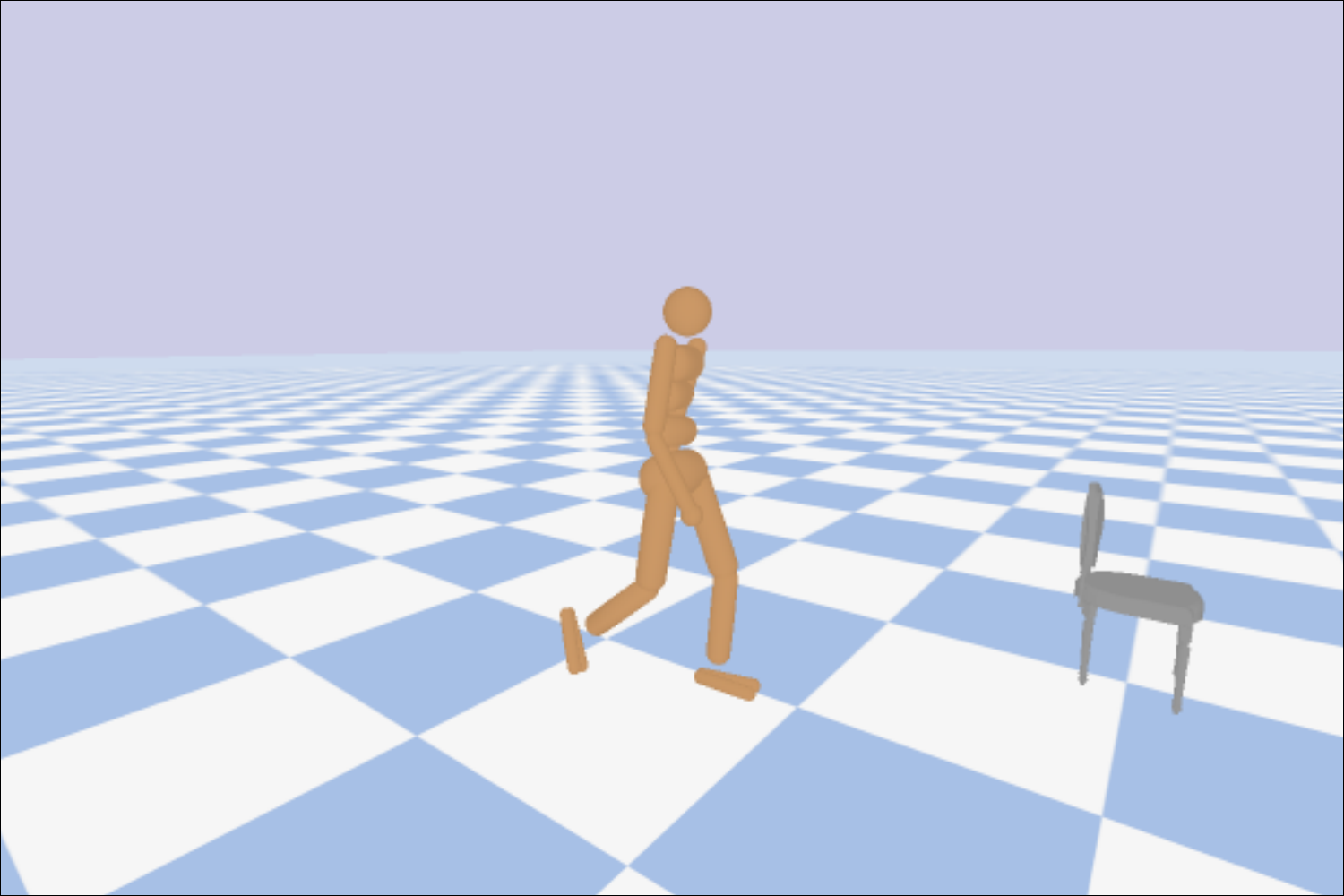} \end{minipage}
 \begin{minipage}{0.12\textwidth} \centering \includegraphics[width=1.00\textwidth]{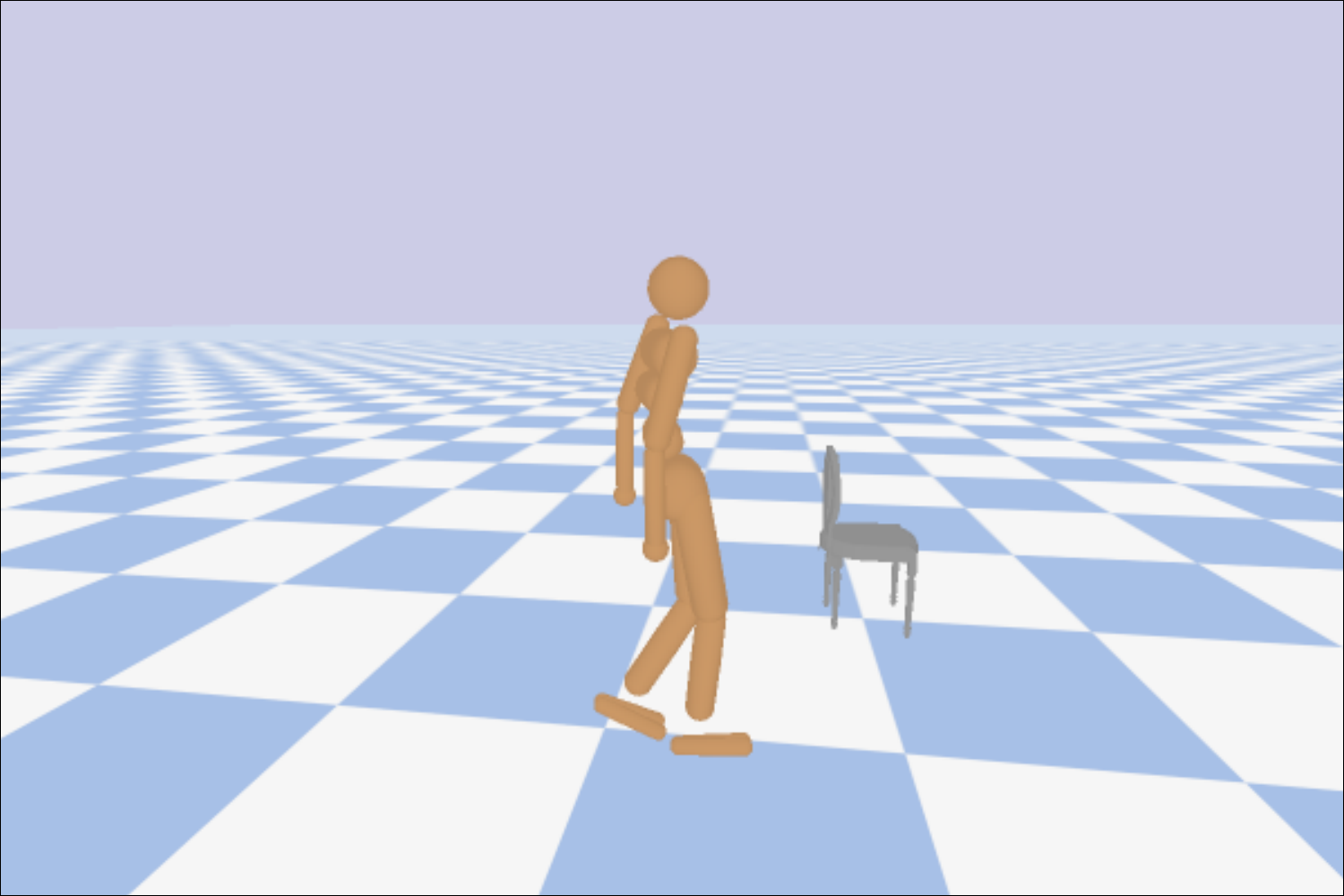} \end{minipage}
 \begin{minipage}{0.12\textwidth} \centering \includegraphics[width=1.00\textwidth]{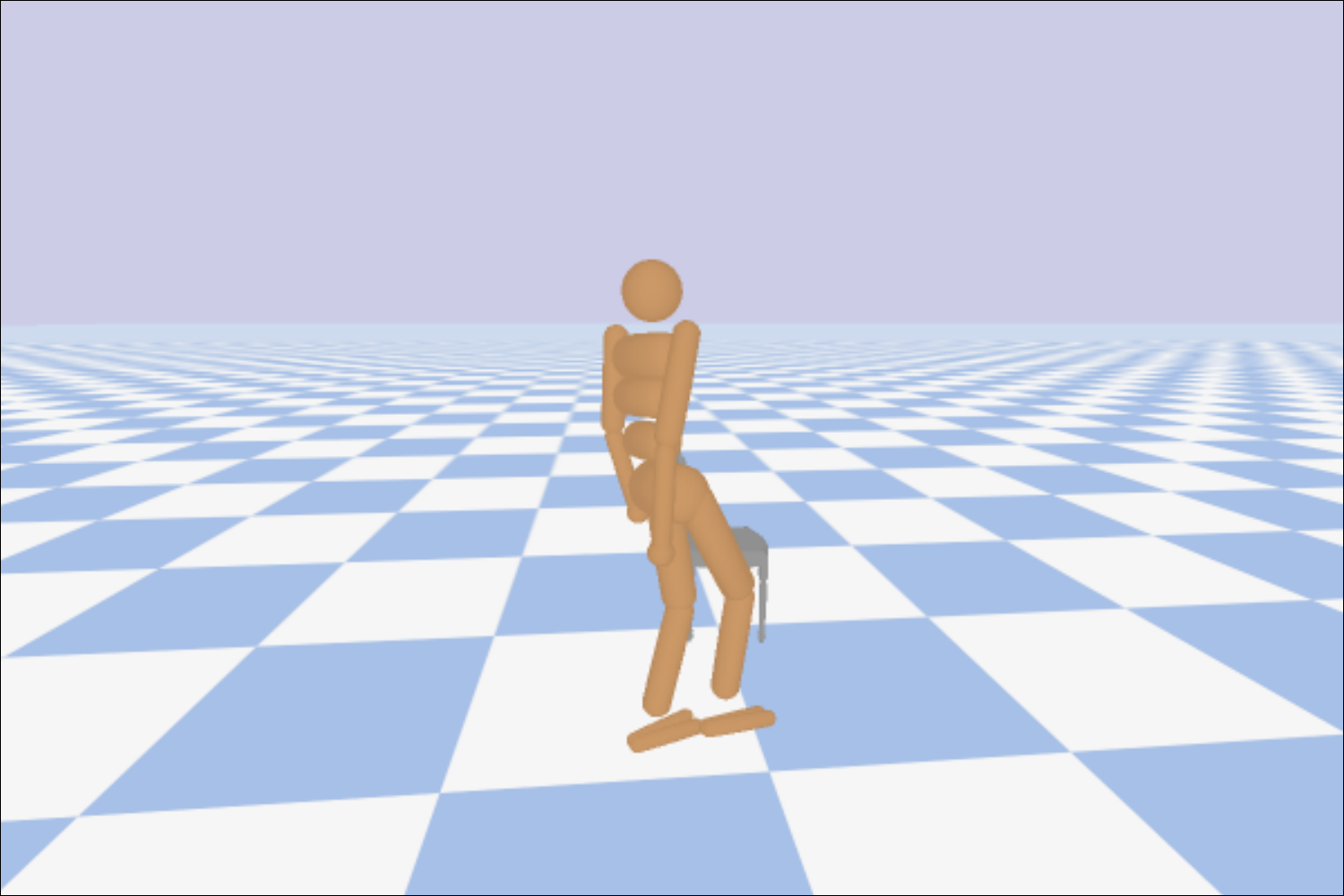} \end{minipage}
 \begin{minipage}{0.12\textwidth} \centering \includegraphics[width=1.00\textwidth]{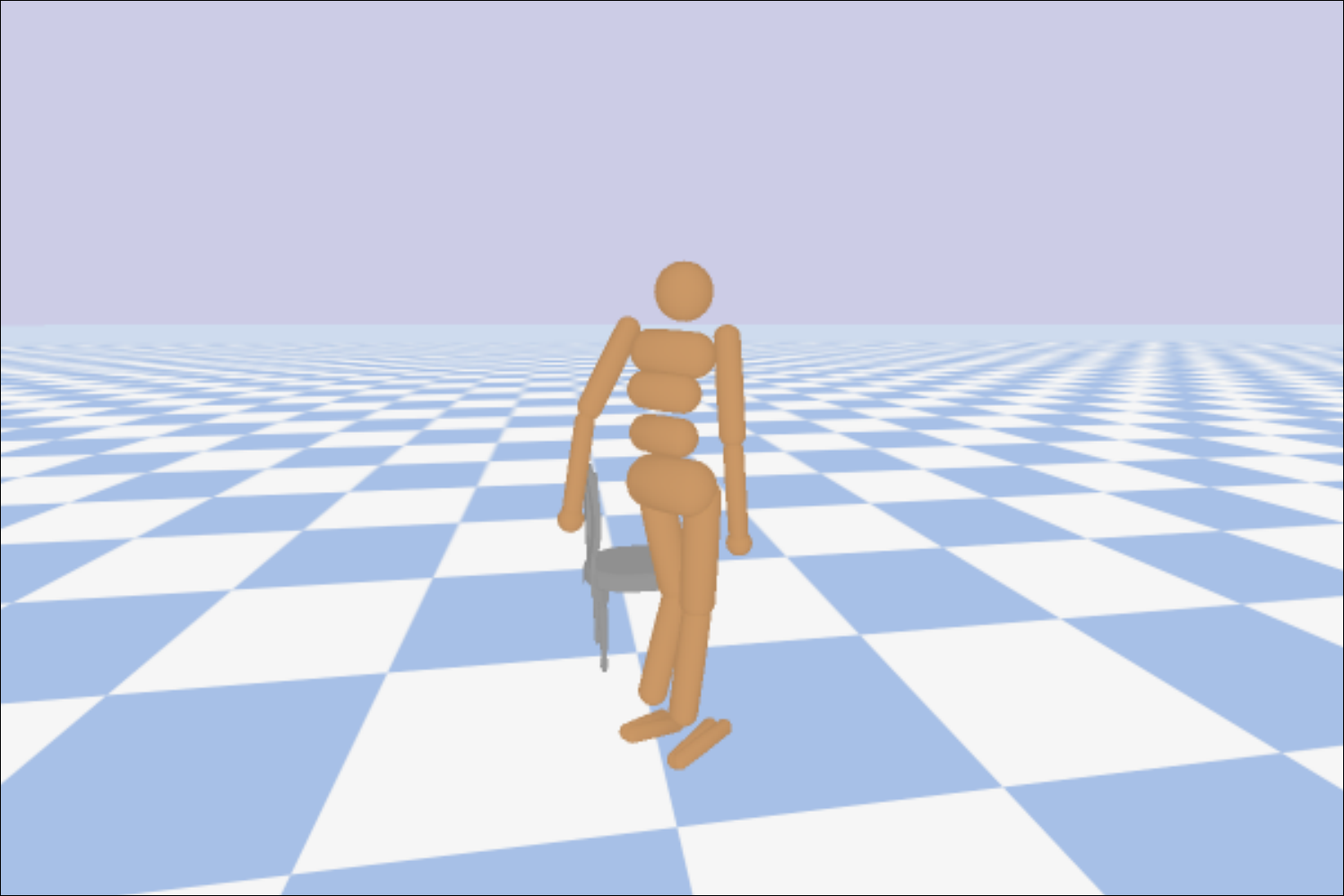} \end{minipage}
 \begin{minipage}{0.12\textwidth} \centering \includegraphics[width=1.00\textwidth]{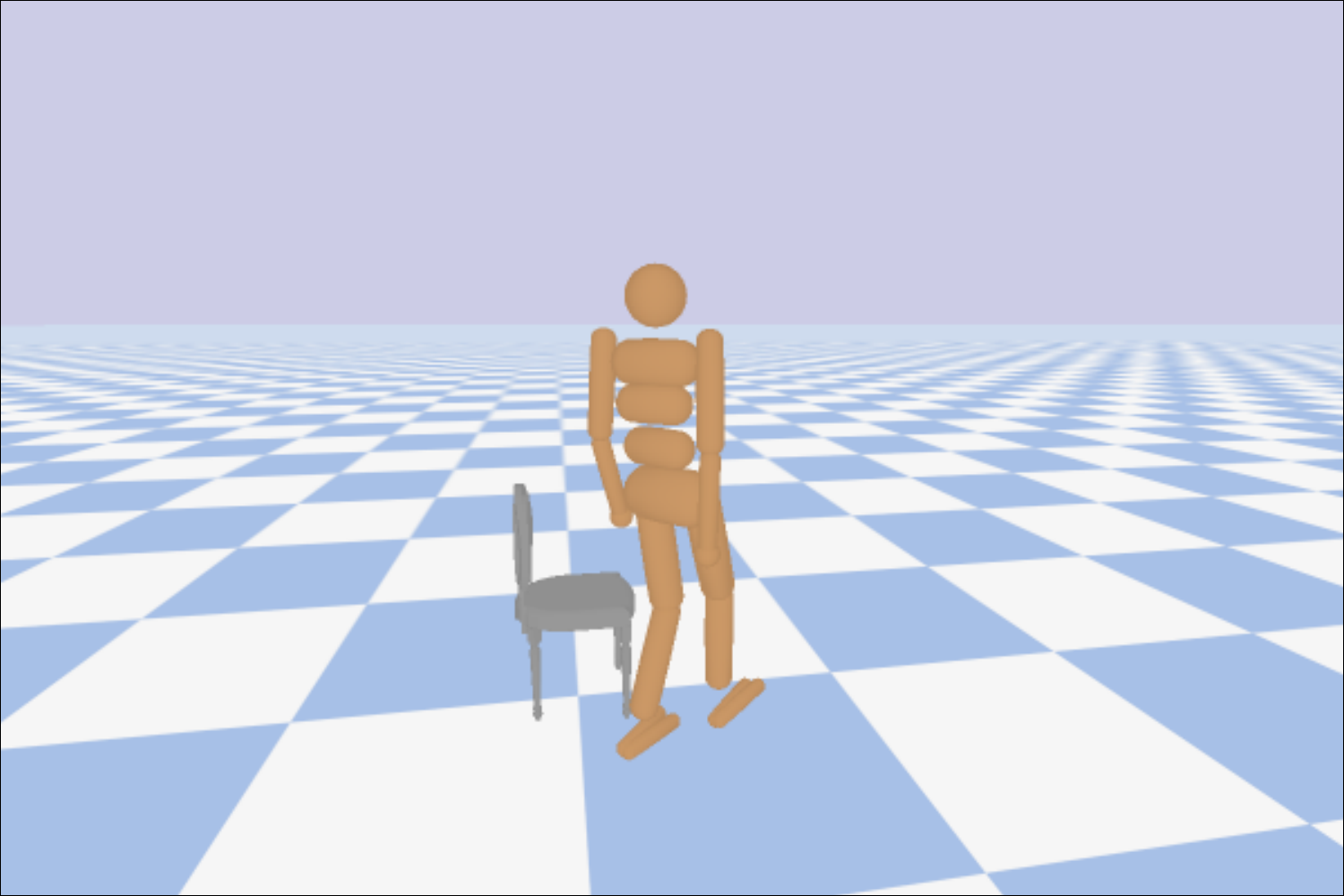} \end{minipage}
 \begin{minipage}{0.12\textwidth} \centering \includegraphics[width=1.00\textwidth]{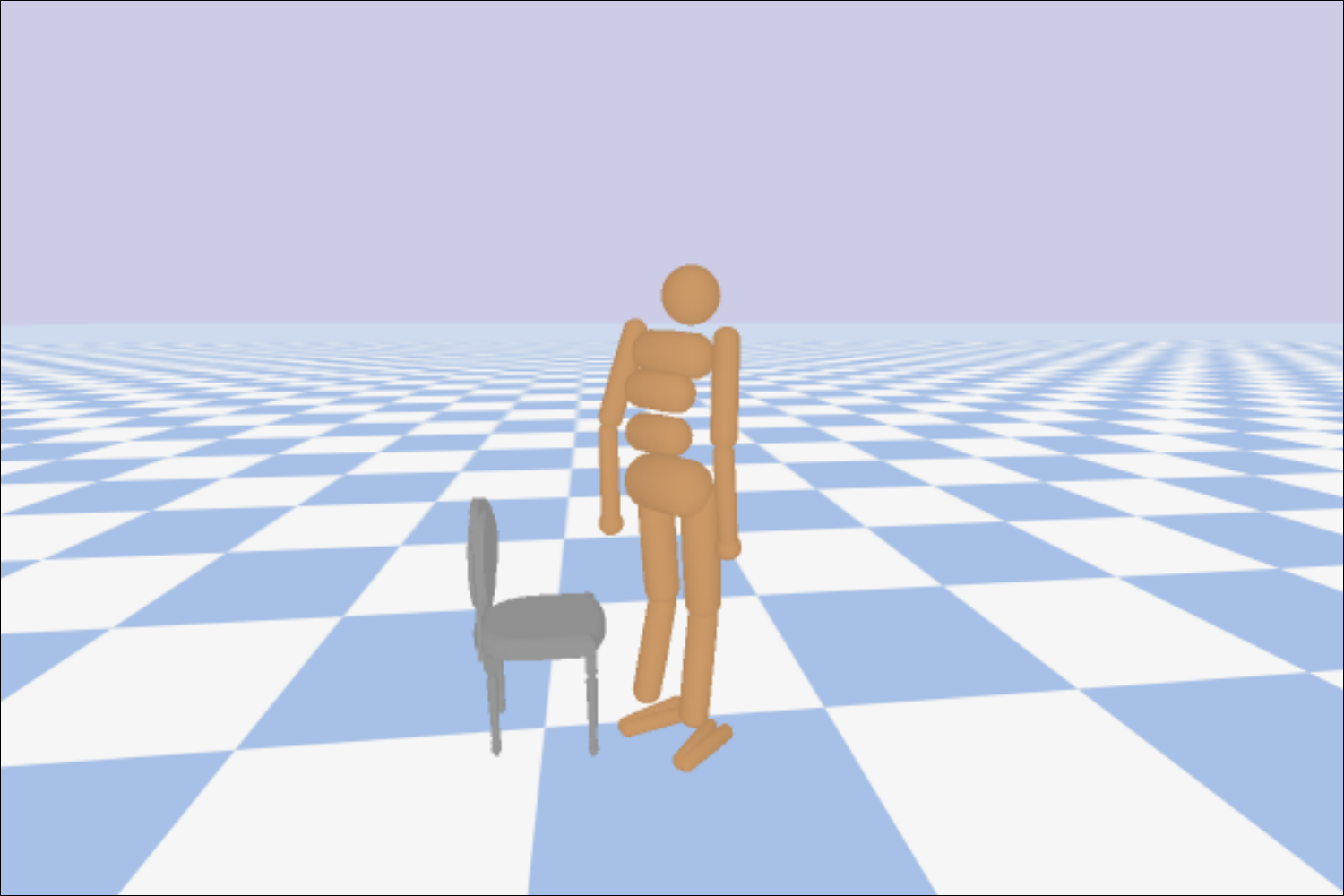} \end{minipage}
 \begin{minipage}{0.12\textwidth} \centering \includegraphics[width=1.00\textwidth]{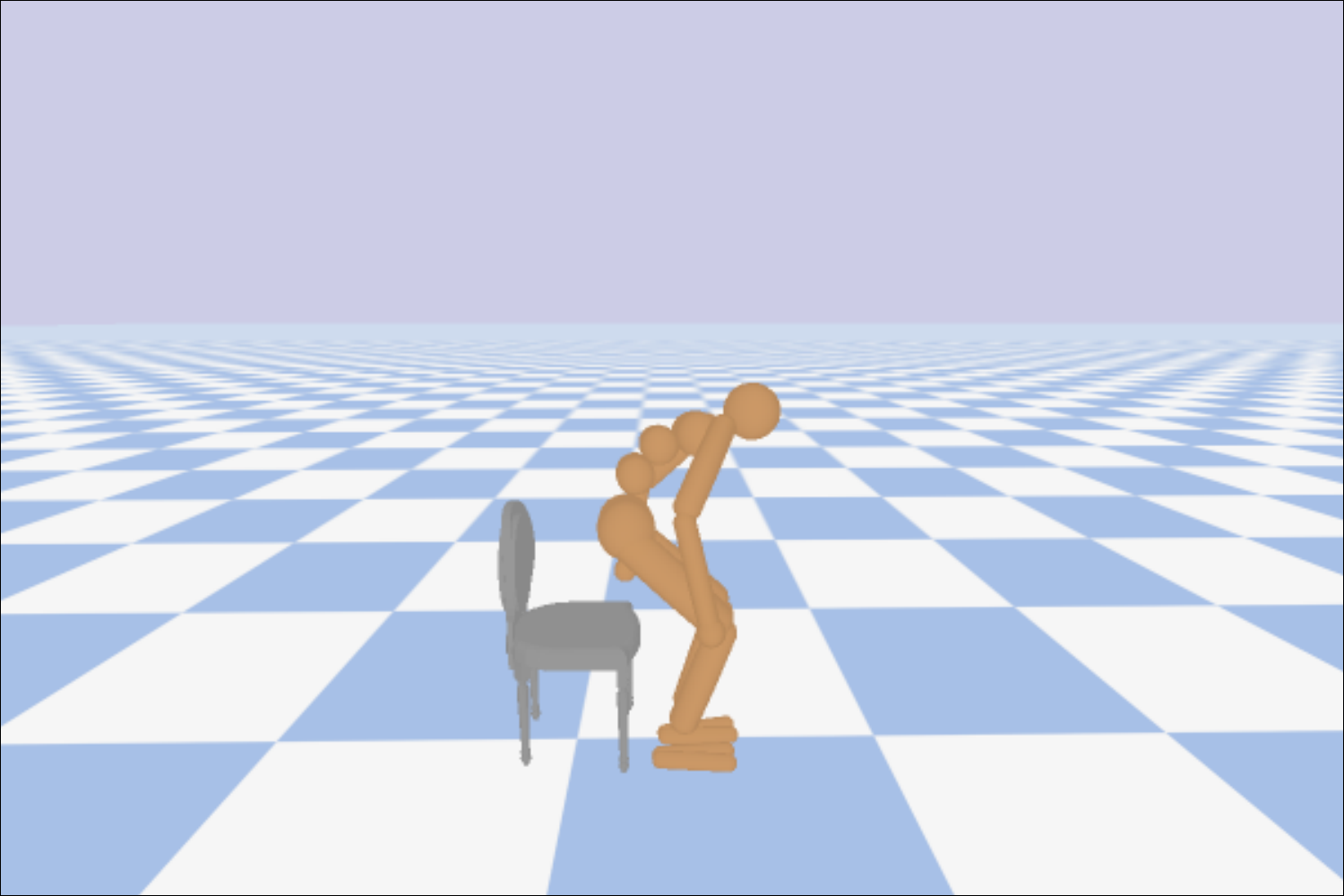} \end{minipage}
 \begin{minipage}{0.12\textwidth} \centering \includegraphics[width=1.00\textwidth]{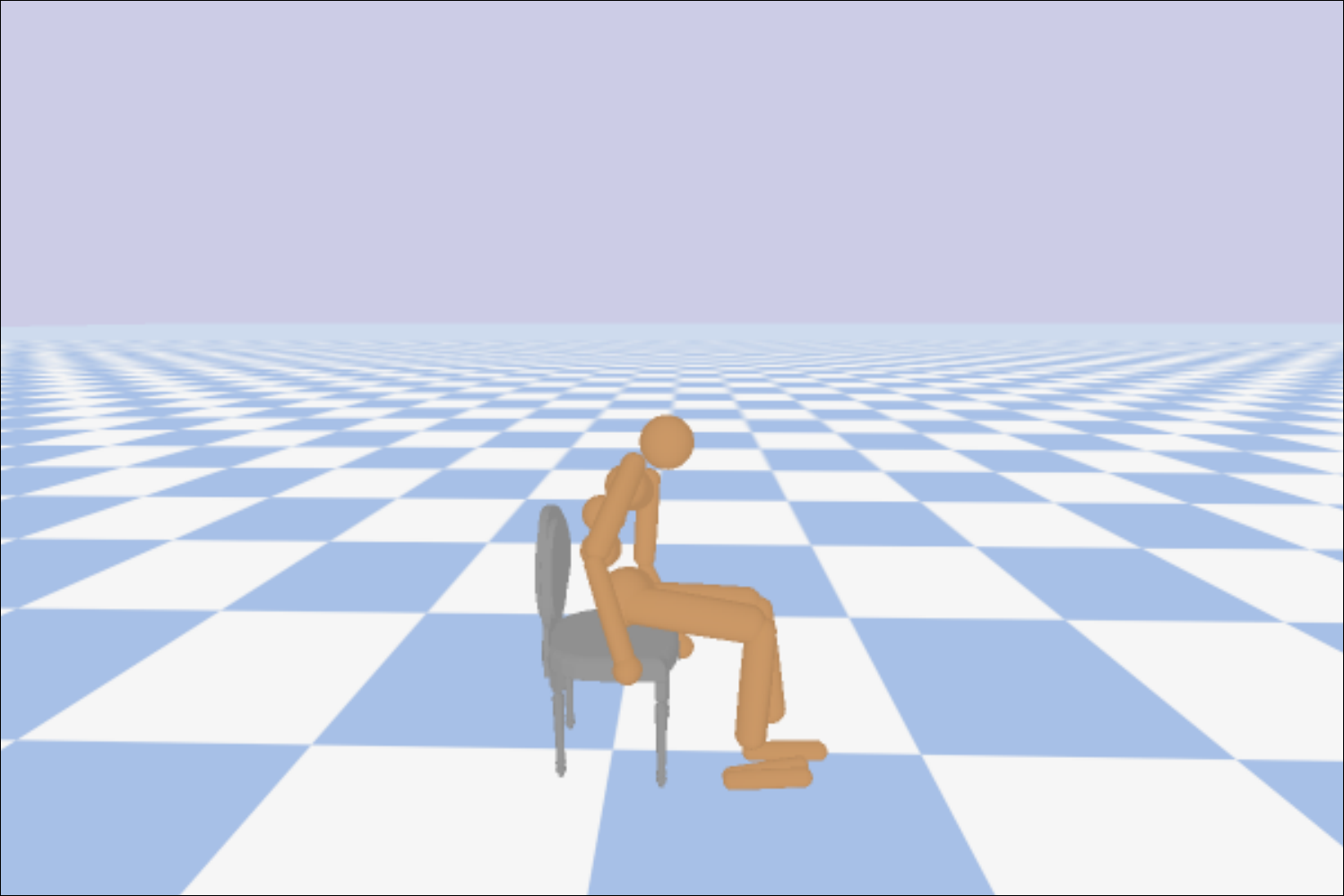} \end{minipage}
 \caption{\small Qualitative results on the Hard setting. The humanoid can sit
down successfully when starting from the back side of the chair.}
 \label{fig:qual-hard}
\end{figure*}

As shown in Tab.~\ref{tab:baselines}, hierarchical approaches outperform
non-hierarchical approaches, validating our hypothesis that hierarchical
models, by breaking a task into reusable subtasks, can attain better
generalization. Besides, our approach outperforms the pre-defined order
baselines. This is because: (1) the main task cannot always be completed by a
fixed order of subtasks, and (2) fixing the order increases training difficulty
because certain missing transitions (e.g. \textit{left
turn}$\rightarrow$\textit{walk}) are necessary for recovery from mistakes.
Finally, our full model outperforms the baselines that only allow turning in
one direction. This suggests the two turning subtasks are complementary and
being used in different scenarios, e.g. in Fig.~\ref{fig:qual-easy},
\textit{walk}$\rightarrow$\textit{right turn}$\rightarrow$\textit{sit} when
starting from the chair's right side (row 3), and
\textit{walk}$\rightarrow$\textit{left turn}$\rightarrow$\textit{sit} when
starting from the chair's left side (row 4).

\vspace{-3mm}

\paragraph{Analysis} As can be seen in Tab.~\ref{tab:baselines}, the success
rate is still low even with the full model (i.e. 31.61\%). This can be
attributed to three factors: (1) failures of subtask execution, (2) failures
due to subtask transitions, and (3) an insufficient subtask repertoire. First,
Tab.~\ref{tab:subtasks} (top) shows the success rate of individual subtasks,
where the initial pose is set to the first frame of the reference motion (i.e.
as in stage one of subtask training). We can see the execution does not always
succeed (e.g. 67.59\% for \textit{right turn}). Second, Tab.~\ref{tab:subtasks}
(bottom) shows the success rate for the same subtasks, but with the initial
pose set to the last frame of the execution of another subtask (i.e. as in
stage two of subtask training). With fine-tuning the success rate after
transitions can be significantly improved, although still not perfect. Finally,
Fig.~\ref{fig:qual-easy} (row 5) shows a failure case where the humanoid needs
a ``back up'' move when it is stuck in the state of directly confronting the
chair. Building a more diverse subtask skill set is an interesting future
research problem.

To analyze the meta controller's behavior, we look at the statistics on the
switching between subtasks. Fig.~\ref{fig:transition} shows the subtask
transition matrices when the humanoid is started either from the right or left
side of the chair. We can see that certain transitions are more favored in
certain starting areas, e.g. \textit{walk}$\rightarrow$\textit{left turn} is
favored over \textit{walk}$\rightarrow$\textit{right turn} when started from
the left side. This is in line with the earlier observation that the two
turning subtasks are complementary.

\vspace{-3mm}

\paragraph{Hard Setting} We now increase the task's difficulty by initializing
the humanoid in Zone 2 and 3 (Fig.~\ref{fig:curriculum}), and show the effect
of the proposed curriculum learning (CL) strategy. Tab.~\ref{tab:curriculum}
shows the results from different initialization zones. First, we observe a
severe drop in the success rate when the humanoid is spawned in Zone 2 and 3
(e.g. from 31.61\% to 4.05\% for ``Zone 3 w/o CL''). However, the success rate
is higher in both zones when the proposed curriculum learning strategy is
applied (e.g. from 4.05\% to 7.05\% in Zone 3). This suggests that a carefully
tailored curriculum can improve the training outcome of a challenging task.
Note that the difference in the minimum distance is less significant (e.g.
0.5549 for ``Zone 2 w/o CL' versus 0.5526 for ``Zone 2''), since without CL the
humanoid can still approach the chair, but will fail to turn and sit due to the
difficulty in learning. Fig.~\ref{fig:qual-hard} shows two successful examples
when the humanoid is spawned from the rear side of the chair. Interestingly,
the humanoid learns a slightly different behavior (e.g.
\textit{walk}$\rightarrow$\textit{sit} without \textit{turn}) compared to when
starting from the front side (row 3 and 4 in Fig.~\ref{fig:qual-easy}).

\section{Conclusion}

We address motion synthesis of an interactive task---sitting onto a chair. We
introduce a hierarchical reinforcement learning approach which relies on a
collection of subtask controllers trained to imitate reusable mocap motions,
and a meta controller trained to execute the subtasks properly to complete the
main task. We experimentally demonstrate the strength of our approach over
different non-hierarchical and hierarchical baselines, and show an application
to motion prediction given an image.

\section{Ethical Impact}

The immediate impact of our work is a new virtual environment and benchmark for
synthesizing human sitting motions. This will facilitate follow-up work in this
area. Ultimately our work will be a small step towards simulating human actions
and activities. This is the key to building virtual environments with realistic
human agents, an important tool for training collaborative robots to safely
interact and work alongside humans as mentioned in the introduction.

\section{Acknowledgements}

This work was partially supported by the National Science Foundation under
Grant No. 1734266.

\bibliography{main}

\begin{thebibliography}{46}
\providecommand{\natexlab}[1]{#1}
\providecommand{\url}[1]{\texttt{#1}}
\providecommand{\urlprefix}{URL }
\expandafter\ifx\csname urlstyle\endcsname\relax
  \providecommand{\doi}[1]{doi:\discretionary{}{}{}#1}\else
  \providecommand{\doi}{doi:\discretionary{}{}{}\begingroup
  \urlstyle{rm}\Url}\fi

\bibitem[{Agrawal and {van de Panne}(2016)}]{agrawal:siggraph2016}
Agrawal, S.; and {van de Panne}, M. 2016.
\newblock Task-based locomotion.
\newblock In \emph{SIGGRAPH}.

\bibitem[{B\"{u}tepage et~al.(2017)B\"{u}tepage, Black, Kragic, and
  Kjellstr\"{o}m}]{butepage:cvpr2017}
B\"{u}tepage, J.; Black, M.~J.; Kragic, D.; and Kjellstr\"{o}m, H. 2017.
\newblock Deep representation learning for human motion prediction and
  classification.
\newblock In \emph{CVPR}.

\bibitem[{Chang et~al.(2015)Chang, Funkhouser, Guibas, Hanrahan, Huang, Li,
  Savarese, Savva, Song, Su, Xiao, Yi, and Yu}]{chang:arxiv2015}
Chang, A.~X.; Funkhouser, T.; Guibas, L.; Hanrahan, P.; Huang, Q.; Li, Z.;
  Savarese, S.; Savva, M.; Song, S.; Su, H.; Xiao, J.; Yi, L.; and Yu, F. 2015.
\newblock Shape{N}et: An Information-Rich 3{D} Model Repository.
\newblock \emph{arXiv preprint arXiv:1512.03012} .

\bibitem[{Chao et~al.(2017)Chao, Yang, Price, Cohen, and Deng}]{chao:cvpr2017}
Chao, Y.-W.; Yang, J.; Price, B.; Cohen, S.; and Deng, J. 2017.
\newblock Forecasting Human Dynamics from Static Images.
\newblock In \emph{CVPR}.

\bibitem[{Clegg et~al.(2018)Clegg, Yu, Tan, Liu, and
  Turk}]{clegg:siggraphasia2018}
Clegg, A.; Yu, W.; Tan, J.; Liu, C.~K.; and Turk, G. 2018.
\newblock Learning to Dress: Synthesizing Human Dressing Motion via Deep
  Reinforcement Learning.
\newblock In \emph{SIGGRAPH Asia}.

\bibitem[{{CMU}(2003)}]{cmu-mocap}
{CMU}. 2003.
\newblock {CMU} {G}raphics {L}ab {M}otion {C}apture {D}atabase.
\newblock \url{http://mocap.cs.cmu.edu}.

\bibitem[{Coumans and Bai(2016--2019)}]{coumans:2019}
Coumans, E.; and Bai, Y. 2016--2019.
\newblock Py{B}ullet, a {P}ython module for physics simulation for games,
  robotics and machine learning.
\newblock \url{http://pybullet.org}.

\bibitem[{Delaitre et~al.(2012)Delaitre, Fouhey, Laptev, Sivic, Gupta, and
  Efros}]{delaitre:eccv2012}
Delaitre, V.; Fouhey, D.~F.; Laptev, I.; Sivic, J.; Gupta, A.; and Efros, A.~A.
  2012.
\newblock Scene Semantics from Long-Term Observation of People.
\newblock In \emph{ECCV}.

\bibitem[{Dhariwal et~al.(2017)Dhariwal, Hesse, Klimov, Nichol, Plappert,
  Radford, Schulman, Sidor, Wu, and Zhokhov}]{openai:baselines}
Dhariwal, P.; Hesse, C.; Klimov, O.; Nichol, A.; Plappert, M.; Radford, A.;
  Schulman, J.; Sidor, S.; Wu, Y.; and Zhokhov, P. 2017.
\newblock {O}pen{AI} {B}aselines.
\newblock \url{https://github.com/openai/baselines}.

\bibitem[{Fragkiadaki et~al.(2015)Fragkiadaki, Levine, Felsen, and
  Malik}]{fragkiadaki:iccv2015}
Fragkiadaki, K.; Levine, S.; Felsen, P.; and Malik, J. 2015.
\newblock Recurrent Network Models for Human Dynamics.
\newblock In \emph{ICCV}.

\bibitem[{Ghosh et~al.(2017)Ghosh, Song, Aksan, and Hilliges}]{ghosh:3dv2017}
Ghosh, P.; Song, J.; Aksan, E.; and Hilliges, O. 2017.
\newblock Learning Human Motion Models for Long-Term Predictions.
\newblock In \emph{3DV}.

\bibitem[{Gui et~al.(2018{\natexlab{a}})Gui, Wang, Liang, and
  Moura}]{gui:eccv2018a}
Gui, L.-Y.; Wang, Y.-X.; Liang, X.; and Moura, J. M.~F. 2018{\natexlab{a}}.
\newblock Adversarial Geometry-Aware Human Motion Prediction.
\newblock In \emph{ECCV}.

\bibitem[{Gui et~al.(2018{\natexlab{b}})Gui, Wang, Ramanan, and
  Moura}]{gui:eccv2018b}
Gui, L.-Y.; Wang, Y.-X.; Ramanan, D.; and Moura, J. M.~F. 2018{\natexlab{b}}.
\newblock Few-Shot Human Motion Prediction via Meta-Learning.
\newblock In \emph{ECCV}.

\bibitem[{Gupta et~al.(2011)Gupta, Satkin, Efros, and Hebert}]{gupta:cvpr2011}
Gupta, A.; Satkin, S.; Efros, A.~A.; and Hebert, M. 2011.
\newblock From 3{D} Scene Geometry to Human Workspace.
\newblock In \emph{CVPR}.

\bibitem[{Heess et~al.(2017)Heess, TB, Sriram, Lemmon, Merel, Wayne, Tassa,
  Erez, Wang, Eslami, Riedmiller, and Silver}]{heess:arxiv2017}
Heess, N.; TB, D.; Sriram, S.; Lemmon, J.; Merel, J.; Wayne, G.; Tassa, Y.;
  Erez, T.; Wang, Z.; Eslami, S. M.~A.; Riedmiller, M.; and Silver, D. 2017.
\newblock Emergence of Locomotion Behaviours in Rich Environments.
\newblock \emph{arXiv preprint arXiv:1707.02286} .

\bibitem[{Heess et~al.(2016)Heess, Wayne, Tassa, Lillicrap, Riedmiller, and
  Silver}]{heess:arxiv2016}
Heess, N.; Wayne, G.; Tassa, Y.; Lillicrap, T.; Riedmiller, M.; and Silver, D.
  2016.
\newblock Learning and Transfer of Modulated Locomotor Controllers.
\newblock \emph{arXiv preprint arXiv:1610.05182} .

\bibitem[{Holden, Komura, and Saito(2017)}]{holden:siggraph2017}
Holden, D.; Komura, T.; and Saito, J. 2017.
\newblock Phase-Functioned Neural Networks for Character Control.
\newblock In \emph{SIGGRAPH}.

\bibitem[{Holden, Saito, and Komura(2016)}]{holden:siggraph2016}
Holden, D.; Saito, J.; and Komura, T. 2016.
\newblock A Deep Learning Framework for Character Motion Synthesis and Editing.
\newblock In \emph{SIGGRAPH}.

\bibitem[{Huang et~al.(2018)Huang, Qi, Zhu, Xiao, Xu, and Zhu}]{huang:eccv2018}
Huang, S.; Qi, S.; Zhu, Y.; Xiao, Y.; Xu, Y.; and Zhu, S.-C. 2018.
\newblock Holistic 3{D} Scene Parsing and Reconstruction from a Single {RGB}
  Image.
\newblock In \emph{ECCV}.

\bibitem[{Jain et~al.(2016)Jain, Zamir, Savarese, and Saxena}]{jain:cvpr2016}
Jain, A.; Zamir, A.~R.; Savarese, S.; and Saxena, A. 2016.
\newblock Structural-{RNN}: Deep Learning on Spatio-Temporal Graphs.
\newblock In \emph{CVPR}.

\bibitem[{Kovar, Gleicher, and Pighin(2002)}]{kovar:siggraph2002}
Kovar, L.; Gleicher, M.; and Pighin, F. 2002.
\newblock Motion Graphs.
\newblock In \emph{SIGGRAPH}.

\bibitem[{Kulkarni et~al.(2016)Kulkarni, Narasimhan, Saeedi, and
  Tenenbaum}]{kulkarni:nips2016}
Kulkarni, T.~D.; Narasimhan, K.; Saeedi, A.; and Tenenbaum, J. 2016.
\newblock Hierarchical Deep Reinforcement Learning: Integrating Temporal
  Abstraction and Intrinsic Motivation.
\newblock In \emph{NIPS}.

\bibitem[{Li et~al.(2018)Li, Zhang, Lee, and Lee}]{li:cvpr2018}
Li, C.; Zhang, Z.; Lee, W.~S.; and Lee, G.~H. 2018.
\newblock Convolutional Sequence to Sequence Model for Human Dynamics.
\newblock In \emph{CVPR}.

\bibitem[{Liu and Hodgins(2017)}]{liu:tog2017}
Liu, L.; and Hodgins, J. 2017.
\newblock Learning to Schedule Control Fragments for Physics-Based Characters
  Using Deep {Q}-Learning.
\newblock \emph{ToG} 36(3).

\bibitem[{Liu and Hodgins(2018)}]{liu:siggraph2018}
Liu, L.; and Hodgins, J. 2018.
\newblock Learning Basketball Dribbling Skills Using Trajectory Optimization
  and Deep Reinforcement Learning.
\newblock In \emph{SIGGRAPH}.

\bibitem[{Liu, {van de Panne}, and Yin(2016)}]{liu:tog2016}
Liu, L.; {van de Panne}, M.; and Yin, K. 2016.
\newblock Guided Learning of Control Graphs for Physics-Based Characters.
\newblock \emph{ToG} 35(3): 29:1--29:14.

\bibitem[{Martinez et~al.(2017)Martinez, Black, , and
  Romero}]{martinez:cvpr2017}
Martinez, J.; Black, M.~J.; ; and Romero, J. 2017.
\newblock On human motion prediction using recurrent neural networks.
\newblock In \emph{CVPR}.

\bibitem[{Merel et~al.(2019{\natexlab{a}})Merel, Ahuja, Pham, Tunyasuvunakool,
  Liu, Tirumala, Heess, and Wayne}]{merel:iclr2019a}
Merel, J.; Ahuja, A.; Pham, V.; Tunyasuvunakool, S.; Liu, S.; Tirumala, D.;
  Heess, N.; and Wayne, G. 2019{\natexlab{a}}.
\newblock Hierarchical Visuomotor Control of Humanoids.
\newblock In \emph{ICLR}.

\bibitem[{Merel et~al.(2019{\natexlab{b}})Merel, Hasenclever, Galashov, Ahuja,
  Pham, Wayne, Teh, and Heess}]{merel:iclr2019b}
Merel, J.; Hasenclever, L.; Galashov, A.; Ahuja, A.; Pham, V.; Wayne, G.; Teh,
  Y.~W.; and Heess, N. 2019{\natexlab{b}}.
\newblock Neural Probabilistic Motor Primitives for Humanoid Control.
\newblock In \emph{ICLR}.

\bibitem[{Merel et~al.(2017)Merel, Tassa, TB, Srinivasan, Lemmon, Wang, Wayne,
  and Heess}]{merel:arxiv2017}
Merel, J.; Tassa, Y.; TB, D.; Srinivasan, S.; Lemmon, J.; Wang, Z.; Wayne, G.;
  and Heess, N. 2017.
\newblock Learning human behaviors from motion capture by adversarial
  imitation.
\newblock \emph{arXiv preprint arXiv:1707.02201} .

\bibitem[{{O}pen{AI}(2017)}]{openai:roboschool}
{O}pen{AI}. 2017.
\newblock {O}pen{AI} {R}oboschool.
\newblock \url{https://blog.openai.com/roboschool/}.

\bibitem[{Peng et~al.(2018{\natexlab{a}})Peng, Abbeel, Levine, and {van de
  Panne}}]{peng:siggraph2018}
Peng, X.~B.; Abbeel, P.; Levine, S.; and {van de Panne}, M. 2018{\natexlab{a}}.
\newblock Deep{M}imic: Example-Guided Deep Reinforcement Learning of
  Physics-Based Character Skills.
\newblock In \emph{SIGGRAPH}.

\bibitem[{Peng et~al.(2017)Peng, Berseth, Yin, and {van de
  Panne}}]{peng:siggraph2017}
Peng, X.~B.; Berseth, G.; Yin, K.; and {van de Panne}, M. 2017.
\newblock Deep{L}oco: Developing Locomotion Skills Using Hierarchical Deep
  Reinforcement Learning.
\newblock In \emph{SIGGRAPH}.

\bibitem[{Peng et~al.(2019)Peng, Chang, Zhang, Abbeel, and
  Levine}]{peng:neurips2019}
Peng, X.~B.; Chang, M.; Zhang, G.; Abbeel, P.; and Levine, S. 2019.
\newblock {MCP}: Learning Composable Hierarchical Control with Multiplicative
  Compositional Policies.
\newblock In \emph{NeurIPS}.

\bibitem[{Peng et~al.(2018{\natexlab{b}})Peng, Kanazawa, Malik, Abbeel, and
  Levine}]{peng:siggraphasia2018}
Peng, X.~B.; Kanazawa, A.; Malik, J.; Abbeel, P.; and Levine, S.
  2018{\natexlab{b}}.
\newblock {SFV}: Reinforcement Learning of Physical Skills from Videos.
\newblock In \emph{SIGGRAPH Asia}.

\bibitem[{Schulman et~al.(2017)Schulman, Wolski, Dhariwal, Radford, and
  Klimov}]{schulman:arxiv2017}
Schulman, J.; Wolski, F.; Dhariwal, P.; Radford, A.; and Klimov, O. 2017.
\newblock Proximal Policy Optimization Algorithms.
\newblock \emph{arXiv preprint arXiv:1707.06347} .

\bibitem[{Tessler et~al.(2017)Tessler, Givony, Zahavy, Mankowitz, and
  Mannor}]{tessler:aaai2017}
Tessler, C.; Givony, S.; Zahavy, T.; Mankowitz, D.~J.; and Mannor, S. 2017.
\newblock A Deep Hierarchical Approach to Lifelong Learning in {M}inecraft.
\newblock In \emph{AAAI}.

\bibitem[{Villegas et~al.(2018)Villegas, Yang, Ceylan, and
  Lee}]{villegas:cvpr2018}
Villegas, R.; Yang, J.; Ceylan, D.; and Lee, H. 2018.
\newblock Neural Kinematic Networks for Unsupervised Motion Retargetting.
\newblock In \emph{CVPR}.

\bibitem[{Walker et~al.(2017)Walker, Marino, Gupta, and
  Hebert}]{walker:iccv2017}
Walker, J.; Marino, K.; Gupta, A.; and Hebert, M. 2017.
\newblock The Pose Knows: Video Forecasting by Generating Pose Futures.
\newblock In \emph{ICCV}.

\bibitem[{Wang, Girdhar, and Gupta(2017)}]{wang:cvpr2017}
Wang, X.; Girdhar, R.; and Gupta, A. 2017.
\newblock Binge Watching: Scaling Affordance Learning From Sitcoms.
\newblock In \emph{CVPR}.

\bibitem[{Yamane, Kuffner, and Hodgins(2004)}]{yamane:siggraph2004}
Yamane, K.; Kuffner, J.~J.; and Hodgins, J.~K. 2004.
\newblock Synthesizing Animations of Human Manipulation Tasks.
\newblock In \emph{SIGGRAPH}.

\bibitem[{Yan et~al.(2018)Yan, Rastogi, Villegas, Sunkavalli, Shechtman, Hadap,
  Yumer, and Lee}]{yan:eccv2018}
Yan, X.; Rastogi, A.; Villegas, R.; Sunkavalli, K.; Shechtman, E.; Hadap, S.;
  Yumer, E.; and Lee, H. 2018.
\newblock {MT}-{VAE}: Learning Motion Transformations to Generate Multimodal
  Human Dynamics.
\newblock In \emph{ECCV}.

\bibitem[{Yao et~al.(2018)Yao, Wang, Ni, Wei, and Yang}]{yao:cvpr2018}
Yao, T.; Wang, M.; Ni, B.; Wei, H.; and Yang, X. 2018.
\newblock Multiple Granularity Group Interaction Prediction.
\newblock In \emph{CVPR}.

\bibitem[{Zaremba and Sutskever(2014)}]{zaremba:arxiv2014}
Zaremba, W.; and Sutskever, I. 2014.
\newblock Learning to Execute.
\newblock \emph{arXiv preprint arXiv:1410.4615} .

\bibitem[{Zhou et~al.(2018)Zhou, Li, Xiao, He, Huang, and Li}]{zhou:iclr2018}
Zhou, Y.; Li, Z.; Xiao, S.; He, C.; Huang, Z.; and Li, H. 2018.
\newblock Auto-Conditioned Recurrent Networks for Extended Complex Human Motion
  Synthesis.
\newblock In \emph{ICLR}.

\bibitem[{Zhu, Zhao, and Zhu(2015)}]{zhu:cvpr2015}
Zhu, Y.; Zhao, Y.; and Zhu, S.-C. 2015.
\newblock Understanding Tools: Task-Oriented Object Modeling, Learning and
  Recognition.
\newblock In \emph{CVPR}.

\end{thebibliography}


\appendix

\appendixpage

\section{Subtask Controller Training}

We apply the early termination strategy~\cite{peng:siggraph2018}: an episode is
terminated immediately if the height of the root falls below 0.78 meters for
\textit{walk} and \textit{turn}, and 0.54 meters for \textit{sit}. These
thresholds are chosen according to the height of the humanoid. For
\textit{turn}, the episode is also terminated when the root's yaw angle differs
from the reference motion for more than \ang{45}. For \textit{walk}, we adopt
the two-stage training strategy described in the Subtask Controller section of the paper. In
target-directed walking, we randomly sample a new 2D target in the front of the
humanoid every 2.5 seconds or when the target is reached. For \textit{sit}, the
chair is placed at a fixed location behind the humanoid, and we use reference
state initialization~\cite{peng:siggraph2018} to facilitate training.

\section{Additional Data and Implementation Details}

For each subtask, Tab.~\ref{tab:mocap} shows the associated mocap clip we used
from the CMU Graphics Lab Motion Capture Database~\cite{cmu-mocap}.
Tab.~\ref{tab:hyperparameter} shows the hyerparamters we used for the PPO
training.

\begin{table}[t]
 \centering
 \small
 \begin{tabular}{l||cc}
   \hline \TBstrut
   Subtask                             & Subject \# & Trial \# \\
   \hline \Tstrut
   \textit{walk}                       & 8          & 1, 4     \\
   \textit{left} / \textit{right turn} & 69         & 13       \\ \Bstrut
   \textit{sit}                        & 143        & 18       \\
   \hline
 \end{tabular}
 \caption{\small Mocap clips adopted from the CMU MoCap database~\cite{cmu-mocap}.}
 \label{tab:mocap}
\end{table}

\begin{table}[t]
 \centering
 \small
 \begin{tabular}{l||cc}
   \hline \TBstrut
   ~                     & Subtasks         & Meta Task        \\
   \hline \Tstrut
   \texttt{nsteps}       & 8192             & 64               \\
   \texttt{nminibatches} & 32               & 8                \\
   \texttt{noptepochs}   & 4                & 2                \\ \Bstrut
   \texttt{lr}           & $1\times10^{-4}$ & $1\times10^{-4}$ \\
   \hline
 \end{tabular}
 \caption{\small Hyperparamters for PPO training.}
 \label{tab:hyperparameter}
\end{table}

\begin{figure}[t!]
 \centering
 \begin{minipage}{0.114\textwidth} \centering \includegraphics[width=1.00\textwidth,trim={80 40 80 40},clip]{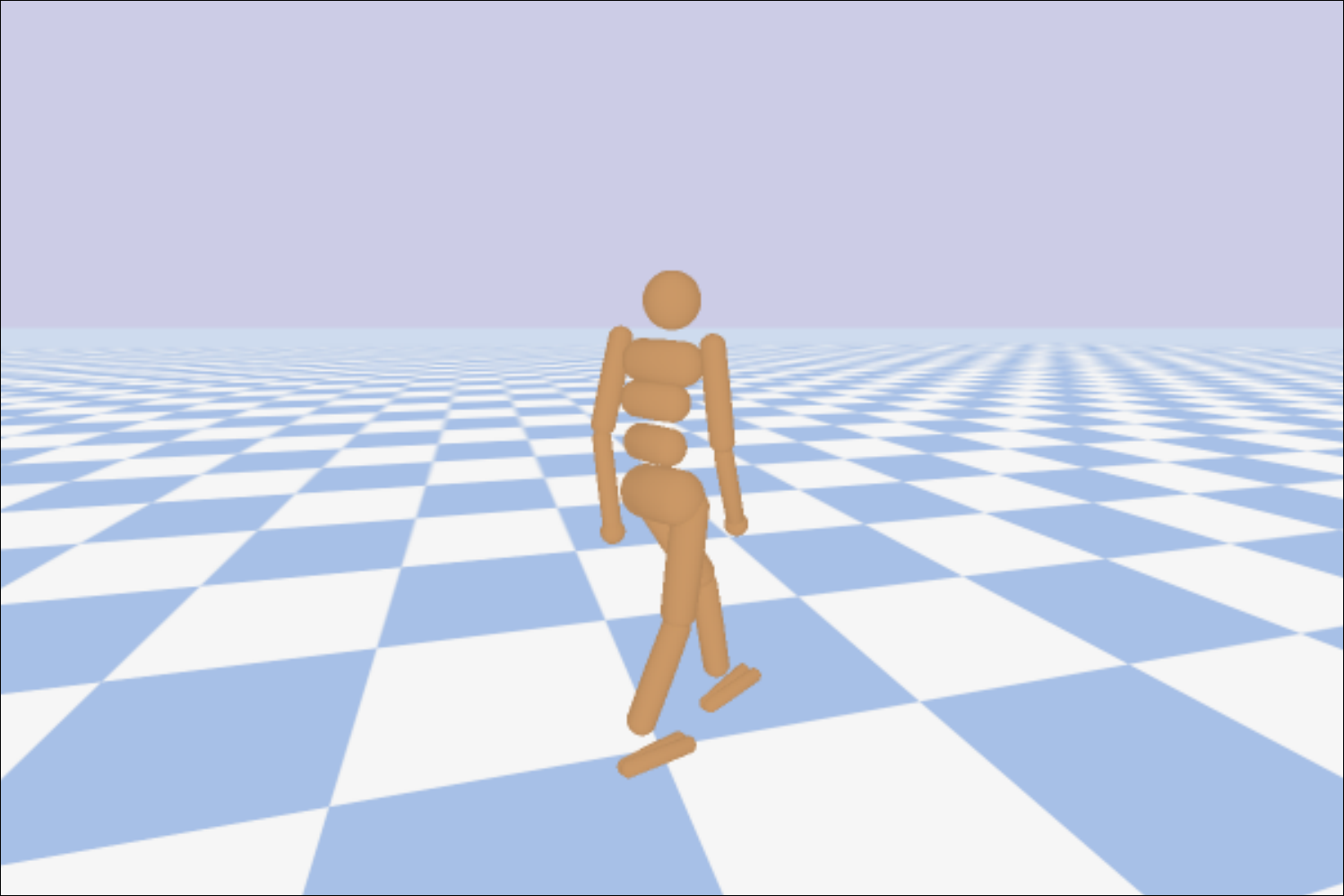} \end{minipage}
 \begin{minipage}{0.114\textwidth} \centering \includegraphics[width=1.00\textwidth,trim={80 40 80 40},clip]{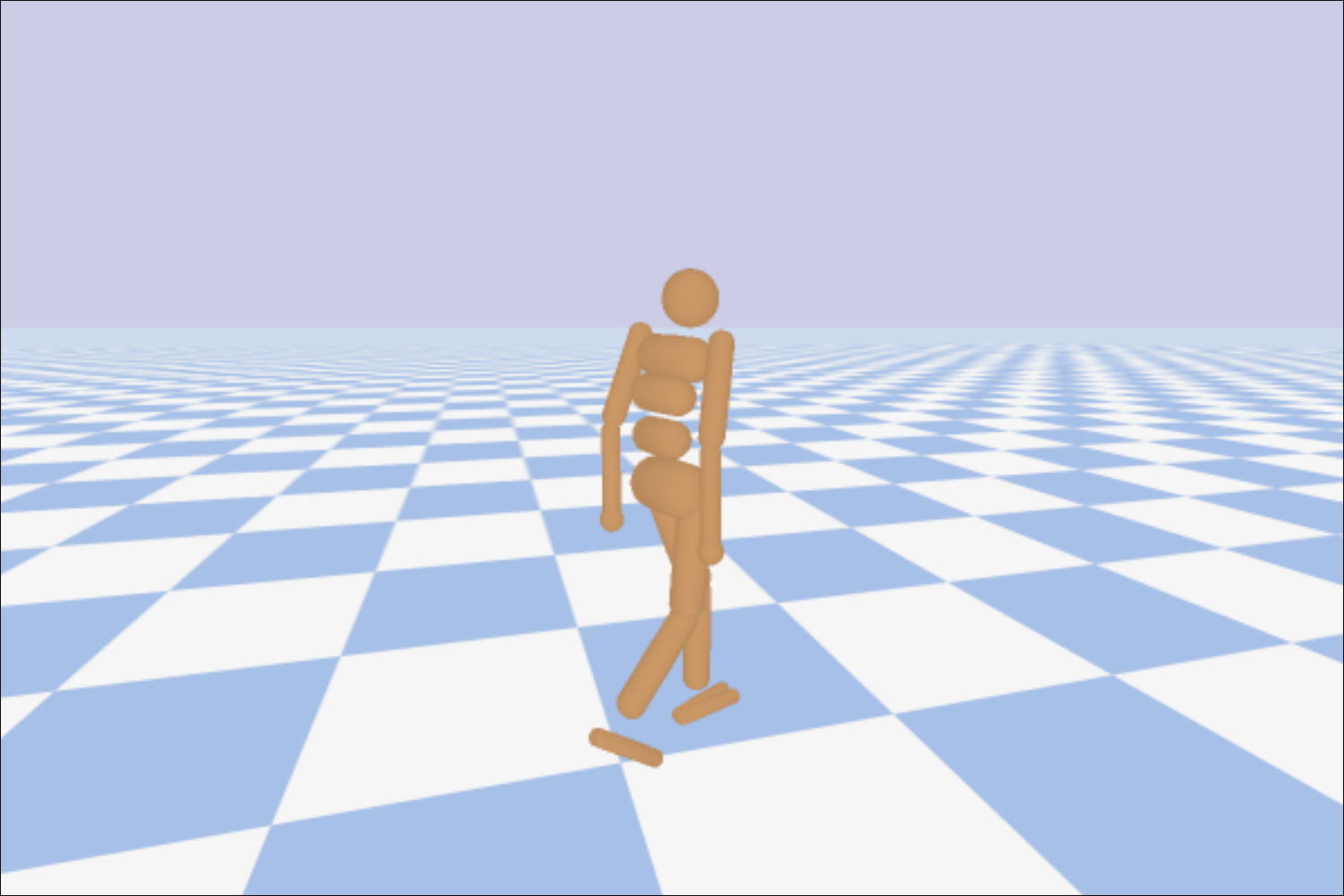} \end{minipage}
 \begin{minipage}{0.114\textwidth} \centering \includegraphics[width=1.00\textwidth,trim={80 40 80 40},clip]{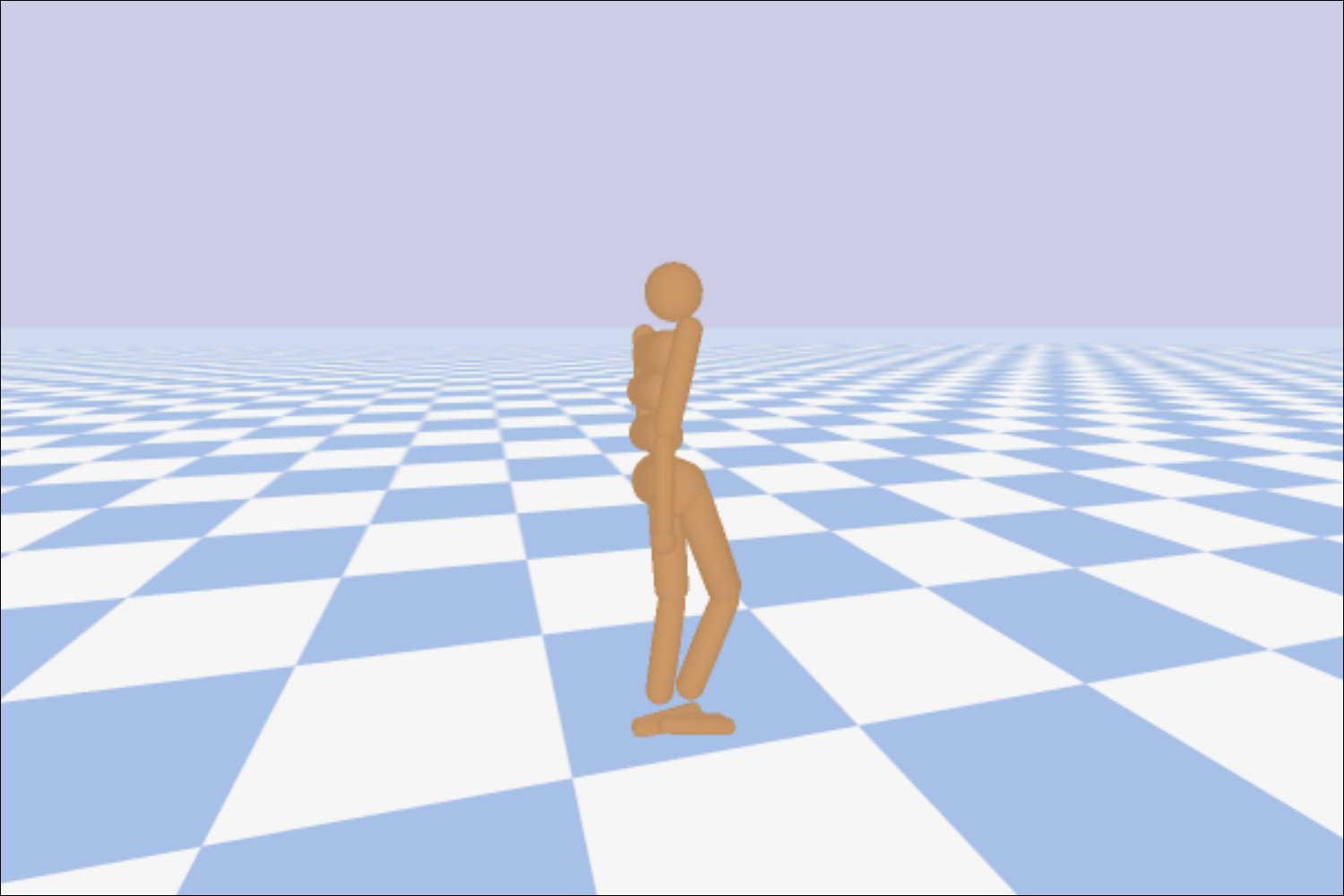} \end{minipage}
 \begin{minipage}{0.114\textwidth} \centering \includegraphics[width=1.00\textwidth,trim={80 40 80 40},clip]{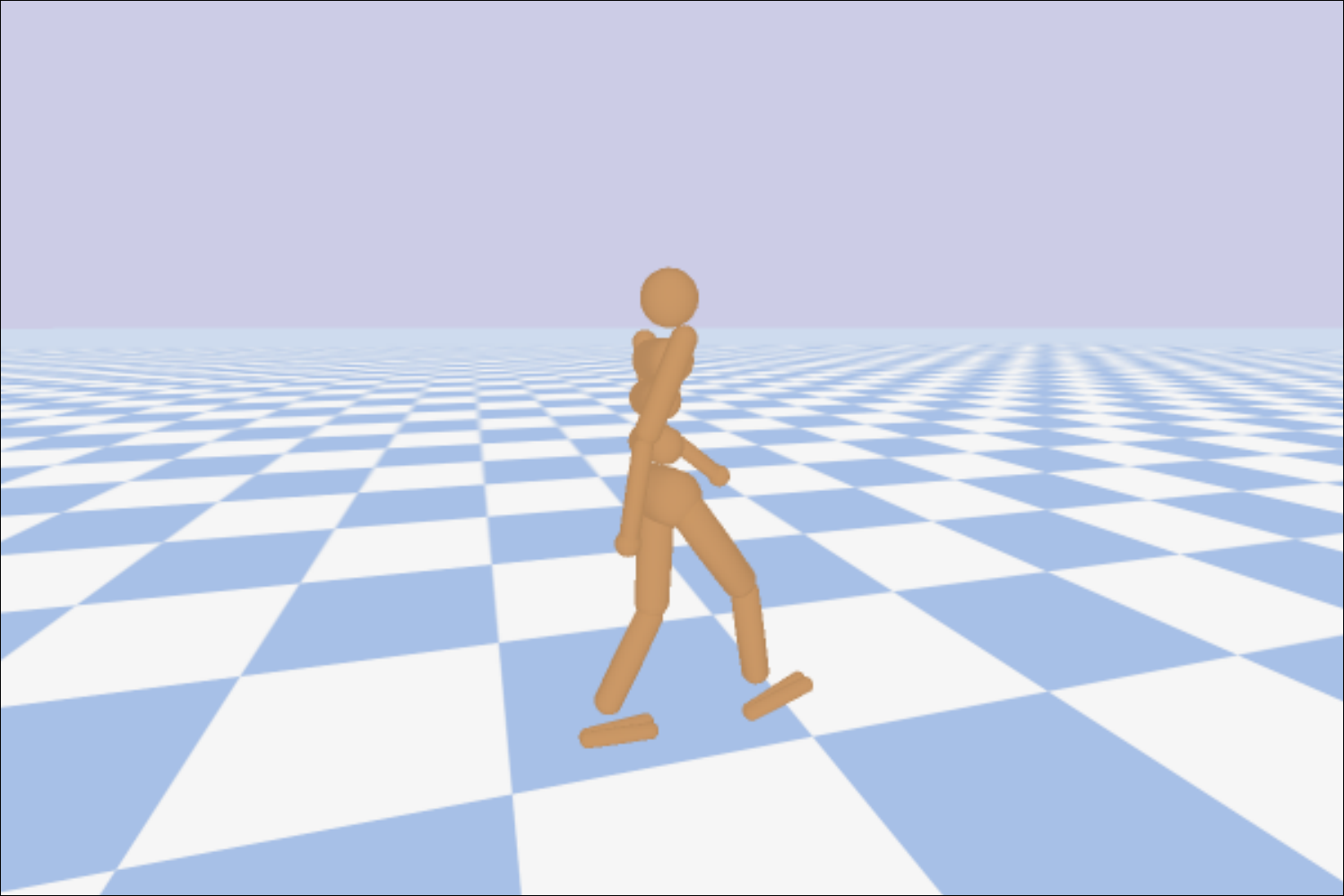} \end{minipage}
 \\ \vspace{1mm}
 \begin{minipage}{0.114\textwidth} \centering \includegraphics[width=1.00\textwidth,trim={80 40 80 40},clip]{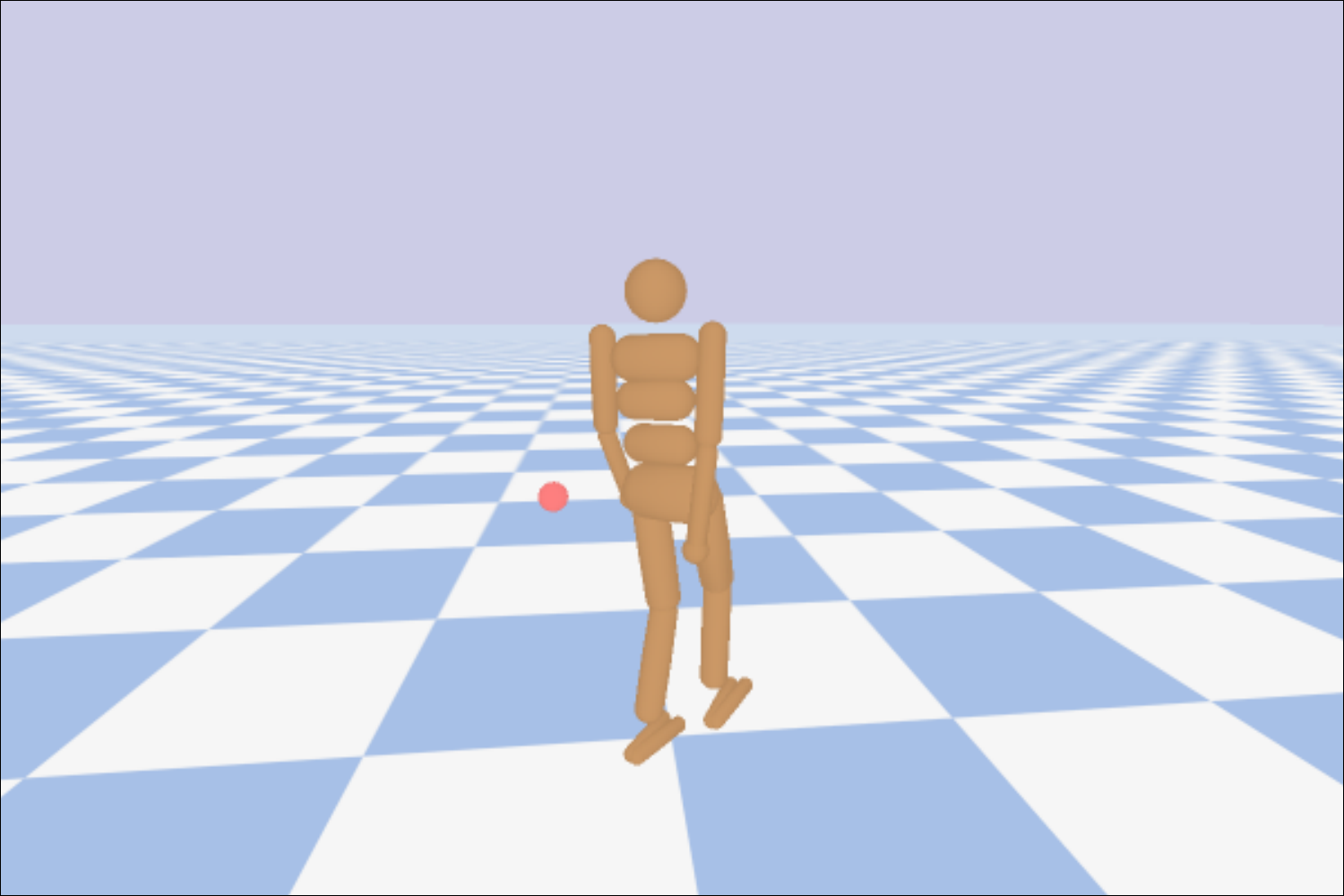} \end{minipage}
 \begin{minipage}{0.114\textwidth} \centering \includegraphics[width=1.00\textwidth,trim={80 40 80 40},clip]{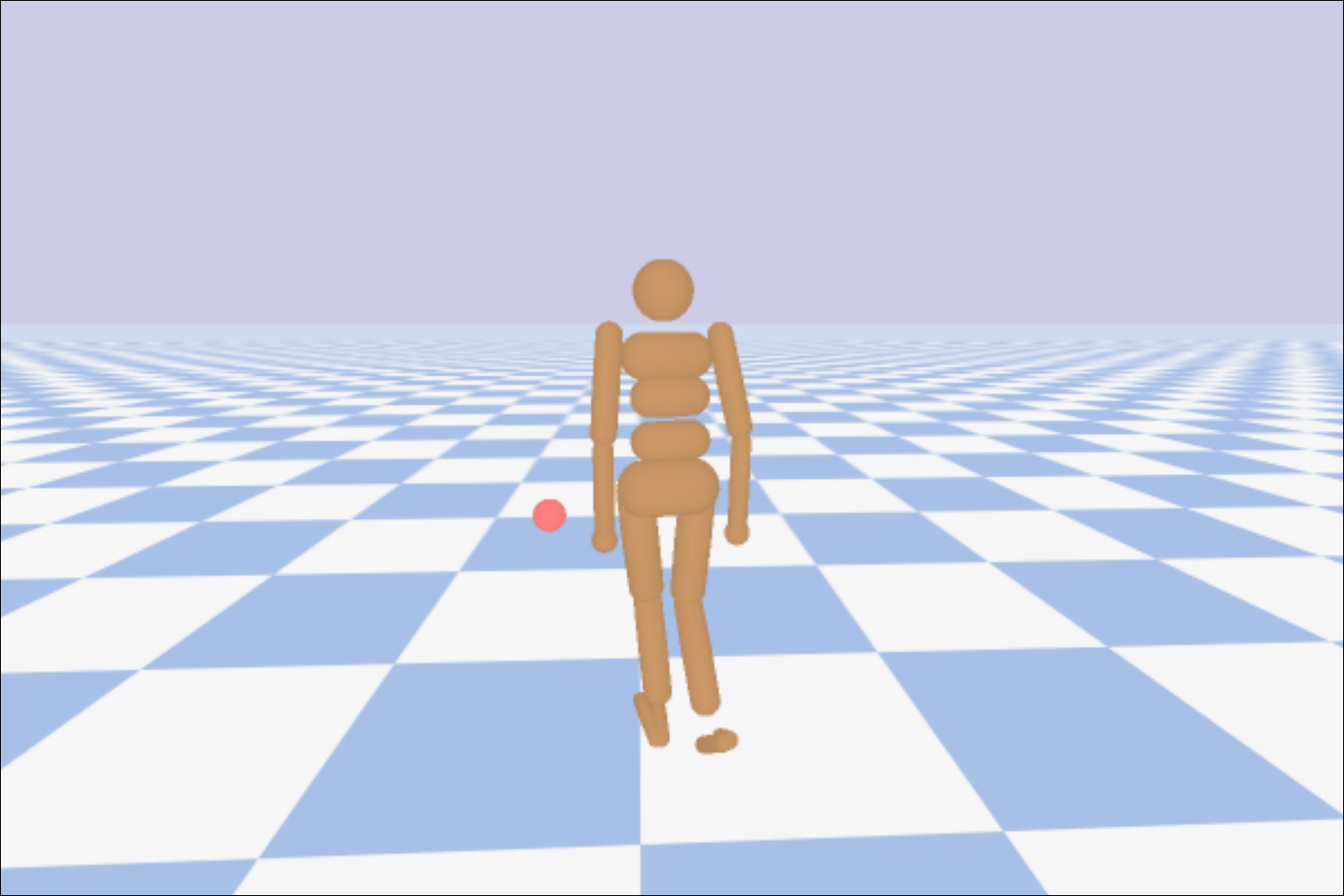} \end{minipage}
 \begin{minipage}{0.114\textwidth} \centering \includegraphics[width=1.00\textwidth,trim={80 40 80 40},clip]{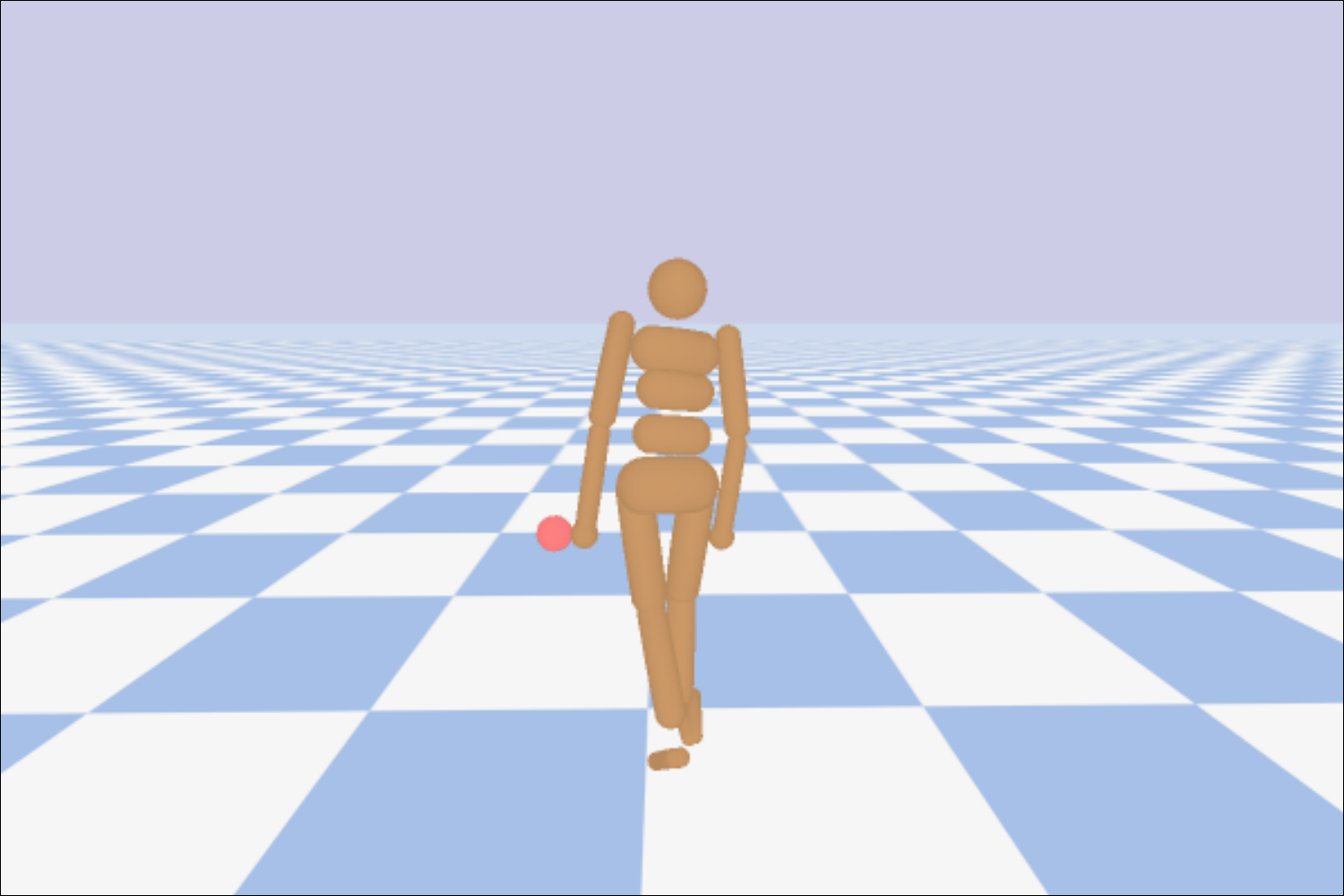} \end{minipage}
 \begin{minipage}{0.114\textwidth} \centering \includegraphics[width=1.00\textwidth,trim={80 40 80 40},clip]{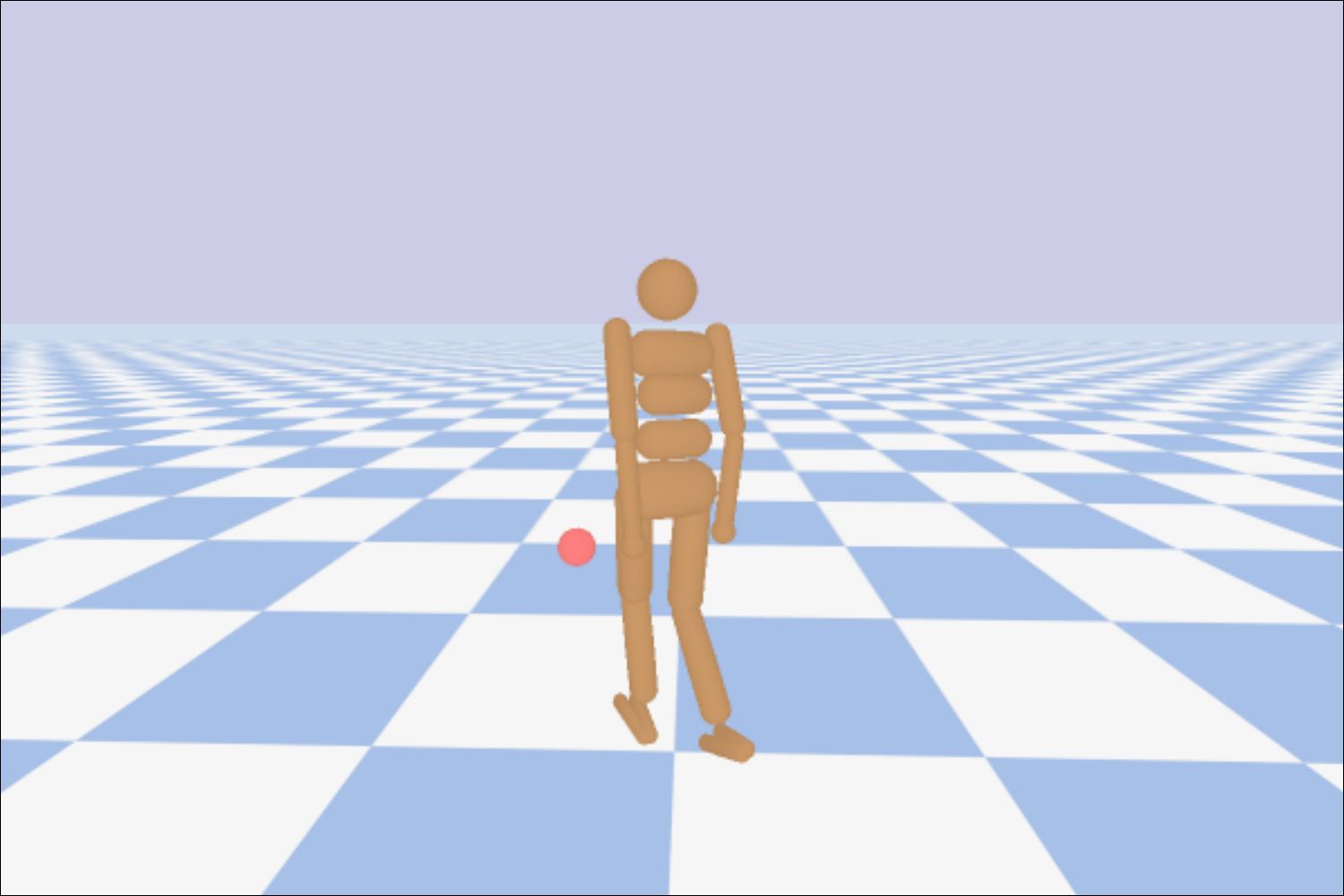} \end{minipage}
 \\ \vspace{1mm}
 \begin{minipage}{0.114\textwidth} \centering \includegraphics[width=1.00\textwidth,trim={80 40 80 40},clip]{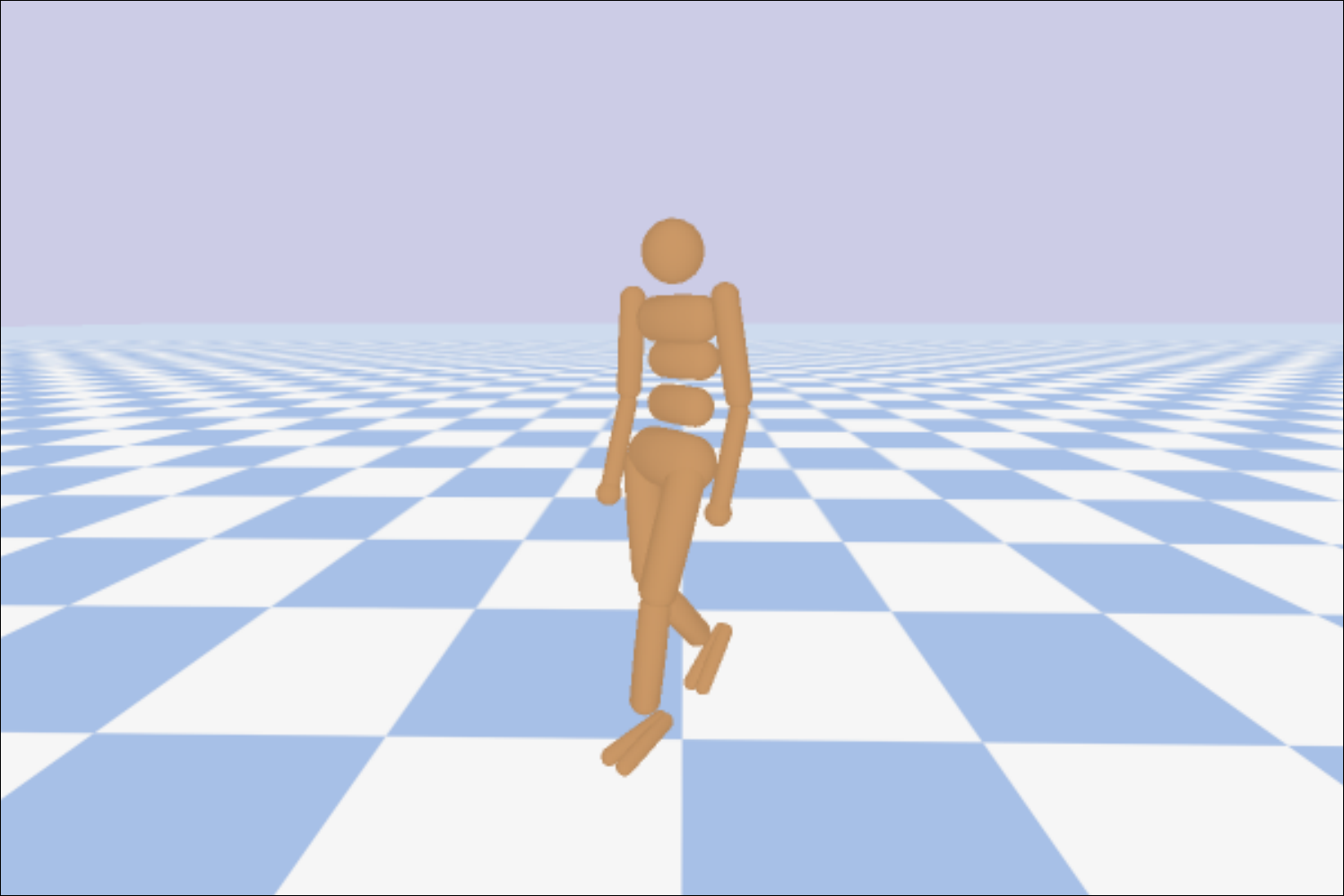} \end{minipage}
 \begin{minipage}{0.114\textwidth} \centering \includegraphics[width=1.00\textwidth,trim={80 40 80 40},clip]{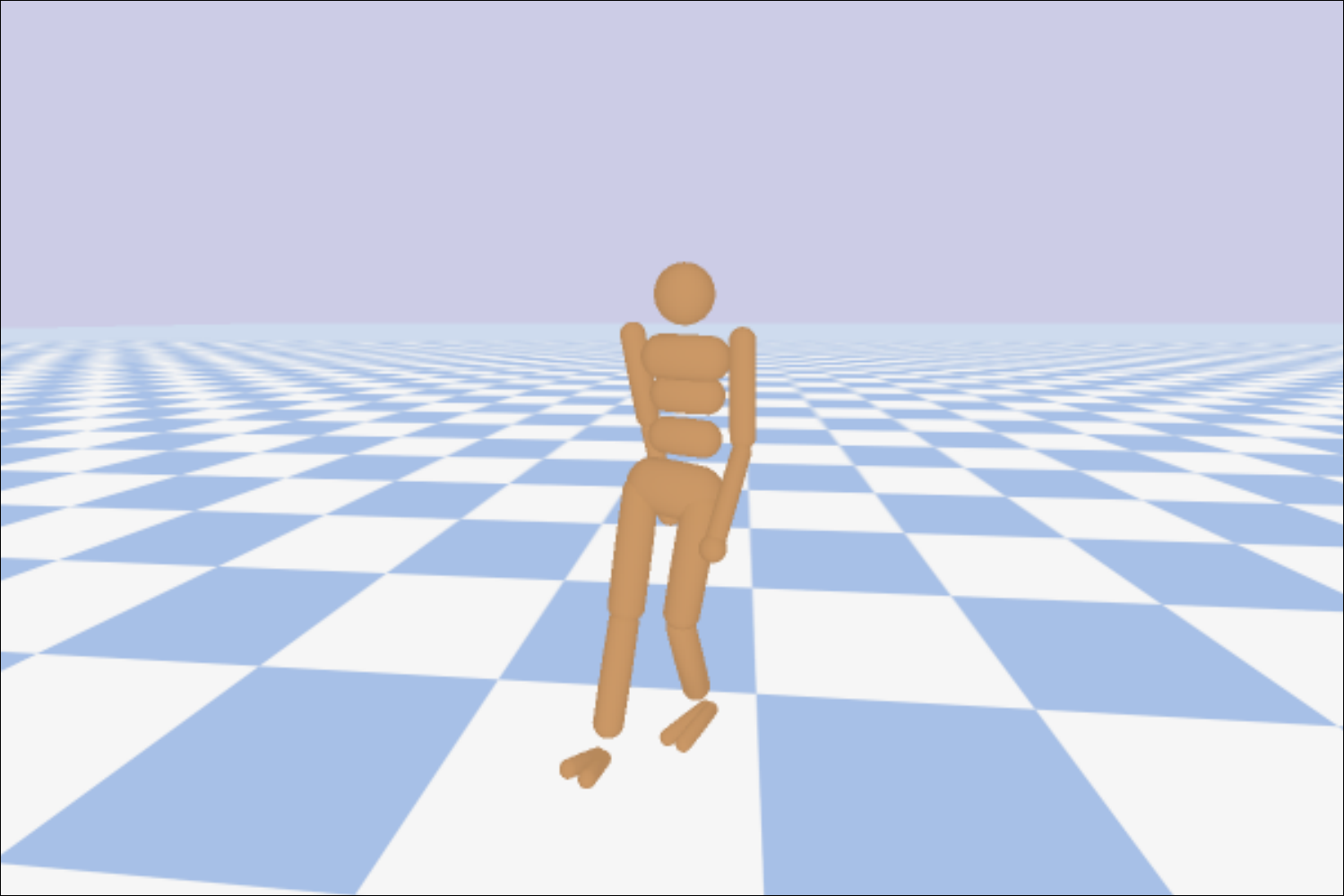} \end{minipage}
 \begin{minipage}{0.114\textwidth} \centering \includegraphics[width=1.00\textwidth,trim={80 40 80 40},clip]{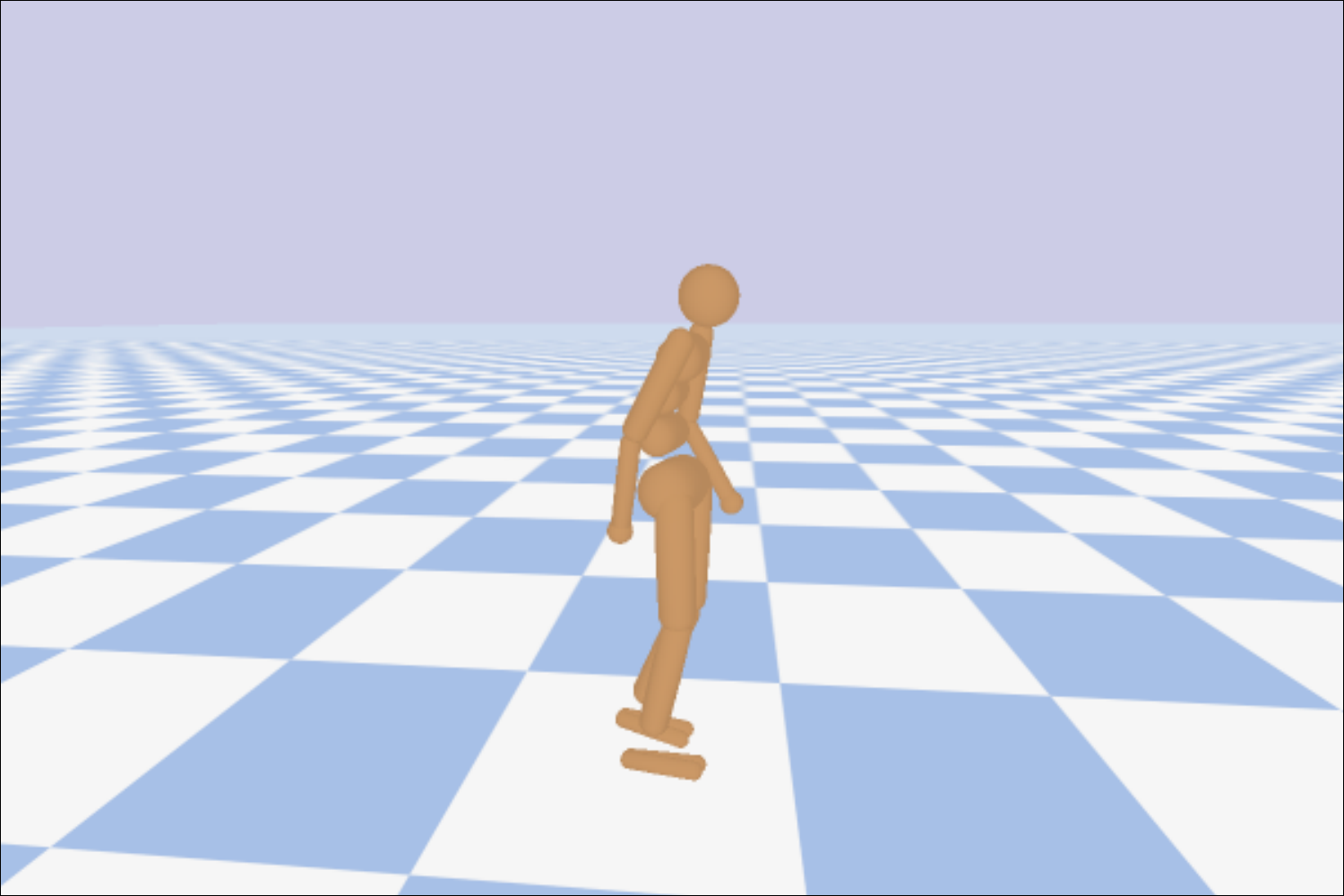} \end{minipage}
 \begin{minipage}{0.114\textwidth} \centering \includegraphics[width=1.00\textwidth,trim={80 40 80 40},clip]{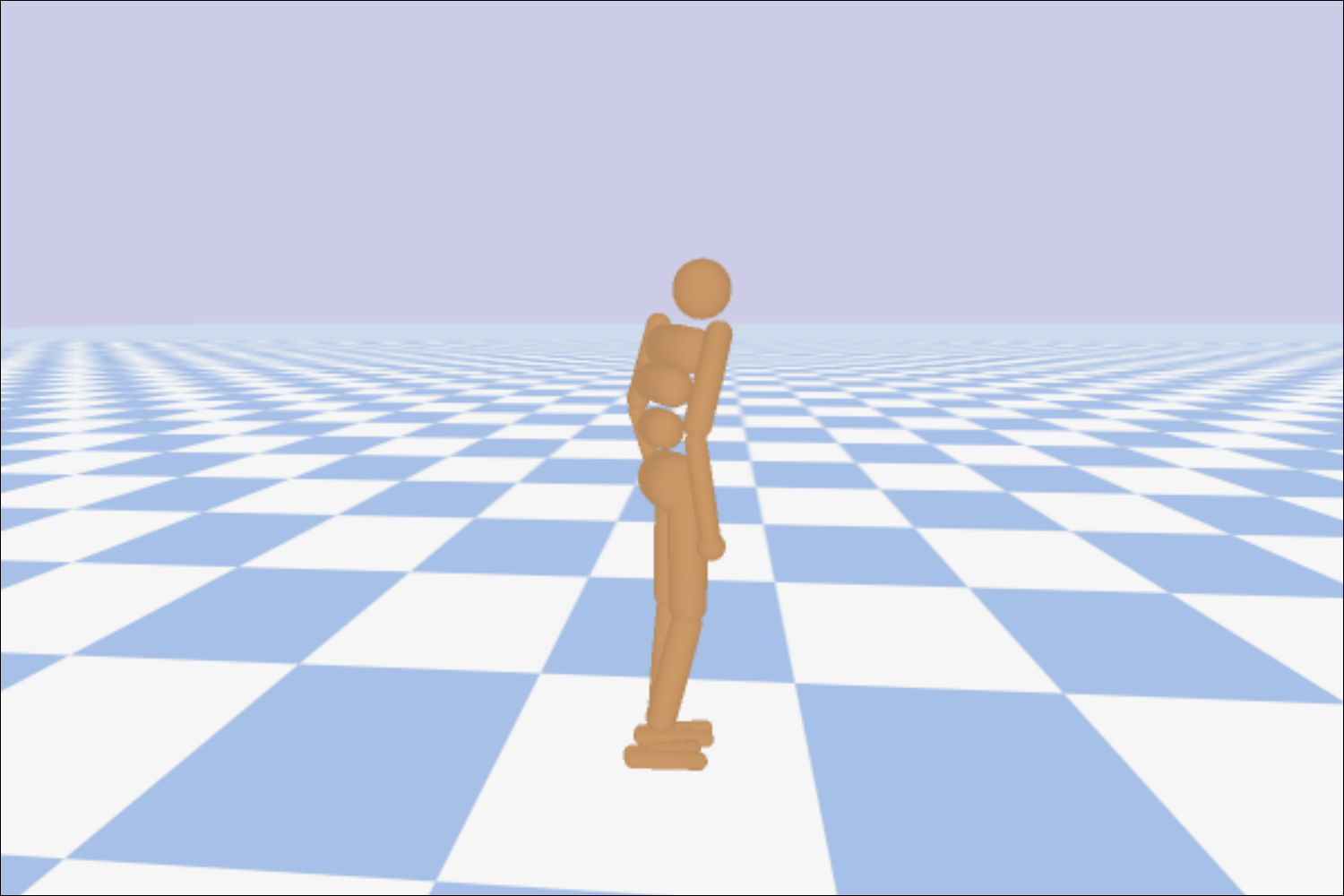} \end{minipage}
 \\ \vspace{1mm}
 \begin{minipage}{0.114\textwidth} \centering \includegraphics[width=1.00\textwidth,trim={80 40 80 40},clip]{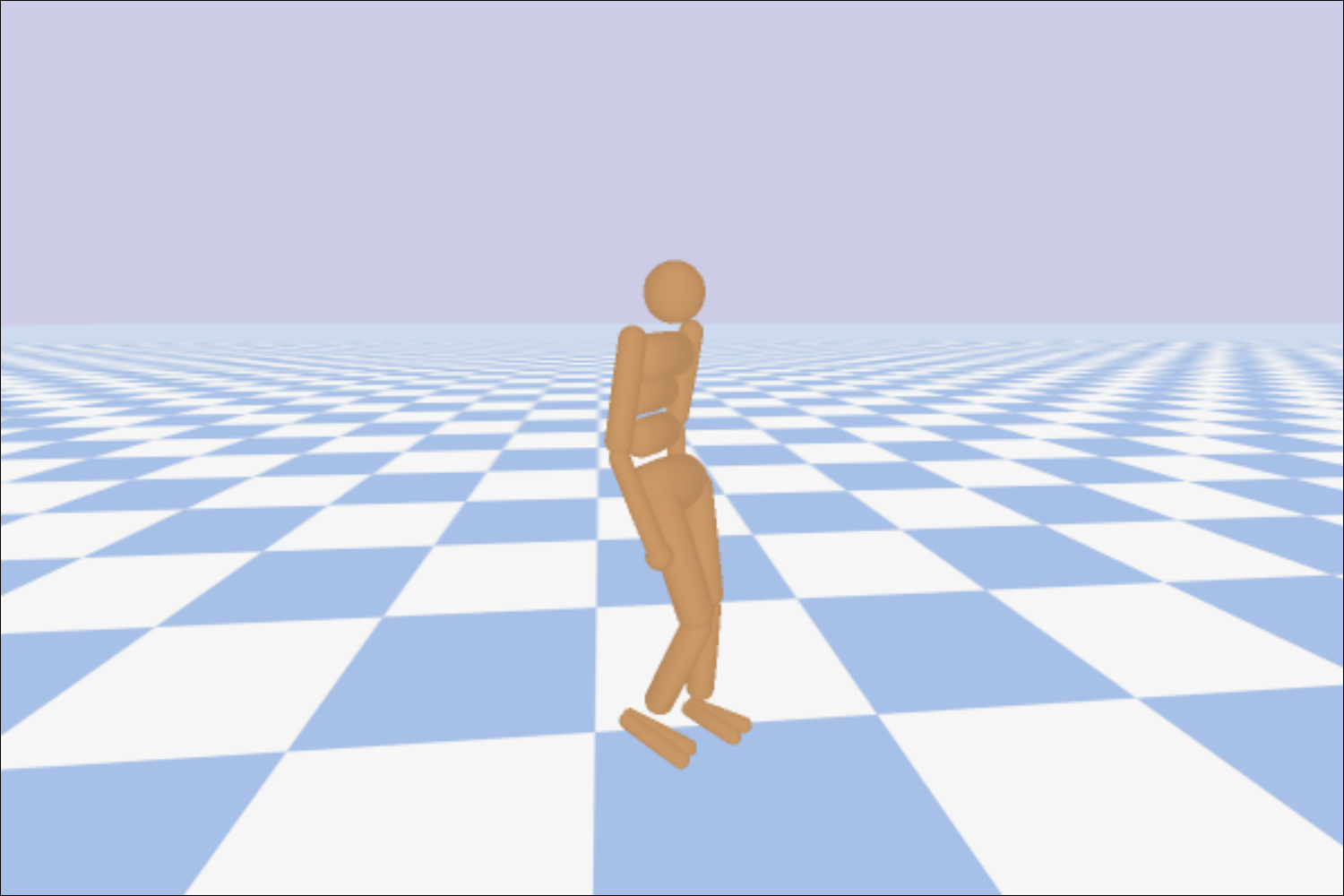} \end{minipage}
 \begin{minipage}{0.114\textwidth} \centering \includegraphics[width=1.00\textwidth,trim={80 40 80 40},clip]{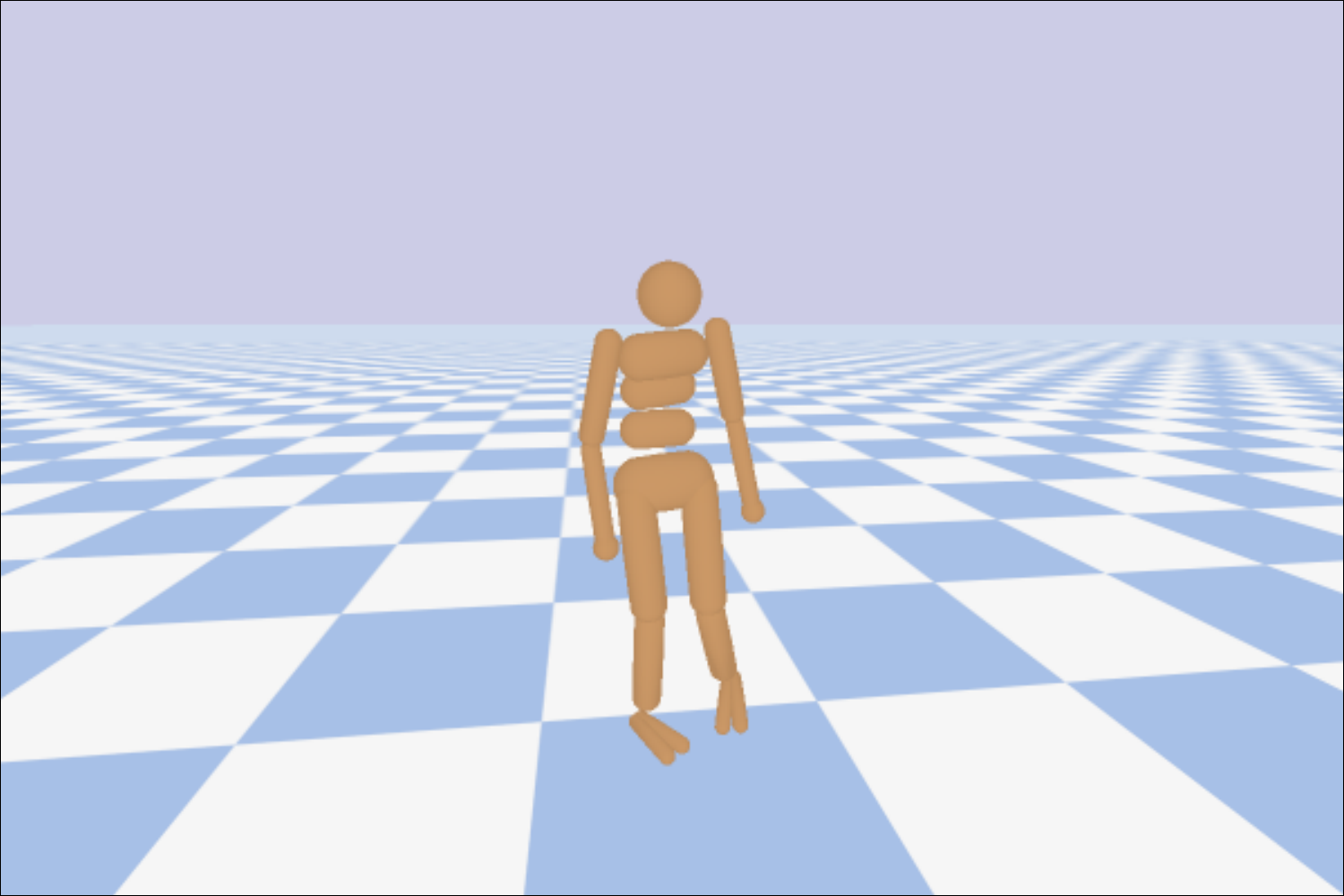} \end{minipage}
 \begin{minipage}{0.114\textwidth} \centering \includegraphics[width=1.00\textwidth,trim={80 40 80 40},clip]{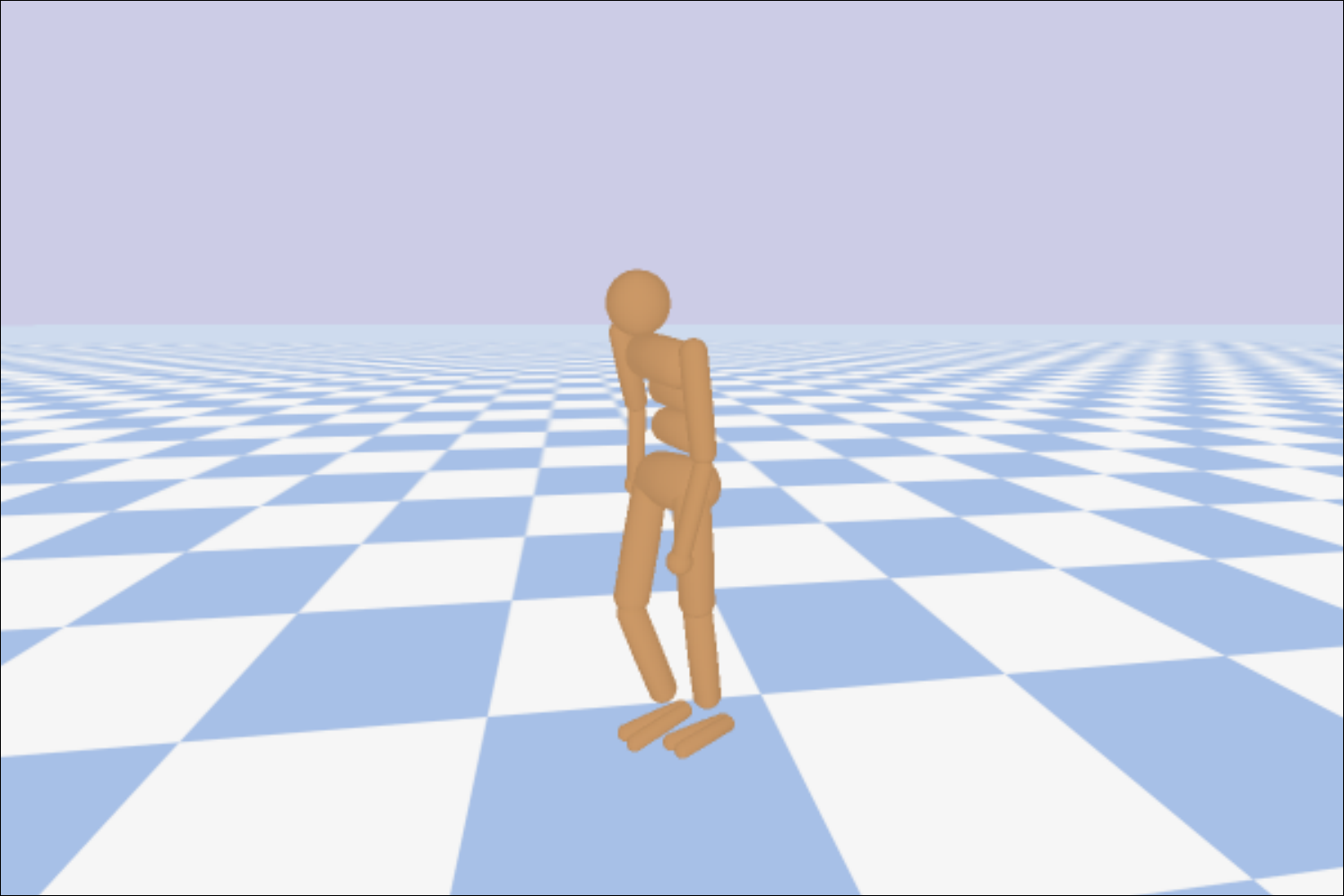} \end{minipage}
 \begin{minipage}{0.114\textwidth} \centering \includegraphics[width=1.00\textwidth,trim={80 40 80 40},clip]{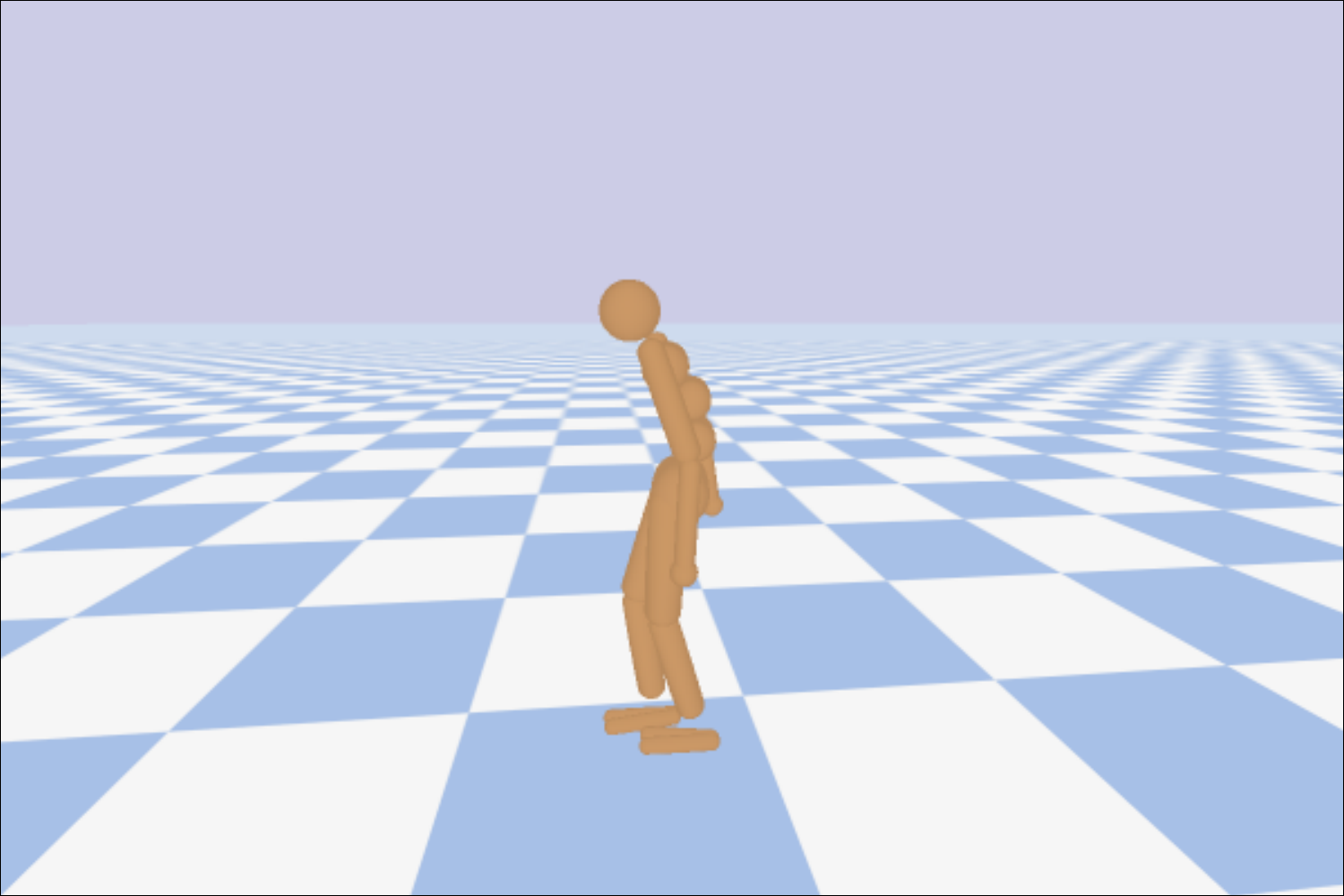} \end{minipage}
 \\ \vspace{1mm}
 \begin{minipage}{0.114\textwidth} \centering \includegraphics[width=1.00\textwidth,trim={80 40 80 40},clip]{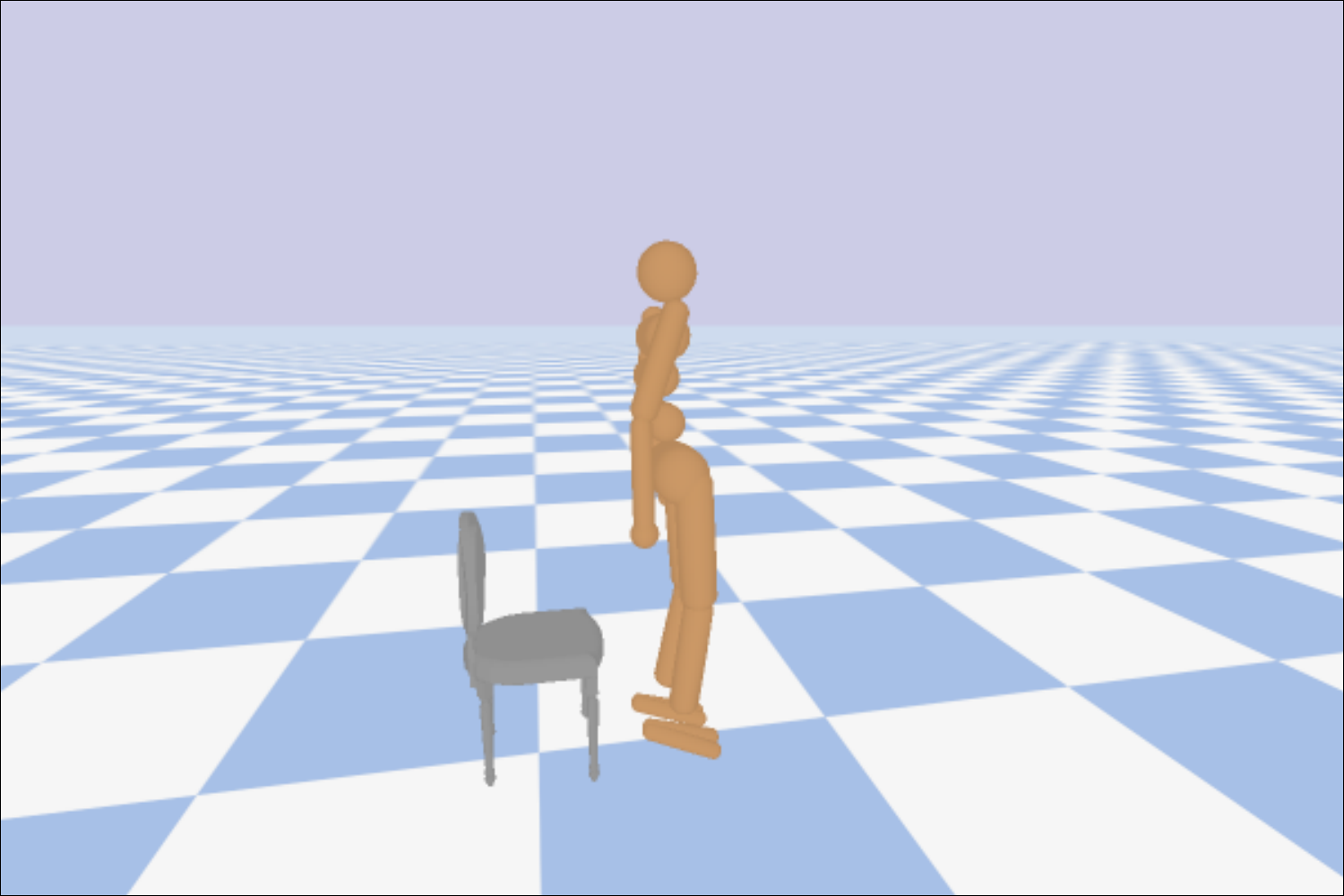} \end{minipage}
 \begin{minipage}{0.114\textwidth} \centering \includegraphics[width=1.00\textwidth,trim={80 40 80 40},clip]{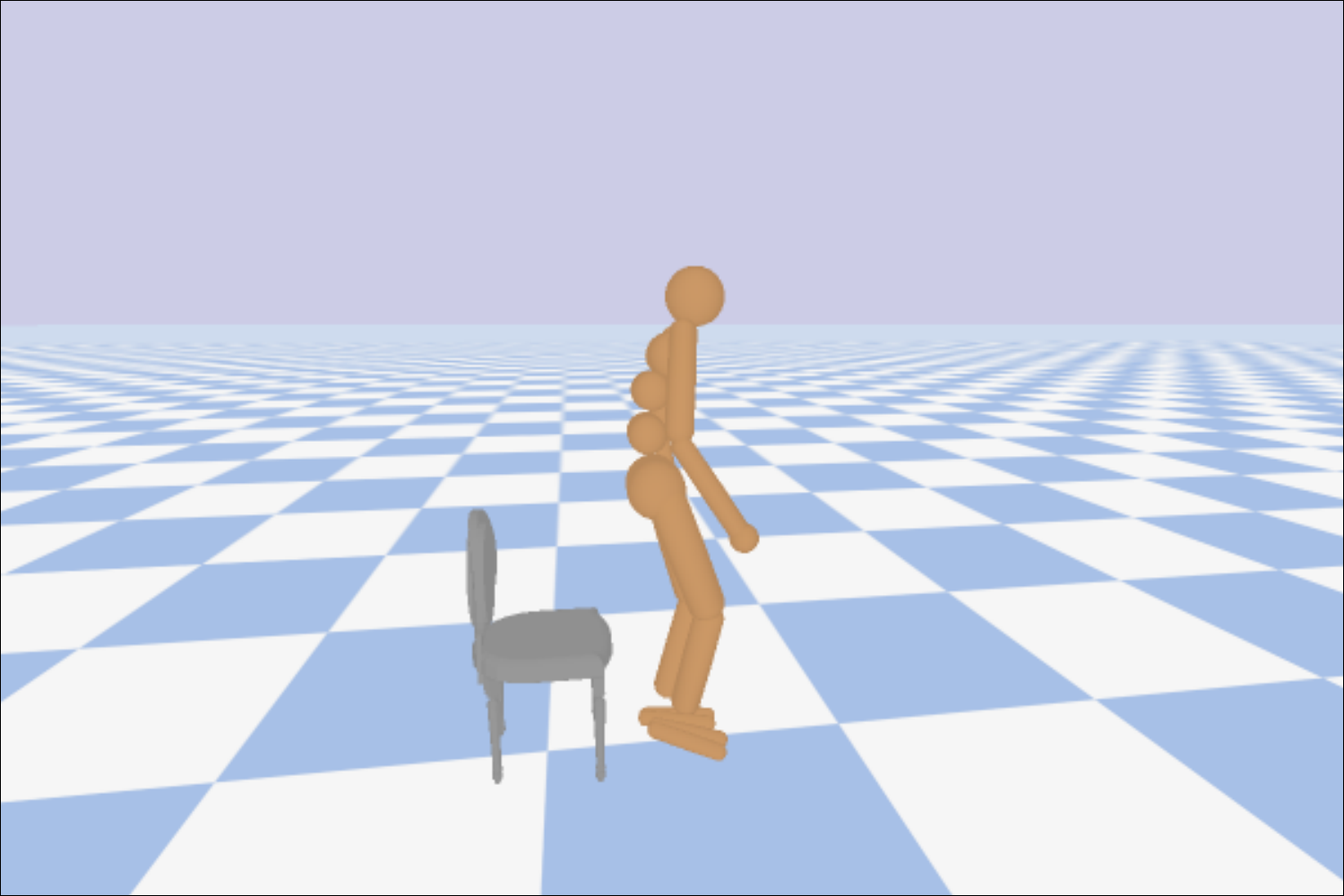} \end{minipage}
 \begin{minipage}{0.114\textwidth} \centering \includegraphics[width=1.00\textwidth,trim={80 40 80 40},clip]{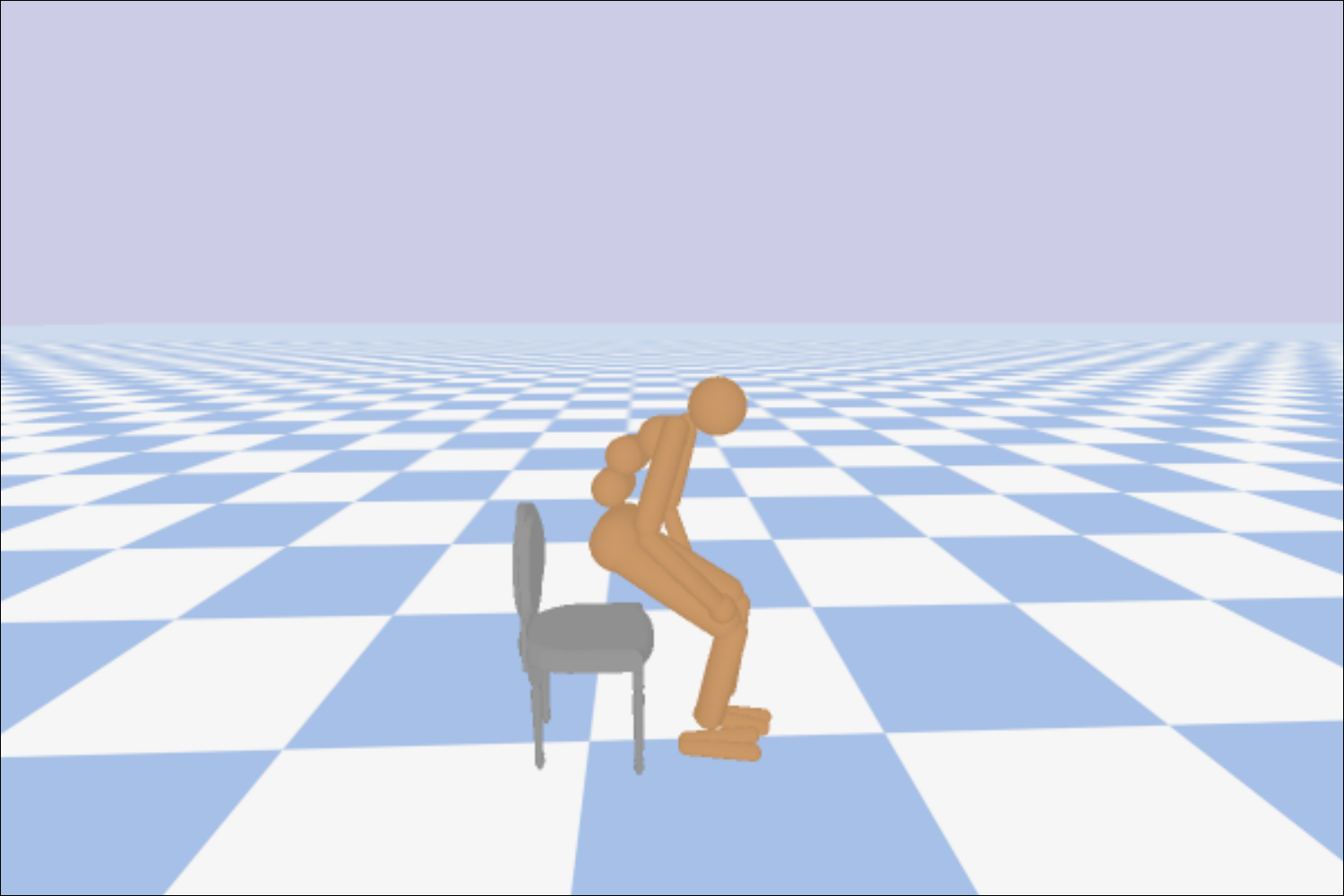} \end{minipage}
 \begin{minipage}{0.114\textwidth} \centering \includegraphics[width=1.00\textwidth,trim={80 40 80 40},clip]{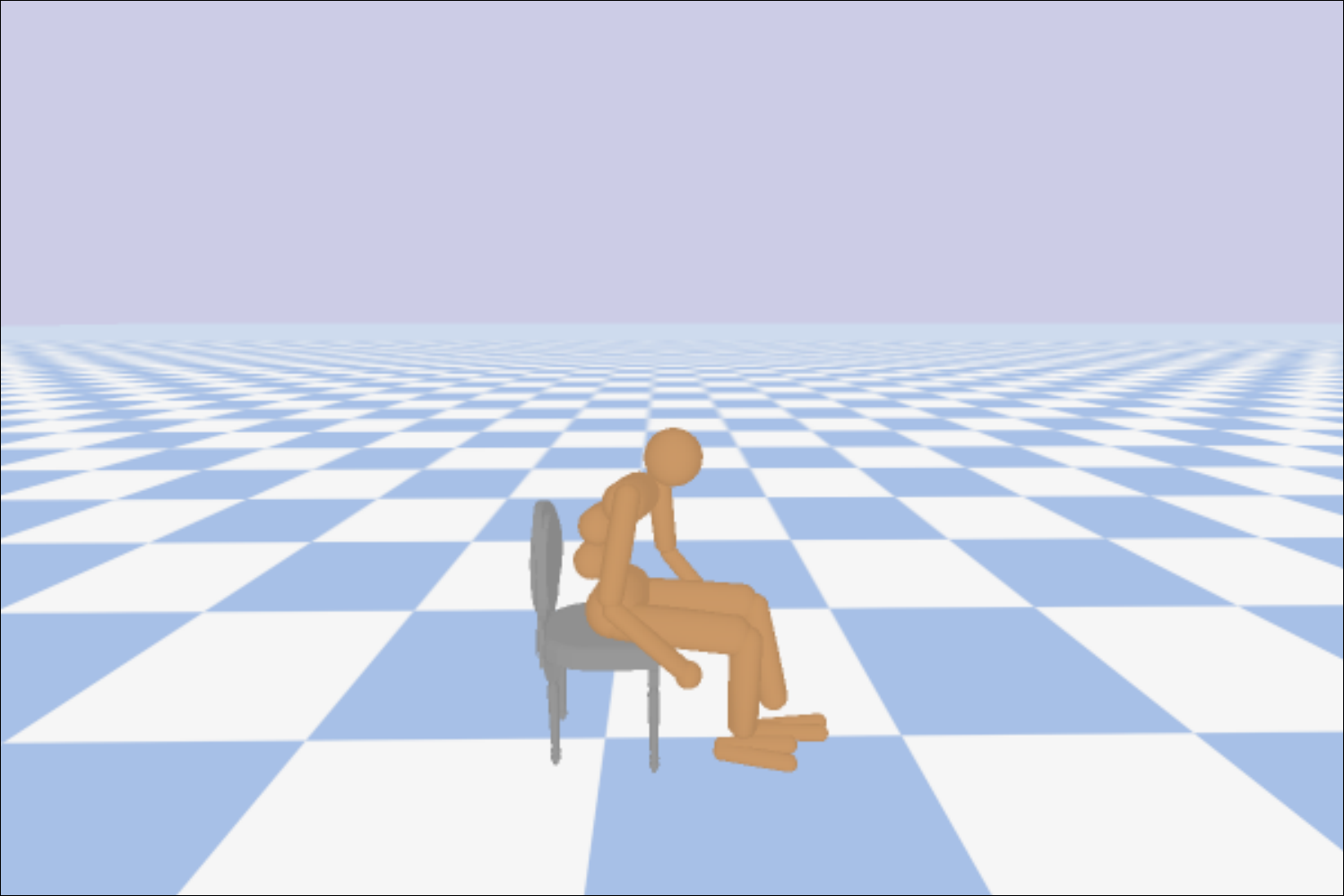} \end{minipage}
 \caption{\small Execution of subtasks. From top to bottom: \textit{forward
walking}, \textit{target directed walking}, \textit{left turn}, \textit{right
turn}, and \textit{sit}.}
 \label{fig:qual-subtask}
\end{figure}

\begin{figure*}[t]
 \centering
 \begin{minipage}{0.12\textwidth} \centering \includegraphics[width=1.00\textwidth]{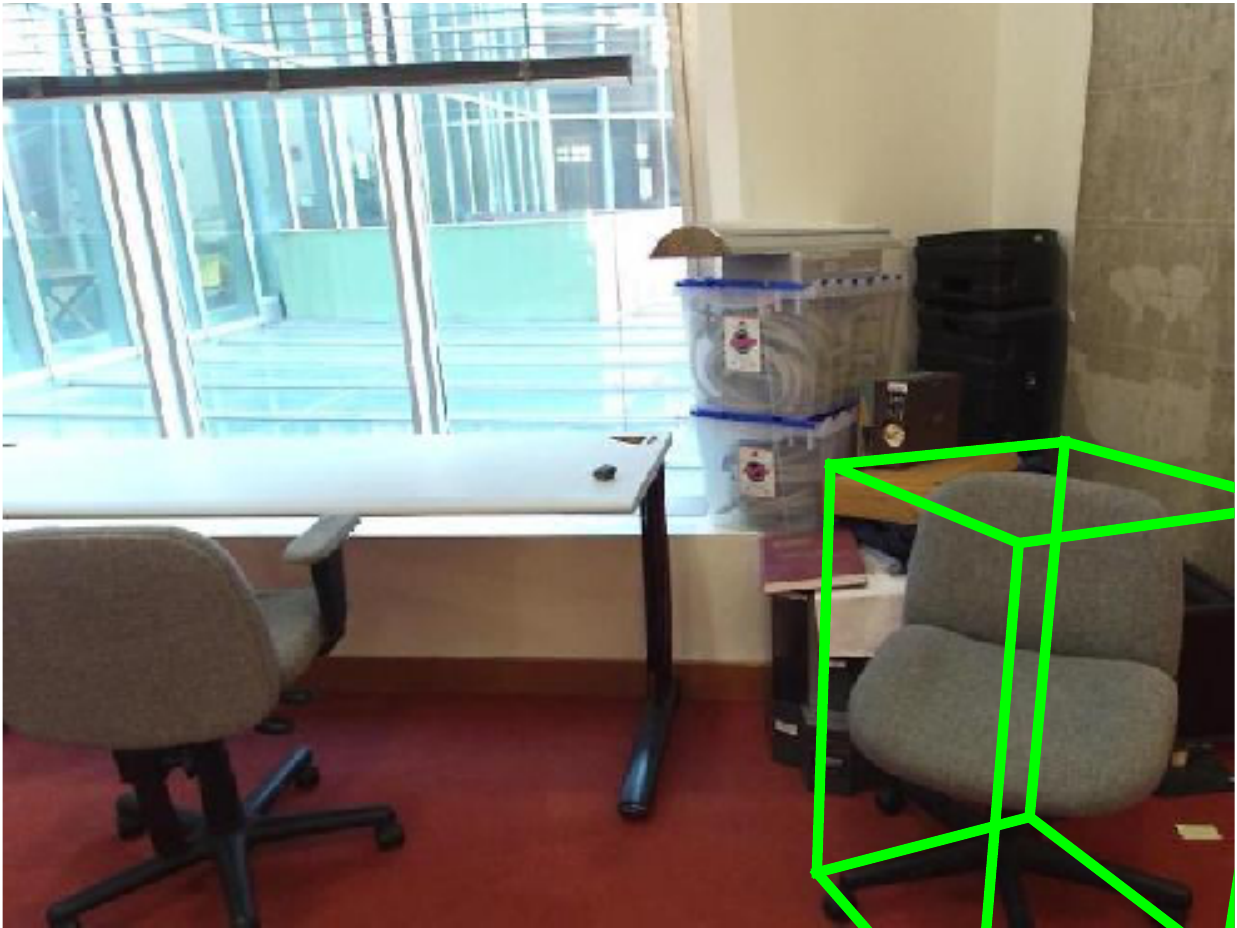} \end{minipage}
 \begin{minipage}{0.12\textwidth} \centering \includegraphics[width=1.00\textwidth]{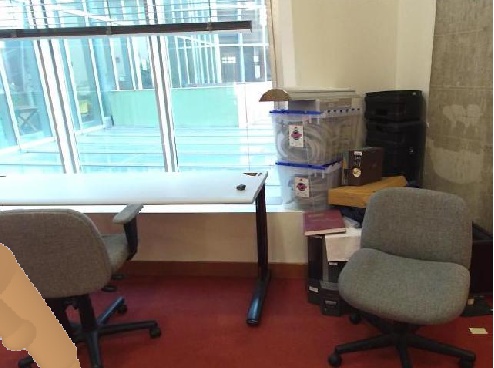} \end{minipage}
 \begin{minipage}{0.12\textwidth} \centering \includegraphics[width=1.00\textwidth]{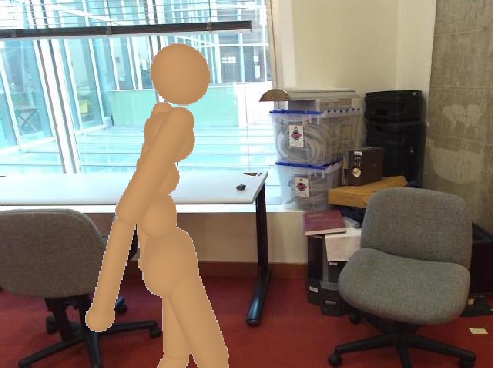} \end{minipage}
 \begin{minipage}{0.12\textwidth} \centering \includegraphics[width=1.00\textwidth]{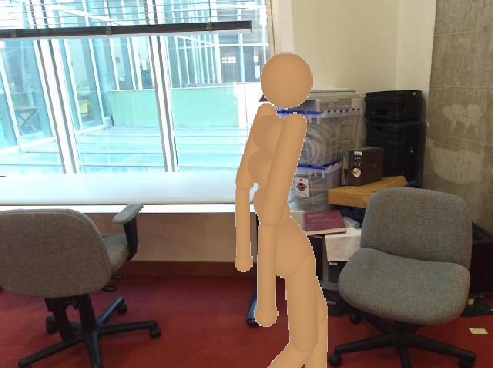} \end{minipage}
 \begin{minipage}{0.12\textwidth} \centering \includegraphics[width=1.00\textwidth]{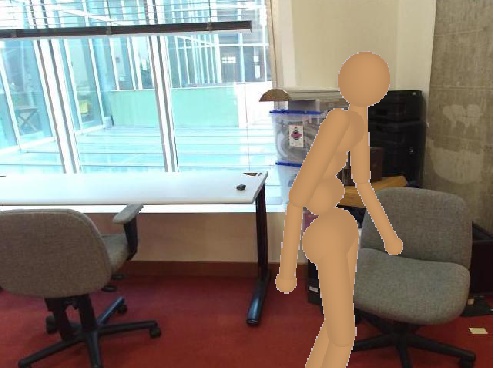} \end{minipage}
 \begin{minipage}{0.12\textwidth} \centering \includegraphics[width=1.00\textwidth]{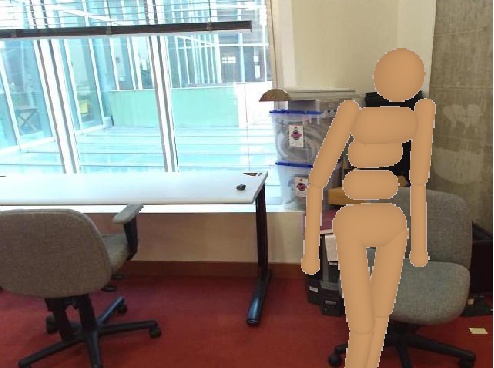} \end{minipage}
 \begin{minipage}{0.12\textwidth} \centering \includegraphics[width=1.00\textwidth]{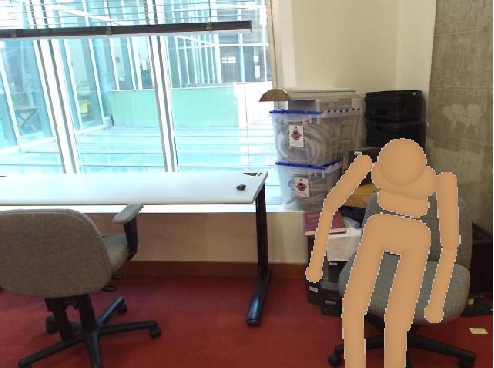} \end{minipage}
 \begin{minipage}{0.12\textwidth} \centering \includegraphics[width=1.00\textwidth]{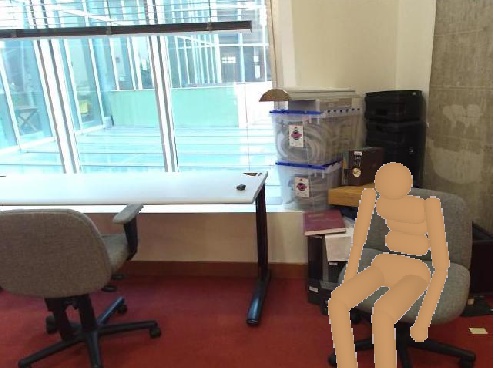} \end{minipage}
 \\ \vspace{1mm}
 \begin{minipage}{0.12\textwidth} \centering \includegraphics[width=1.00\textwidth]{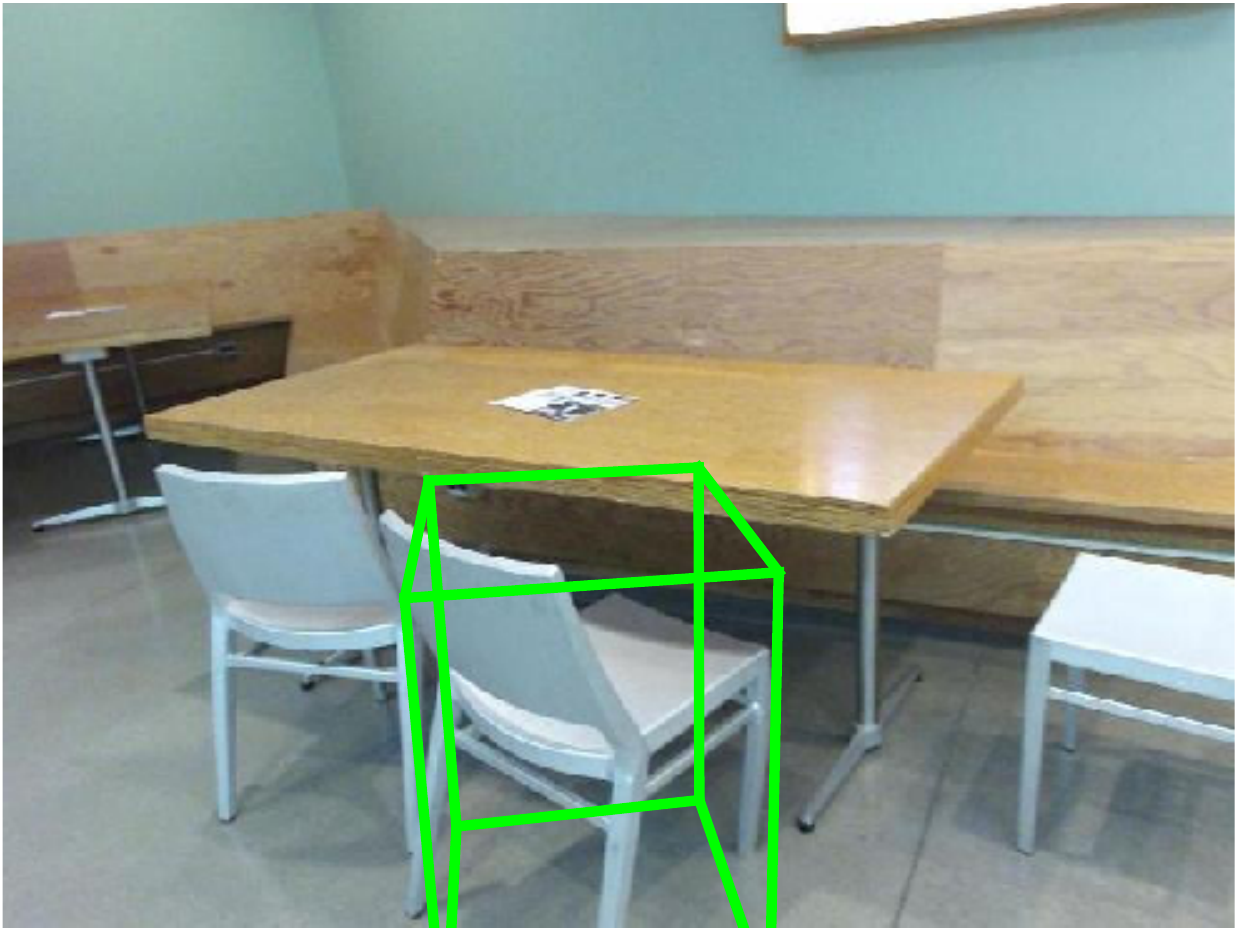} \end{minipage}
 \begin{minipage}{0.12\textwidth} \centering \includegraphics[width=1.00\textwidth]{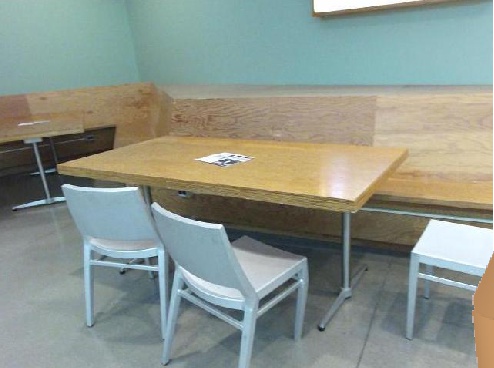} \end{minipage}
 \begin{minipage}{0.12\textwidth} \centering \includegraphics[width=1.00\textwidth]{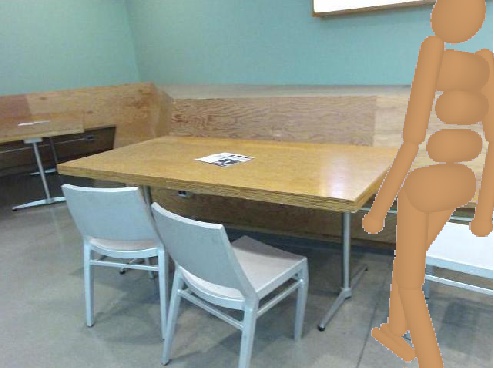} \end{minipage}
 \begin{minipage}{0.12\textwidth} \centering \includegraphics[width=1.00\textwidth]{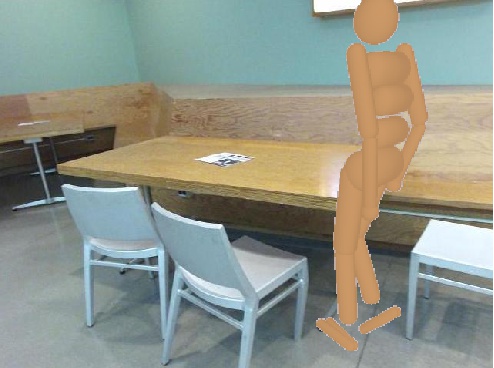} \end{minipage}
 \begin{minipage}{0.12\textwidth} \centering \includegraphics[width=1.00\textwidth]{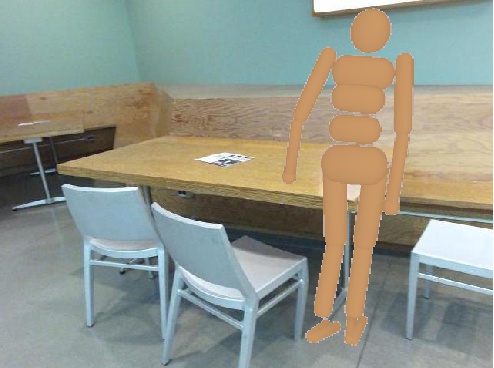} \end{minipage}
 \begin{minipage}{0.12\textwidth} \centering \includegraphics[width=1.00\textwidth]{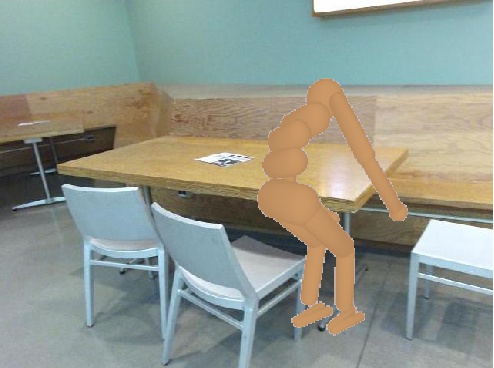} \end{minipage}
 \begin{minipage}{0.12\textwidth} \centering \includegraphics[width=1.00\textwidth]{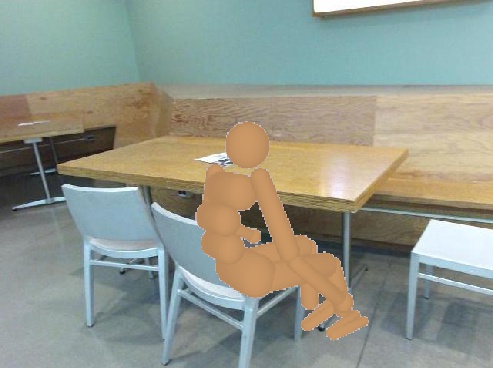} \end{minipage}
 \begin{minipage}{0.12\textwidth} \centering \includegraphics[width=1.00\textwidth]{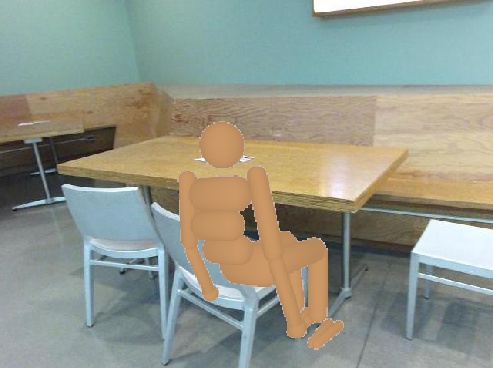} \end{minipage}
 \caption{\small Synthesizing sitting motions from a single image. The first
column shows the 3D reconstruction output from~\cite{huang:eccv2018}.}
 \label{fig:application}
\end{figure*}

\section{Qualitative Results on Subtasks}

We show qualitative results of the individual subtask controllers trained using
their corresponding reference motions. Each row in Fig.~\ref{fig:qual-subtask}
shows the humanoid performance of one particular subtask: walk in one direction
(row 1), following a target (row 2), turn in place both left (row 3) and right
(row 4), and sit on a chair (row 5).

\section{Motion Synthesis in Human Living Space}

We show a vision-based application of our approach by synthesizing sitting
motions from a single RGB image that depicts human living space with chairs.
First, we recover the 3D scene configuration using the method by Huang et
al.~\cite{huang:eccv2018}. We then align the observed scene with the simulated
environment using the detected chair and its estimated 3D position and
orientation. This enables us to transfer the synthesized sitting motion to the
observed scene. Fig.~\ref{fig:application} shows two images rendered with
synthesized humanoid motion. While the motion looks physically plausible in
these examples, this is not always the case in general, since we do not model
the other objects (e.g. tables) in the scene. An interesting future direction
is to learn the motion by simulating scenes with cluttered objects. It is also
possible to synthesize motions based on the humans observed in the image, given
the recent advance on extracting 3D human pose from a single
image~\cite{peng:siggraphasia2018}.

\end{document}